\PassOptionsToPackage{table}{xcolor}

\documentclass[onecolumn]{fairmeta}

\usepackage{hyperref}

\usepackage{amsmath}
\usepackage{amssymb}
\usepackage{mathtools}
\usepackage{bm}
\usepackage{microtype}
\usepackage{graphicx}
\usepackage{subcaption}
\usepackage{booktabs}
\usepackage{epigraph}
\usepackage{xspace}
\usepackage{xstring}

\usepackage{wrapfig}
\usepackage{array}
\usepackage{nicefrac}
\usepackage{amsfonts}
\usepackage{url}
\newcommand{\score}[4]{#1.#2$_{\tiny \pm  \text{\scriptsize #3.#4}}$}

\usepackage{nicematrix}
\newcommand{\pmSep}{%
  \ifnum\theiRow>2\relax 
    $\,\pm\,$%
  \fi
}
\usepackage{multirow}

\usepackage{tcolorbox}
\tcbuselibrary{skins}
\tcbuselibrary{breakable}
\usepackage{float}
\usepackage{siunitx}
\newcommand{\circa}{{\raise.17ex\hbox{$\scriptstyle\sim$}}}

\usepackage{pifont}
\newcommand{\cmark}{\ding{51}}%
\newcommand{\xmark}{\ding{55}}%

\usepackage[textsize=tiny]{todonotes}

\usepackage{tikz}
\usetikzlibrary{patterns}

\newtcolorbox{questionbox}[1]{
  colback=metabg,
  colframe=metafg,
  fonttitle=\bfseries,
  title=#1
}

\definecolor{icmlblue}{RGB}{0,76,153} 
\definecolor{icmlgray}{gray}{0.15}
\definecolor{icmllight}{gray}{0.96}

\tcbset{
  icmlbox/.style={
    enhanced,
    breakable,
    boxrule=0.4pt,
    colframe=icmlblue!65!black,
    colback=icmllight,
    coltitle=black,
    fonttitle=\bfseries,
    left=1.0mm,right=1.0mm,top=0.8mm,bottom=0.8mm,
    boxsep=1.2mm,
    arc=1.2mm,
    outer arc=1.2mm,
    titlerule=0pt,
    toptitle=0.4mm,
    bottomtitle=0.2mm,
    attach title to upper,
    title={#1},
  },
  icmlbox/.default={}
}
\newtcolorbox{callout}[1][]{%
  enhanced, breakable, frame hidden,
  left=2mm,right=2mm,top=2mm,bottom=2mm,
  boxrule=1pt,
  colback=metabg,
  arc=10pt,
  before skip=15pt,
  grow to left by=3pt,
  grow to right by=3pt,
  #1
}

\newcommand{\benchmark}{EgoBabyVLM\xspace}

\title{\benchmark: Benchmarking Cross-Modal Learning from Naturalistic Egocentric Video Data}

\author[1*\dagger]{Dongyan Lin}
\author[1*\dagger]{Phillip Rust}
\author[1,\dagger]{Angel Villar Corrales}
\author[2]{Alvin W. M. Tan}
\author[1]{Mahi Luthra}
\author[1]{Charles-Éric Saint-James}
\author[1]{Rashel Moritz}
\author[3]{Sheila Krogh-Jespersen}
\author[1]{Vanessa Stark}
\author[1]{Surya Parimi}
\author[1]{Jiayi Shen}
\author[1]{Youssef Benchekroun}
\author[1]{Yosuke Higuchi}
\author[1]{Martin Gleize}
\author[1]{Tom Fizycki}
\author[4]{Nicolas Hamilakis}
\author[4]{Manel Khentout}
\author[5]{Sho Tsuji}
\author[1\ddagger]{Balázs Kégl}
\author[1]{Juan Pino}
\author[2]{Michael C. Frank}
\author[1,4]{Emmanuel Dupoux}

\affiliation[1]{Meta Superintelligence Labs}
\affiliation[2]{Stanford University}
\affiliation[3]{Meta Reality Labs}
\affiliation[4]{École Normale Supérieure}
\affiliation[5]{The University of Tokyo}

\contribution[*]{Equal contribution}
\contribution[\dagger]{Core contributors}
\contribution[\ddagger]{Work done at Meta}

\abstract{
Children acquire language grounding with remarkable robustness from limited visuo-linguistic input in ways that surpass today's best large multimodal models. Recent research suggests current vision-language models (VLMs) trained on curated web data fail to generalize to the sparse, weakly-aligned egocentric streams produced by wearable devices, embodied agents, and infant head-cams—and no fixed evaluation pipeline exists for measuring progress on this regime.
We train VLMs on datasets with varying degrees of semantic alignment between visual and linguistic inputs, including naturalistic infant and adult egocentric videos, and evaluate them with a comprehensive suite spanning multimodal language grounding and unimodal vision and language tasks.
At the core of this suite is \emph{Machine-DevBench}, a corpus-grounded benchmark of lexical and grammatical competence, automatically generated from the model's training vocabulary across logarithmic frequency bins to eliminate the train/eval mismatch and low statistical power of prior developmental benchmarks.
Our results show that current VLM paradigms hinge on the tight semantic alignment of curated data and fail to exploit the weakly-aligned signal that dominates naturalistic egocentric input—the very regime in which humans thrive.
To motivate progress, we introduce the \benchmark Challenge to drive the development of models capable of grounded language learning from the kind of naturalistic data that human infants experience.
}

\date{\today}
\correspondence{\email{dongyanlin@meta.com}, \email{philliprust@meta.com}}

\metadata[Code]{\url{https://github.com/facebookresearch/egobabyvlm}}
\metadata[Homepage]{\url{https://facebookresearch.github.io/egobabyvlm}}
\metadata[Leaderboard]{\url{https://facebookresearch.github.io/egobabyvlm/leaderboard.html}}

\begin{document}

\maketitle
\section{Introduction}
\label{section:intro}

\vspace{-1mm}
{\small\textit{Instead of trying to produce a programme to simulate the adult mind, why not rather try to produce one which simulates the child's? If this were then subjected to an appropriate course of education one would obtain the adult brain.}}
\vspace{-2mm}

\hfill {\small--- Alan Turing, 1950}


Human children develop robust visual and linguistic systems efficiently from sparse, naturalistic egocentric experience~\citep{frank2023bridging, warstadt2023findings, long2024babyview, vong2024grounded}, whereas today's vision--language models (VLMs) require web-scale curated data and still fail to generalize to the kind of inputs that wearable devices and embodied agents produce: WearVQA reports only 24--52\% QA accuracy in egocentric wearable scenarios~\citep{chang2025wearvqa}, and WAGIBench shows that even SOTA VLMs cannot reliably infer intent from wearable footage~\citep{veerabadran2025benchmarking}. This generalization gap reflects a deeper mismatch in semantic alignment between modalities: curated image--caption pairs are tightly aligned by construction, whereas egocentric video and the spontaneous speech that accompanies it are sparse and only weakly aligned moment-by-moment. With high-quality curated pretraining data finite~\citep{villalobos2022will},
developing methods that learn effectively from raw, naturalistic streams is becoming increasingly important.

We study these questions by treating vision--language models (VLMs) as computational models of children's multimodal learning, training them on sensory inputs comparable in scale to those experienced by young children.
We compared four datasets spanning curated captions to developmentally plausible, naturalistic egocentric video (see Tab.~\ref{tab:dataset_comparison}).
%

Using these datasets, we trained contrastive (CLIP+) and generative (LLaVA) VLMs as baselines. To evaluate the resulting models, we introduce \emph{Machine-DevBench}, a scalable benchmark inspired by developmental psychology literature that is designed to assess language grounding in ways that parallel children's language understanding. Our benchmark evaluates emerging lexical and grammatical competence through corpus-grounded, contrastive vision–language tasks spanning noun recognition, property attribution, and eight core grammatical constructions.

Additionally, to study the common assumption in both cognitive science and machine learning that cross-modal learning boosts unimodal capabilities through multimodal fusion, we evaluated the unimodal encoders before and after multimodal finetuning.
Our evaluations include standard vision and text benchmarks, as well as a novel Visual Property Swap (VP-Swap) task designed to evaluate the benefit of introducing visual information into a text encoder through multimodal fusion objective.

Our contributions are threefold.
First, we quantified the cross-modality semantic alignment using several complementary methods on datasets representing different degrees of naturalism, and revealed that infant egocentric data exhibits substantially weaker alignment than the curated datasets that today's VLMs are trained on, or even instructional videos.
Second, we present the \benchmark Challenge Suite: a fixed-size, publicly available baby egocentric video training corpus paired with our Machine-DevBench multimodal language grounding benchmark and a suite of unimodal vision and language evaluation tasks. \benchmark is publicly available and includes both a dedicated evaluation pipeline and a public leaderboard.
Third, by benchmarking both contrastive (CLIP+) and generative (LLaVA) VLMs on the \benchmark suite, we show that both architecture families perform near chance on naturalistic egocentric data, while approaching off-the-shelf pretrained CLIP~\citep{radford2021learning} on curated captions, indicating that data quality and semantic alignment---rather than architectural choice or scale alone---drive language acquisition and grounding.
This helps explain the generalization gap of deployed VLMs in egocentric settings such as WearVQA~\citep{chang2025wearvqa} and WAGIBench~\citep{veerabadran2025benchmarking}.

\section{Related Work}

\subsection{Vision-Language Models and the Egocentric Generalization Gap}

Progress on VLMs has been driven by both contrastive and generative architectures. CLIP \citep{radford2021learning} pioneered contrastive image--text learning at web scale, enabling zero-shot transfer to downstream tasks. Flamingo \citep{alayrac2022flamingo} extended this with a late-fusion architecture supporting in-context multimodal learning, and LLaVA \citep{liu2023visual} established the now-standard recipe of pairing a pretrained vision encoder with a generative language model via a trained projector. These advances have produced strong results on broad-coverage VLM benchmarks such as MMBench \citep{liu2024mmbench} and MMMU \citep{yue2024mmmu}, which assess adult-level capabilities under web-scale curated training.

These benchmarks, however, do not characterize VLM behavior on the sparse, weakly-aligned egocentric streams that wearable devices and embodied agents produce. A growing body of egocentric benchmarks documents this generalization gap across complementary axes: long-form temporal reasoning (EgoSchema~\citep{mangalam2023egoschema}), first-person perception and planning (EgoThink~\citep{cheng2023egothink}, EgoPlan-Bench~\citep{chen2023egoplan}), and embodied scene understanding (OpenEQA~\citep{majumdar2024openeqa}), with frontier VLMs falling well short of human performance on each. This gap motivates evaluation infrastructure targeted at the regime in which deployed multimodal systems must operate.

\subsection{Measuring Cross-Modal Semantic Alignment}

Automatic metrics based on pretrained VLMs, such as CLIPScore \citep{hessel-etal-2021-clipscore} or VQAScore \citep{vqascore}, are commonly used to quantify prompt--output consistency in text-to-image/video generation. While typically used to compare models trained on the same dataset, we propose a derivative method to assess the intrinsic semantic alignment of different datasets.
Beyond generation, prior work on video--language learning shows that narration--visual correspondence is often noisy and temporally misaligned in instructional video, motivating objectives and models that better handle weak alignment or explicitly model temporal correspondence \citep{miech-etal-2020, xu-etal-2021-videoclip, han-etal-2022, yang2023tempclr, lin2024multigranularity}. These efforts, however, are not primarily targeted at naturalistic egocentric data, where relevant visual evidence may be particularly sparse and intermittent.
Most recently, \citet{tan2025assessing} studied alignment in such egocentric settings and, consistent with our findings, reported sparse semantic alignment.
Those analyses, however, are restricted to CLIP and does not assess the downstream consequences of this sparsity.
The present work evaluates a broader range of models---including those explicitly trained on video---and quantifies how semantic alignment quality affects downstream performance.

\subsection{Children-Inspired Machine Learning}

\textbf{Datasets. }Developmental psychology has long recognized the value of studying children's naturalistic experiences. Early work by \citet{fausey2016faces} collected 143 hours of infant egocentric video using head-mounted cameras, revealing developmental trajectories in visual experience (from faces to hands). SAYCam \citep{sullivan2021saycam} provided \circa400 hours of audiovisual recordings from three children, enabling preliminary investigations of learning from child-perspective data. BabyView \citep{long2024babyview} further scaled this approach with high-resolution egocentric videos from 31 children spanning 6--30 months of age, totaling more than 800 hours in their 2025.1 version---sufficient for training modern neural networks while maintaining developmental plausibility.

\textbf{Models. }CVCL \citep{vong2024grounded} pioneered using child egocentric data for vision--language modeling, training CLIP on 61 hours of SAYCam data from a single child and demonstrating emergent language grounding through embedding-based evaluation. This work established that models can learn word--object associations from naturalistic infant input, though questions remain about the robustness and generality of such learning \citep{vong2025robustness}.

\textbf{Evaluation benchmarks.} Developmental benchmarks pioneered evaluating models against psychology paradigms: DevBench \citep{tan2024devbench} for multimodal lexicon, syntax, and semantics; KiVA \citep{yiu2024kiva} and BabyVision \citep{chen2026babyvision} for visual reasoning; BabyVLM \citep{wang2025babyvlm} and BabyVLM-v2 \citep{wang2025babyvlmv2} for pretraining on developmental video; and BabyLM \citep{warstadt2023findings} and BabySLM \citep{lavechin2023babyslm} for developmentally-constrained text and speech. These benchmarks, however, are typically limited by small sample sizes, narrow linguistic coverage, and vocabulary mismatch when applied to models trained on small naturalistic corpora.

\providecommand{\yescell}{\textcolor{green!55!black}{\cmark}}
\providecommand{\nocell}{\textcolor{red!70!black}{\xmark}}

\begin{table}[t]
    \centering
    \caption{Comparison of \benchmark with prior developmentally-inspired benchmarks.}
    \label{tab:benchmark_comparison}
    \footnotesize
    \resizebox{1.0\textwidth}{!}{%
    \setlength{\tabcolsep}{0.95pt}
    \renewcommand{\arraystretch}{1.}
    \begin{tabular}{l|ccc|ccccc}
        \toprule
         & \multicolumn{3}{c|}{\textbf{Reference Baselines}} & \multicolumn{5}{c}{\textbf{Evaluation Suite}} \\
        \cmidrule(lr){2-4} \cmidrule(lr){5-9}
        \textbf{Benchmark}
         & \rotatebox{25}{Generative}
         & \rotatebox{25}{Contrastive}
         & \rotatebox{25}{Egocentric}
         & \rotatebox{25}{Multimodal}
         & \rotatebox{25}{Language}
         & \rotatebox{25}{Dense-Visual}
         & \rotatebox{25}{Corpus-Based}
         & \rotatebox{25}{OOD-Styles}
         \\
        \midrule
        DevBench~{\scriptsize\citep{tan2024devbench}}      & \yescell & \yescell & \yescell  & \yescell & \yescell & \nocell  & \nocell  & \yescell \\
        BabyLM~{\scriptsize\citep{warstadt2023findings}}   & \yescell & \nocell  & \nocell  & \yescell & \yescell & \nocell  & \nocell  & \yescell \\
        BabyVLM~{\scriptsize\citep{wang2025babyvlm}}       & \yescell & \yescell & \yescell  & \yescell & \nocell  & \nocell  & \nocell  & \nocell \\
        BabyVLM-v2~{\scriptsize\citep{wang2025babyvlmv2}}  & \yescell & \nocell  & \yescell  & \yescell & \nocell  & \nocell  & \nocell  & \nocell \\
        BabyVision~{\scriptsize\citep{chen2026babyvision}} & \yescell  & \nocell  & \nocell  & \yescell & \nocell  & \nocell  & \nocell  & \yescell \\
        KiVA~{\scriptsize\citep{yiu2024kiva}}              & \yescell  & \nocell  & \nocell  & \yescell & \nocell  & \nocell  & \nocell  & \nocell \\
        \midrule
        \textbf{\benchmark (Ours)}            & \yescell & \yescell & \yescell & \yescell & \yescell & \yescell & \yescell & \yescell \\
        \bottomrule
    \end{tabular}
	}
    \vspace{2pt}
\end{table}

Table~\ref{tab:benchmark_comparison} summarizes how our proposed \benchmark advances over prior developmentally-inspired benchmarks. On the \emph{reference baseline} axis, prior work either omits infant-data baselines (BabyVision, KiVA) or uses only the smaller SayCam corpus and a single architecture family (e.g., generative-only in BabyVLM-v2); \benchmark provides both contrastive (CLIP+) and generative (LLaVA) baselines pretrained on BabyView, the largest naturalistic infant egocentric corpus to date. On the \emph{evaluation} axis, \benchmark spans multimodal, language, and dense visual probes; supports flexible customization to any training corpus via Machine-DevBench and LongTail-Swap; samples evaluation vocabulary in-domain from the model's training distribution; and tests both realistic and cartoon visual inputs to probe visual OOD generalization.

\vspace{-2mm}

\section{Characterizing the Semantic Alignment of Children's Egocentric Input}
\label{q1}

\vspace{-2mm}

We hypothesize that naturalistic egocentric input such as BabyView and Ego4D exhibits substantially weaker semantic cross-modal alignment---correspondence between visual content and accompanying language---than the curated datasets VLMs are trained on, such as COCO.

\begin{table}[t]
  \centering
    \begin{minipage}[t]{0.58\textwidth}
      \vspace{0pt}
      \centering
      \footnotesize
      \setlength{\tabcolsep}{2.0pt}
      \renewcommand{\arraystretch}{1.2}
      \resizebox{\linewidth}{!}{%
      \begin{NiceTabular}{lrrrr}
        \toprule
        \textbf{Dataset} & \textbf{BabyView} & \textbf{Ego4D} & \textbf{HowTo*} & {\scriptsize\textbf{COCO-MC$^+$}} \\
        \midrule
        Video [hours] & 863 & 2550 & 1111 & -- \\
        Speech [hours] & 673 & 485 & 774 & -- \\
        Avg. utterance length [words] & 5 & 4 & 13 & 10 \\
        Semantic alignment [\scriptsize$\mathcal{A}_{\text{ViT-B/16}}$] & 0.012 & 0.017 & 0.193 & $\bm{0.916}$ \\
        Semantic alignment [\scriptsize$\mathcal{A}_{\text{PE-Core-G}}$] & 0.059 & 0.043 & 0.418 & $\bm{0.969}$ \\
        Speech coverage ratio & 0.78 & 0.19 & 0.70 & -- \\
        \bottomrule
      \end{NiceTabular}%
      }
      \vspace{1em}
      \caption{Comparison of multimodal datasets. *subset of HowTo100M. $^+$COCO enriched with images from MetaCLIP dataset.}
      \label{tab:dataset_comparison}
    \end{minipage}
    \hfill
    \begin{minipage}[t]{0.38\textwidth}
      \vspace{0pt}
      \centering
      \includegraphics[width=\linewidth]{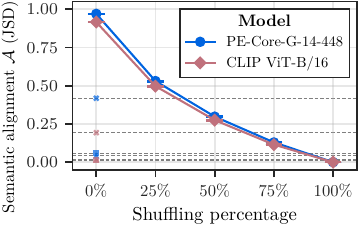}
      \captionof{figure}{Semantic alignment scores for progressively shuffled COCO-MC. {\scriptsize $\blacksquare$}~BabyView, $\blacktriangle$ Ego4D, $\boldsymbol{\times}$ HowTo.}
      \label{fig:coco_shuffling}
    \end{minipage}
%
\end{table}

\begin{wrapfigure}{r}{0.45\textwidth}
    \centering
    \vspace{-1em}







\begin{tikzpicture}[scale=0.5, every node/.style={scale=0.5}]
      \definecolor{circleA}{HTML}{C0717C}
      \definecolor{circleB}{HTML}{0064E0}
      \definecolor{darkgray}{rgb}{0.15, 0.15, 0.15}

      \def\Rvideo{2.5}
      \def\Rspeech{1.118}   
      \def\Raligned{0.43}    

      \def\Sx{-0.6}\def\Sy{0}
      \def\Ax{-0.8}\def\Ay{-0.3}

      \fill[circleB, opacity=0.45] (0,0) circle (\Rvideo);
      \fill[circleA, opacity=0.6] (\Sx,\Sy) circle (\Rspeech);
      \fill[pattern=north east lines, pattern color=darkgray] (\Ax,\Ay) circle (\Raligned);

      \draw[darkgray, thick] (0,0) circle (\Rvideo);
      \draw[darkgray, thick] (\Sx,\Sy) circle (\Rspeech);
      \draw[darkgray, thick] (\Ax,\Ay) circle (\Raligned);

      \node[align=center, font=\Large\bfseries] at (1.6, 0) {Video\\Data};

      \node[align=right, font=\Large, anchor=east] (speech) at (-3.5, 1.4)
          {\textbf{Speech}\\\textbf{Data}};
      \draw[->, darkgray, thick] (speech.east) -- (-1.5, 0.7);

      \node[align=right, font=\Large, anchor=east] (aligned) at (-3.5, -1.5)
          {\textbf{{Semantically}}\\\textbf{{aligned}}};
      \draw[->, darkgray, thick] (aligned.east) -- (\Ax,\Ay);
  \end{tikzpicture}
    \caption{We aim to measure how much of a child's paired speech and visual input are semantically aligned (shaded area).}
    \label{fig:venn-diagram}
    \vspace{-1em}
\end{wrapfigure}

To quantify semantic alignment between visual and linguistic modalities, we transcribed videos with WhisperX \citep{bain2023whisperx} and kept the clean utterances (see App.~\ref{method:whisperx}), based on which we extract time-aligned frames. We quantified semantic alignment via the Jensen-Shannon divergence (JSD) between the empirical distributions of CLIPScores for matched temporal pairings versus shuffled pairings.\footnote{The original CLIPScore is defined as $w \cdot \max\left( \cos(\mathbf{c}, \mathbf{v}),\ 0 \right)$
 \citep{hessel-etal-2021-clipscore}. We used unscaled ($w = 1$) and unclipped scores to capture the full distributional properties of the embedding space, rather than aiming for positive sample-level scores.} We extracted embeddings using an off-the-shelf ViT-B/16~CLIP model \citep{radford2021learning} trained on web-scale image-text pairs. To ensure our alignment measurements generalize beyond image-trained models, we also evaluated using the Perception Encoder \citep[PE-Core-G-14-448;][]{bolya2025perception}, a state-of-the-art CLIP variant extended to video data. We report 95\% confidence intervals obtained via bootstrapping (1000 samples). Formally, given a collection of $N$ image-text pairs with visual embeddings $\mathbf{v_i}$ and textual embeddings $\mathbf{c_i}$, we define the alignment score $\mathcal{A}$ as:
\begin{equation}
\mathcal{A} = \text{JSD}(P \parallel Q), \qquad P \sim \{ \cos(\mathbf{c}_i, \mathbf{v}_i) \}_{i=1}^N, \qquad Q \sim \{ \cos(\mathbf{c}_{\pi(i)}, \mathbf{v}_i) \}_{i=1}^N .
\end{equation}
and $\pi(\cdot)$ denotes a random permutation of indices. $\mathcal{A}$ is bounded between $[0, 1]$, with a higher score indicating stronger semantic relatedness in original pairings compared to random combinations. To validate this metric's sensitivity, we systematically degraded COCO's alignment by creating shuffled variants at 25\%, 50\%, 75\%, and 100\% levels, where image-caption pairs were randomly reassigned while maintaining dataset size.. We describe our approach for extracting video clips (for Perception Encoder) and sampling frames (for CLIP) with temporally aligned utterance transcriptions in detail in App.~\ref{method:frame-extraction}.

To further validate and contextualize our semantic alignment metric, we also compared with three additional scoring methods: 1) using VQAScore \citep{vqascore}, 2) measuring semantic textual similarity between the transcribed utterance and a synthetic caption of the video clip or frame, and 3) CLIPScoring with a variant of Perception Encoder that we LoRA-finetuned on these synthetic captions. We present these results in App.~\ref{app:semantic_alignment_further}.

\paragraph{Dataset comparison.} Tab.~\ref{tab:dataset_comparison} summarizes dataset scale, speech coverage, and alignment. We used the 2025.1 version of BabyView containing \circa863h of video and \circa673h of speech. For HowTo100M, we used a random subset that is on the same order of magnitude in size as BabyView which we hereafter refer to as ``HowTo''.
For COCO, to match the volume of visual content in the video datasets, we curated a dataset version enriched with MetaCLIP~1.2 \citep{xu2024demystifying, xu-etal-2024-altogether} images (see App~\ref{methods:coco-mc} for details), hereafter termed \textbf{COCO-MC}.
To isolate the effect of semantic alignment from data scale, we matched all four corpora to the same order of magnitude of utterances and frames (see Tab.\ref{tab:dataset_sizes}) when training our LLaVA and CLIP+ baselines.
Across datasets, semantic alignment differs by orders of magnitude: curated image--caption pairs (COCO-MC) are highly aligned, instructional video (HowTo) is moderately aligned, and naturalistic egocentric video (BabyView, Ego4D) is weakly aligned (Tab.~\ref{tab:dataset_comparison}). This confirms our hypothesis that children develop robust visual and language capabilities from weakly aligned data. But exactly how misaligned is the data? To calibrate our semantic alignment metric, we systematically degraded COCO-MC's alignment by shuffling increasing percentages of image-caption pairs, which degrades alignment monotonically (Fig.~\ref{fig:coco_shuffling}), validating that $\mathcal{A}$ is sensitive to controlled misalignment. Relative to this calibration, instructional video remains substantially more aligned than naturalistic egocentric video, while the latter approaches the near-random regime (Tab.~\ref{tab:dataset_comparison}).

Taken together, naturalistic egocentric data sits at the weakly-aligned end of the spectrum, while curated captions sit at the highly-aligned end (Tab.~\ref{tab:dataset_comparison}; also see \citet{tan2025assessing}). We next introduce the \benchmark Challenge, which uses this weakly-aligned naturalistic corpus as a training ground for evaluating models' ability to acquire grounded language under realistic, child-centric conditions.

\begin{figure}[t]
    \centering
    \includegraphics[width=\textwidth]{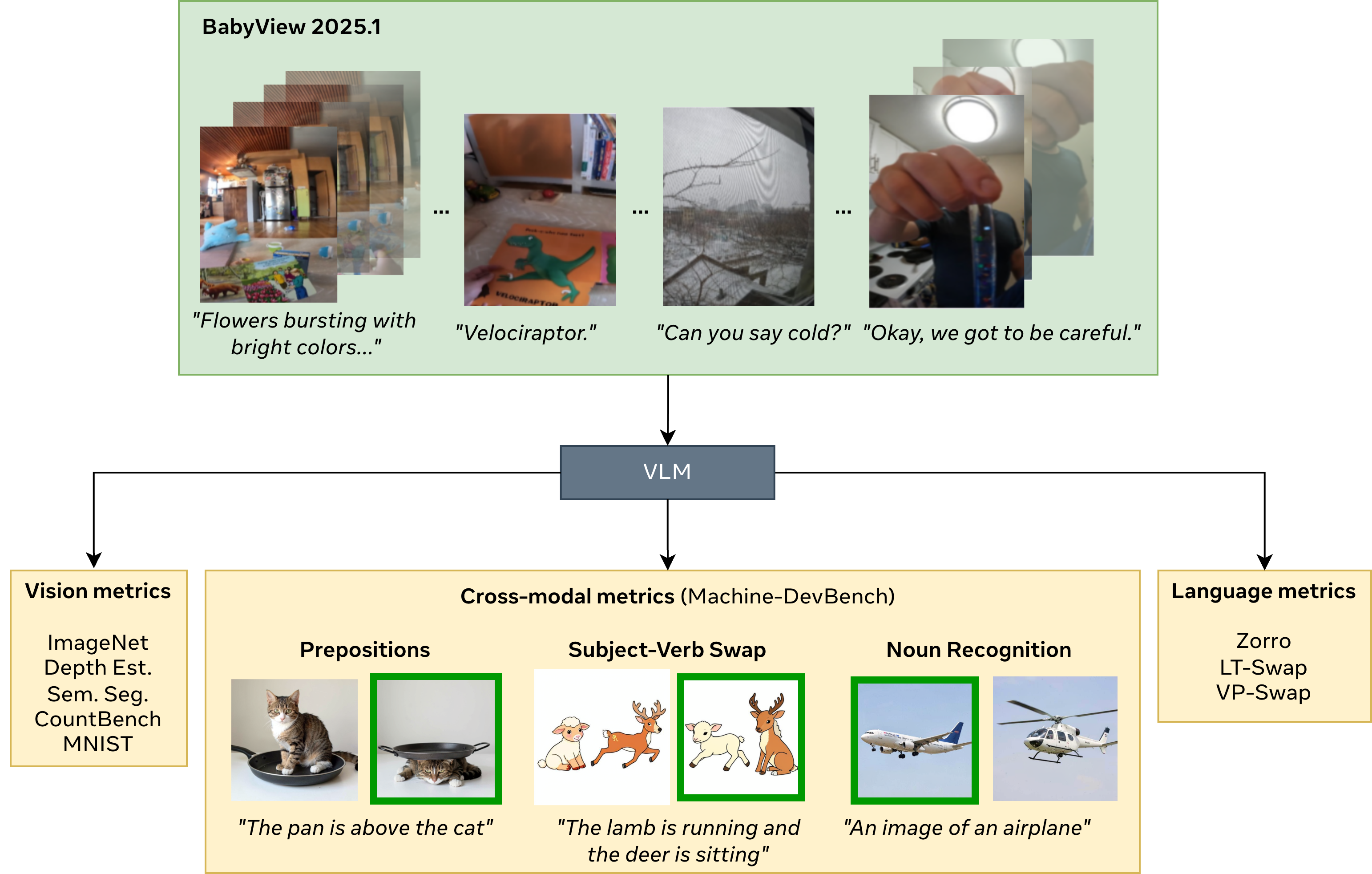}
    \caption{
    	Overview of the \benchmark challenge.
    	Models are trained exclusively on a fixed-size, developmentally plausible egocentric dataset---BabyView 2025.1---and evaluated along three complementary dimensions: (1) unimodal vision benchmarks, including object recognition, semantic segmentation, depth estimation, and counting; (2) unimodal language benchmarks, including syntactic and semantic understanding; and (3) cross-modal language grounding through our proposed Machine-DevBench benchmark, which evaluates diverse semantic and grammatical capabilities.
    }
    \label{fig:benchmark}
\end{figure}

\section{The \benchmark Challenge}
\label{babyvlm-challenge}

The analysis in the preceding section reveals a core tension: naturalistic egocentric data---the kind infants learn from---is weakly semantically aligned and sparse, which contrasts with the curated large-scale datasets modern VLMs are trained on.
The \benchmark Challenge asks: can we close this gap through architectural or algorithmic innovation alone, without changing the data?


Formally, the \benchmark challenge, illustrated in Fig.~\ref{fig:benchmark}, aims to develop VLMs solely trained on egocentric baby data that can:
\begin{enumerate}
    \item Learn robust language grounding from sparse, naturalistic and weakly-aligned multimodal input, as measured by Machine-DevBench (App.~\ref{methods:machinedevbench}) and other language understanding benchmarks.
    \item Maintain or improve unimodal encoder quality through cross-modal learning, rather than degrading it
\end{enumerate}

Models must be trained on the BabyView 2025.1 dataset, openly available on Databrary,\footnote{\url{https://databrary.org/}} which consists of 863 hours of naturalistic egocentric videos collected from children's perspective.
Beyond this corpus, no additional image, video, text, or audio data may be used for pretraining or finetuning, including for any encoder initialization.
This restriction makes the challenge a meaningful test of \emph{learning from infant-comparable input} rather than reliance on transferred web-scale priors.

Each submission is evaluated on three families of tasks (Fig.~\ref{fig:benchmark}):
\begin{itemize}
    \item \textbf{Cross-modal language grounding.} Evaluated using Machine-DevBench, a corpus-grounded probe of lexical and grammatical competence, consisting of over $\sim$3{,}700 trials across 10 tasks, sampled from the model's own training vocabulary across logarithmic frequency bins. Submissions are scored as a \textbf{Lexical} subgroup aggregate (noun + adjective recognition), a \textbf{Grammatical} subgroup aggregate (eight sentence-level tasks) and an \textbf{Overall} aggregate. We refer to App.~\ref{methods:machinedevbench} for further details.
    \item \textbf{Vision tasks.} Six benchmarks evaluating the trained vision encoder, organized into two subgroups.
    The \textbf{Object recognition} subgroup includes ImageNet-1k under three evaluation protocols---$k$-nearest neighbors, linear probing, and ABX discrimination (a zero-shot method inspired by phoneme classification~\citep{poli2025fastabx})---as well as MNIST~\citep{lecun1998mnist} with  linear probing and ABX discrimination, and COCO-Stuff semantic segmentation~\citep{caesar2018cocostuff}.
    The \textbf{Visual properties} subgroup includes NYU Depth~v2 depth estimation~\citep{silberman2012indoor} and CountBench object counting~\citep{paiss2023teaching} using linear and ABX evaluation protocols.
    Submissions report both subgroup aggregates, the \textbf{Overall} vision score, and the \emph{delta} of the \textbf{Overall} score relative to a unimodal-only baseline (DINOv2 trained on the same frames), hence penalizing cross-modal training that degrades vision representations. See App.~\ref{methods:vision-evals} for details.
    \item \textbf{Language tasks.} Five benchmarks evaluating the trained text encoder, organized into two subgroups.
    The \textbf{Syntax} subgroup includes Zorro for grammatical acceptability~\citep{warstadt2020blimp}, together with the LongTail-Swap~\citep{algayres2025longtail} InflectionSwap and AgreementSwap variants.
    The \textbf{Semantics} subgroup includes LongTail-Swap WordSwap and our novel Visual Property Swap (VP-Swap; App.~\ref{methods:vp-swap}) task, designed to probe grounded vision-related semantics.
    Submissions report both subgroup aggregates, the \textbf{Overall} language score, , and the \emph{delta} of the \textbf{Overall} score relative to a unimodal-only baseline (e.g. BERT or GPT2 trained on the same text corpus), hence penalizing cross-modal training that degrades text representations
    See App.~\ref{methods:text-evals} for details.

\end{itemize}

%


We publicly release the \benchmark benchmark, including  a public leaderboard
, as well as an open-source
codebase 
containing training and evaluation harness.
This ensures hat all submissions are scored under the same calibrated pipeline, enabling fair and reproducible comparison across models.
We consider the challenge successful if BabyView-only submissions close a meaningful fraction of the Machine-DevBench gap to web-scale VLMs without degrading same-data unimodal baselines---evidence that the
primary bottleneck for grounded language learning from naturalistic input is algorithmic rather than data scale.
More broadly, we hope \benchmark helps drive the development of architectures and training objectives better suited to the weakly-aligned input children receive, advancing both our understanding of language acquisition and the design of data-efficient grounded AI.

\section{Findings}
\label{sec:findings}

The analyses in Section~\ref{q1} established that naturalistic egocentric data is weakly aligned compared to curated vision–language datasets. We now investigate how this alignment gap impacts learning and what it implies for the \benchmark Challenge, structured around two questions.


\begin{callout}{\textbf{Q1:}}
{How does semantic alignment impact cross-modal learning, especially on egocentric data?}
\end{callout}

\paragraph{Motivation and approach.}
To isolate the effect of semantic alignment on cross-modal learning, we fixed the model and training recipe and varied only the training dataset (keeping dataset size fixed, see Tab.~\ref{tab:dataset_sizes}), including COCO-MC, HowTo, Ego4D, BabyView, and shuffled-COCO-MC variants to probe the alignment--performance relationship.
We consider two baseline VLM architectures. \textbf{CLIP+}, a contrastive model trained with three interleaved objectives---InfoNCE between image and text embeddings, the DINOv2 self-supervised loss on its visual encoder, and masked language modeling on its text encoder---following the rationale that children engage in multiple forms of learning simultaneously. This interleaved optimization strategy prevents catastrophic forgetting (see Apps.~\ref{sec:multimodal_alignment} and~\ref{methods:interleaved-training} for further details and App.~\ref{section:ablation} for ablation studies on interleaved training).
\textbf{LLaVA}, a generative baseline \citep{liu2023visual} trained using an image captioning objective.
Both architectures use a DINOv2 ViT-B/14 vision encoder and a text encoder (BERT-base for CLIP+, GPT-2 for LLaVA), each pretrained on the target dataset's visual domain and its corresponding text corpus (Apps.~\ref{methods:dinov2-training} and~\ref{methods:bert-training}).

\begin{table}[t]
    \centering
    %
    %
    %
    \setlength{\tabcolsep}{3pt}
    \begin{minipage}[t]{0.51\textwidth}
        \vspace{0pt}
        \centering
        \caption{
            Language grounding accuracy (\%) on Machine-DevBench, aggregated by task type, for off-the-shelf models (top) and our models trained on the small-scale datasets (bottom). We report mean and std across 3 training seeds, with best two results in each section / column bold and underlined, respectively. Full per-task accuracies in App.~Tab.~\ref{tab:model_accuracy_full}.
        }
        \label{tab:model_accuracy}
        \resizebox{\linewidth}{!}{%
        \footnotesize
        \setlength{\tabcolsep}{2.5pt}
        \renewcommand{\arraystretch}{1.12}
        \begin{tabular}{l c c | c}
        \toprule
        \textbf{Model}
          & \textbf{\shortstack[c]{Lexical\\Agg$\uparrow$}}
          & \textbf{\shortstack[c]{Grammatical\\Agg$\uparrow$}}
          & \textbf{\shortstack[c]{Overall\\Agg$\uparrow$}}
          \\
        \midrule
        Chance
          & {50.0} & {50.0} & {50.0} \\
        CLIP-L {\scriptsize\citep{radford2021learning}}
          & {87.3} & {70.4} & {78.8} \\
        LLaVA-v1.6-Mistral-7B {\scriptsize\citep{liu2023visual}}
          & {\underline{90.0}} & {\textbf{96.8}} & {\textbf{93.4}} \\
        PE-Core-B {\scriptsize\citep{bolya2025perception}}
          & {\textbf{98.7}} & {\underline{78.6}} & {\underline{88.6}} \\
        Gemma 4 {\scriptsize\citep{gemma2024}}
          & {92.1} & {75.5} & {83.8} \\
        \midrule
        BabyView-CLIP+
          & {\score{53}{4}{0}{7}} & {\score{53}{8}{2}{4}} & {\score{53}{6}{1}{5}} \\
        BabyView-LLaVA
          & {\score{51}{2}{2}{5}} & {\score{49}{7}{4}{3}} & {\score{50}{5}{3}{4}} \\
        Ego4D-CLIP+
          & {\score{50}{2}{2}{8}} & {\score{51}{5}{3}{9}} & {\score{50}{8}{3}{4}} \\
        Ego4D-LLaVA
          & {\score{48}{8}{1}{9}} & {\score{49}{9}{4}{0}} & {\score{49}{3}{3}{0}} \\
        HowTo-CLIP+
          & {\score{60}{6}{2}{1}} & {\score{51}{6}{3}{6}} & {\score{56}{1}{2}{8}} \\
        HowTo-LLaVA
          & {\score{52}{4}{1}{2}} & {\score{50}{7}{3}{9}} & {\score{51}{5}{2}{5}} \\
        COCO-MC-CLIP+
          & {\textbf{\score{68}{8}{1}{8}}} & {\underline{\score{64}{1}{3}{0}}} & {\textbf{\score{66}{5}{2}{4}}} \\
        COCO-MC-LLaVA
          & {\score{59}{6}{2}{3}} & {\textbf{\score{64}{2}{2}{3}}} & {\score{61}{9}{2}{3}} \\
        COCO-MC\textsubscript{shuf25\%}-CLIP+
          & {\underline{\score{67}{8}{0}{8}}} & {\score{64}{0}{3}{6}} & {\underline{\score{65}{9}{2}{2}}} \\
        COCO-MC\textsubscript{shuf25\%}-LLaVA
          & {\score{56}{7}{2}{5}} & {\score{58}{8}{2}{9}} & {\score{57}{7}{2}{7}} \\
        COCO-MC\textsubscript{shuf50\%}-CLIP+
          & {\score{65}{3}{0}{8}} & {\score{63}{9}{2}{8}} & {\score{64}{6}{1}{8}} \\
        COCO-MC\textsubscript{shuf50\%}-LLaVA
          & {\score{54}{5}{0}{9}} & {\score{56}{0}{2}{2}} & {\score{55}{3}{1}{6}} \\
        COCO-MC\textsubscript{shuf75\%}-CLIP+
          & {\score{64}{3}{2}{0}} & {\score{58}{1}{3}{8}} & {\score{61}{2}{2}{9}} \\
        COCO-MC\textsubscript{shuf75\%}-LLaVA
          & {\score{50}{6}{1}{5}} & {\score{51}{1}{3}{7}} & {\score{50}{9}{2}{6}} \\
        COCO-MC\textsubscript{shuf100\%}-CLIP+
          & {\score{49}{2}{4}{8}} & {\score{48}{2}{4}{4}} & {\score{48}{8}{4}{6}} \\
        COCO-MC\textsubscript{shuf100\%}-LLaVA
          & {\score{48}{8}{1}{5}} & {\score{50}{0}{2}{8}} & {\score{49}{4}{2}{2}} \\
        \bottomrule
        \end{tabular}%
        }
    \end{minipage}
    \hfill
    \begin{minipage}[t]{0.45\textwidth}
        \captionof{figure}{
            \textbf{(top)} Semantic alignment vs.\ Machine-DevBench. Color: dataset; marker: encoder for image/video--text cosine similarities; error bars: bootstrapped 95\% CIs ($n{=}1000$).
            \textbf{(bottom)} Noun-recognition accuracy vs.\ training-set word frequency. Shaded area: $\pm 1$~std across 3 seeds.}
        \label{fig: aligment}
        \vspace{0pt}
        \centering
        \includegraphics[width=\linewidth]{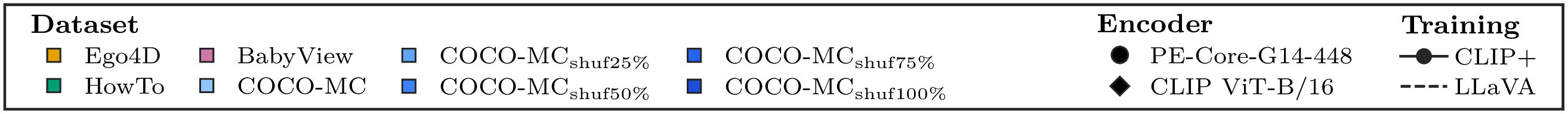}\\[2pt]
        \includegraphics[width=0.85\linewidth]{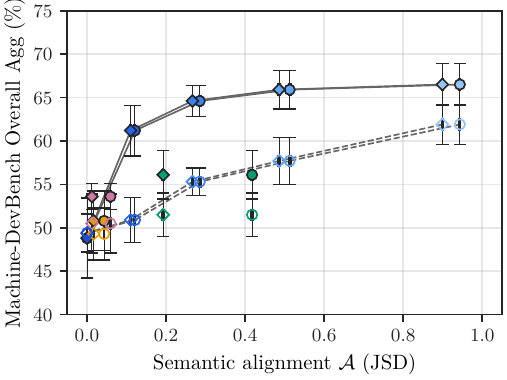}\\[0pt]
        \includegraphics[width=0.85\linewidth]{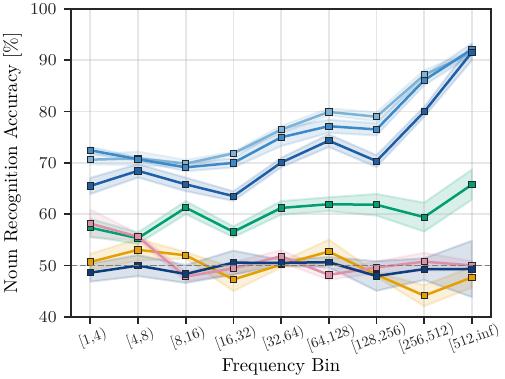}
    \end{minipage}
\end{table}


Our hypothesis is straightforward: if semantic alignment facilitates language grounding, then models trained on more aligned datasets should achieve better performance on language understanding tasks.
To test this, we evaluate each trained model on Machine-DevBench and report a \textbf{Lexical} aggregate, a \textbf{Grammatical} aggregate, and an \textbf{Overall} aggregate (Tab.~\ref{tab:model_accuracy}; full per-task accuracies in App.~Tab.~\ref{tab:model_accuracy_full}).
Because Machine-DevBench samples evaluation vocabulary from each model's own training corpus across logarithmic frequency bins, it lets us cleanly relate downstream grounding performance to (i) the dataset's semantic alignment and (ii) the train-set frequency of the words being probed---neither of which is possible with prior developmental benchmarks.
To further validate Machine-DevBench, we perform additional cross-modal evaluations on DevBench~\citep{tan2024devbench}, reporting both accuracy and model--human similarity, which are provided in App.~\ref{methods:devbench} (Tab.~\ref{tab:devbench_accuracy}--\ref{tab:devbench_similarity}).

\paragraph{Results.}
As shown in Tab.~\ref{tab:model_accuracy} and Fig.~\ref{fig: aligment}, performance on Machine-DevBench scales monotonically with cross-modal alignment across \textbf{Lexical}, \textbf{Grammatical}, and \textbf{Overall} accuracies.
Weakly aligned naturalistic egocentric datasets (BabyView and Ego4D) yield near-chance performance ($\approx$50\%) whereas instructional video (HowTo) improves to 51--56\%.
In contrast, the curated COCO-MC dataset achieves 62--67\%, with the best-aligned models starting to close the gap with off-the-shelf CLIP-B (78.8\%) despite using orders of magnitude less training data
This trend is consistent for both CLIP+ and LLaVA backbones.
For reference, we additionally report large-scale pretrained VLMs such as LLaVA, PE-Core and Gemma, which exceed 80\% accuracy on \benchmark.

As illustrated in  Fig.~\ref{fig: aligment} (top), the COCO-MC shuffle progression confirms alignment as the causal factor: progressively mismatching captions and images monotonically degrades cross-modal performance (CLIP+: 66.5 $\to$ 65.9 $\to$ 64.6 $\to$ 61.2 $\to$ 48.8 at 0/25/50/75/100\% shuffle), collapsing to chance level at full shuffling.
Per-frequency-bin noun recognition (Fig.~\ref{fig: aligment} bottom) further shows that better-aligned models consistently outperform weaker ones across all frequency bins, with the gap becoming larger for more common nouns.

To further validate these claims, we evaluate our models on the original DevBench~\citep{tan2024devbench} and observe the same overall trend (App.~Tab.~\ref{tab:devbench_accuracy}--\ref{tab:devbench_similarity}).
Importantly, Machine-DevBench scores strongly correlate with both DevBench Agg Accuracy and DevBench Agg Alignment across all trained models  (App.~Fig.~\ref{fig:mdb_vs_devbench_scatter}), further supporting Machine-DevBench as a reliable proxy for the human-curated benchmark.

\begin{table}[t]
    \centering
    \caption{
        Unimodal scores aggregated by task type, before and after cross-modal
        finetuning (FT).
        \textbf{(a)} Vision scores of the DINOv2 encoder.
        \textbf{(b)} Language scores of BERT (for CLIP+) and GPT-2 (for
        LLaVA) text encoders.
        For BabyView/Ego4D/HowTo/COCO-MC, within each Train Data block for
        vision and each Encoder sub-block for text, the highest value is
        bolded. For COCO-MC\textsubscript{shufX\%}, the higher between CLIP and LLaVA is
        bolded.
        Cross-modal finetuning generally yields limited gains for unimodal representations and often degrades
        performance, particularly on language tasks. Full per-task scores in App.~Tab.~\ref{tab:vision_comparison_full} and App.~Tab.~\ref{tab:text_comparison_full}.}

    \label{tab:vision_comparison}\label{tab:text_comparison}
    \setlength{\tabcolsep}{2pt}
    \begin{subtable}[t]{0.49\textwidth}
        \vspace{0pt}
        \centering
        \caption{Evaluation on unimodal vision tasks}
        \label{tab:unimodal vision}
        \resizebox{\linewidth}{!}{%
        \footnotesize
        \setlength{\tabcolsep}{3pt}
        \renewcommand{\arraystretch}{1.05}
        \begin{tabular}{l c c c | c}
        \toprule
        \textbf{\shortstack[c]{Train\\Data}} & \textbf{FT}
            & \textbf{\shortstack[c]{Object\\recogn.\\Agg$\uparrow$}}
            & \textbf{\shortstack[c]{Visual\\props.\\Agg$\uparrow$}}
            & \textbf{\shortstack[c]{Overall\\Agg$\uparrow$}} \\
        \midrule
        Chance & --
            & {1.9} & {3.7} & {2.8} \\
        DINOv2-B/14 {\scriptsize\citep{oquab2023dinov2}} & --
            & {\score{76}{3}{0}{2}}
            & {\score{43}{5}{0}{7}}
            & {\score{59}{9}{0}{4}} \\
        CLIP-B/16 {\scriptsize\citep{radford2021learning}} & --
            & {\score{75}{5}{0}{1}}
            & {\score{45}{7}{0}{8}}
            & {\score{60}{6}{0}{4}} \\
        \midrule
        \multirow{3}{*}{BabyView} & \xmark
            & {\textbf{\score{50}{9}{0}{2}}}
            & {\textbf{\score{39}{2}{0}{6}}}
            & {\textbf{\score{45}{1}{0}{4}}} \\
        & CLIP+
            & \score{50}{9}{0}{4} & \score{38}{0}{0}{4} & \score{44}{4}{0}{4} \\
        & LLaVA
            & {\score{43}{2}{0}{6}}
            & {\score{30}{3}{1}{2}}
            & {\score{36}{7}{0}{9}} \\
        \midrule
        \multirow{3}{*}{Ego4D} & \xmark
            & {\textbf{\score{53}{0}{0}{2}}}
            & {\score{37}{3}{0}{2}}
            & {\score{45}{1}{0}{2}} \\
        & CLIP+
            & {\score{52}{9}{0}{2}}
            & {\textbf{\score{37}{6}{0}{2}}}
            & {\textbf{\score{45}{2}{0}{2}}} \\
        & LLaVA
            & {\score{35}{4}{8}{6}}
            & {\score{27}{9}{4}{4}}
            & {\score{31}{6}{6}{5}} \\
        \midrule
        \multirow{3}{*}{HowTo} & \xmark
            & {\score{55}{1}{0}{2}}
            & {\score{39}{2}{0}{4}}
            & {\score{47}{1}{0}{3}} \\
        & CLIP+
            & {\textbf{\score{55}{1}{0}{1}}}
            & {\textbf{\score{39}{5}{0}{4}}}
            & {\textbf{\score{47}{3}{0}{2}}} \\
        & LLaVA
            & {\score{44}{2}{3}{4}}
            & {\score{32}{8}{2}{3}}
            & {\score{38}{5}{2}{8}} \\
        \midrule
        \multirow{3}{*}{COCO-MC} & \xmark
            & {\textbf{\score{66}{7}{0}{1}}}
            & {\score{44}{1}{1}{6}}
            & {\score{55}{4}{0}{9}} \\
        & CLIP+
            & {\score{66}{4}{0}{1}}
            & {\textbf{\score{44}{5}{0}{8}}}
            & {\textbf{\score{55}{5}{0}{4}}} \\
        & LLaVA
            & \score{57}{8}{0}{4}
            & {\score{40}{2}{2}{9}}
            & \score{49}{0}{1}{7} \\
        \midrule
        COCO-MC\textsubscript{shuf25\%} & CLIP+
            & {\textbf{\score{66}{3}{0}{3}}}
            & {\textbf{\score{44}{6}{1}{0}}}
            & {\textbf{\score{55}{4}{0}{5}}} \\
        & LLaVA
            & \score{53}{9}{0}{7}
            & {\score{36}{4}{0}{6}}
            & \score{45}{2}{0}{7} \\
        COCO-MC\textsubscript{shuf50\%} & CLIP+
            & {\textbf{\score{66}{2}{0}{1}}}
            & {\textbf{\score{44}{8}{1}{1}}}
            & {\textbf{\score{55}{5}{0}{5}}} \\
        & LLaVA
            & \score{54}{4}{0}{5}
            & {\score{38}{5}{1}{4}}
            & \score{46}{4}{1}{0} \\
        COCO-MC\textsubscript{shuf75\%} & CLIP+
            & {\textbf{\score{66}{2}{0}{1}}}
            & {\textbf{\score{44}{6}{1}{3}}}
            & {\textbf{\score{55}{4}{0}{6}}} \\
        & LLaVA
            & \score{13}{7}{2}{9}
            & {\score{22}{9}{2}{8}}
            & \score{18}{3}{2}{8} \\
        COCO-MC\textsubscript{shuf100\%} & CLIP+
            & {\textbf{\score{66}{6}{0}{1}}}
            & {\textbf{\score{44}{4}{1}{2}}}
            & {\textbf{\score{55}{5}{0}{6}}} \\
        & LLaVA
            & \score{14}{7}{2}{9}
            & {\score{22}{3}{2}{1}}
            & \score{18}{5}{2}{5} \\
        \bottomrule
        \end{tabular}%
        }
    \end{subtable}
    \hfill
    \begin{subtable}[t]{0.5\textwidth}
        \vspace{0pt}
        \centering
        \caption{Evaluation on unimodal language tasks}
        \label{tab:unimodal text}
        \resizebox{\linewidth}{!}{%
        \footnotesize
        \setlength{\tabcolsep}{3pt}
        \renewcommand{\arraystretch}{1.07}
        \begin{tabular}{l c c c c | c}
        \toprule
        \textbf{\shortstack[c]{Train\\Data}}
            & \textbf{\shortstack[c]{Encoder}}
            & \textbf{FT}
            & \textbf{\shortstack[c]{Syntax\\Agg$\uparrow$}}
            & \textbf{\shortstack[c]{Semantics\\Agg$\uparrow$}}
            & \textbf{\shortstack[c]{Overall\\Agg$\uparrow$}} \\
        \midrule
        Chance & -- & --
            & {50.0} & {50.0} & {50.0} \\
        \midrule
        \multirow{4}{*}{BabyView} & BERT  & \xmark
            & {\textbf{\score{72}{9}{0}{2}}}
            & {\textbf{\score{71}{8}{0}{5}}}
            & {\textbf{\score{72}{4}{0}{3}}} \\
        & BERT  & CLIP+
            & {\score{71}{8}{0}{0}}
            & {\score{71}{0}{0}{0}}
            & {\score{71}{4}{0}{0}} \\
        & GPT-2 & \xmark
            & {\textbf{\score{61}{7}{0}{9}}}
            & {\textbf{\score{55}{8}{1}{3}}}
            & {\textbf{\score{58}{8}{1}{1}}} \\
        & GPT-2 & LLaVA
            & {\score{55}{9}{2}{2}}
            & {\score{53}{2}{0}{7}}
            & {\score{54}{6}{1}{4}} \\
        \midrule
        \multirow{4}{*}{Ego4D} & BERT  & \xmark
            & {\textbf{\score{76}{8}{0}{7}}}
            & {\textbf{\score{73}{7}{0}{6}}}
            & {\textbf{\score{75}{2}{0}{6}}} \\
        & BERT  & CLIP+
            & {\score{73}{5}{1}{2}}
            & {\score{72}{4}{0}{2}}
            & {\score{73}{0}{0}{7}} \\
        & GPT-2 & \xmark
            & {\textbf{\score{57}{4}{0}{3}}}
            & {\textbf{\score{57}{3}{1}{5}}}
            & {\textbf{\score{57}{3}{0}{9}}} \\
        & GPT-2 & LLaVA
            & {\score{52}{9}{2}{9}}
            & {\score{54}{8}{1}{1}}
            & {\score{53}{9}{2}{0}} \\
        \midrule
        \multirow{4}{*}{HowTo} & BERT  & \xmark
            & {\textbf{\score{76}{4}{0}{7}}}
            & {\textbf{\score{79}{5}{1}{2}}}
            & {\textbf{\score{77}{9}{1}{0}}} \\
        & BERT  & CLIP+
            & {\score{71}{8}{0}{7}}
            & {\score{72}{7}{0}{3}}
            & {\score{72}{3}{0}{5}} \\
        & GPT-2 & \xmark
            & {\textbf{\score{56}{5}{0}{7}}}
            & {\textbf{\score{59}{7}{1}{4}}}
            & {\textbf{\score{58}{1}{1}{1}}} \\
        & GPT-2 & LLaVA
            & {\score{53}{0}{1}{6}}
            & {\score{57}{4}{0}{6}}
            & {\score{55}{2}{1}{1}} \\
        \midrule
        \multirow{4}{*}{COCO-MC} & BERT  & \xmark
            & {\score{71}{1}{0}{8}}
            & {\score{81}{4}{0}{6}}
            & {\score{76}{2}{0}{7}} \\
        & BERT  & CLIP+
            & {\textbf{\score{71}{5}{0}{3}}}
            & {\textbf{\score{81}{4}{0}{1}}}
            & {\textbf{\score{76}{4}{0}{2}}} \\
        & GPT-2 & \xmark
            & {\textbf{\score{58}{4}{1}{9}}}
            & {\textbf{\score{71}{8}{0}{8}}}
            & {\textbf{\score{65}{1}{1}{4}}} \\
        & GPT-2 & LLaVA
            & {\score{55}{8}{0}{8}}
            & {\score{60}{3}{0}{5}}
            & {\score{58}{1}{0}{7}} \\
        \midrule
        COCO-MC\textsubscript{shuf25\%} & BERT  & CLIP+
            & {\textbf{\score{71}{2}{0}{2}}}
            & {\textbf{\score{80}{9}{0}{3}}}
            & {\textbf{\score{76}{1}{0}{2}}} \\
        & GPT-2 & LLaVA
            & {\score{54}{6}{2}{2}}
            & {\score{62}{0}{0}{7}}
            & {\score{58}{3}{1}{5}} \\
        COCO-MC\textsubscript{shuf50\%} & BERT  & CLIP+
            & {\textbf{\score{70}{7}{0}{3}}}
            & {\textbf{\score{80}{5}{0}{2}}}
            & {\textbf{\score{75}{6}{0}{2}}} \\
        & GPT-2 & LLaVA
            & {\score{55}{1}{1}{0}}
            & {\score{63}{0}{0}{4}}
            & {\score{59}{1}{0}{7}} \\
        COCO-MC\textsubscript{shuf75\%} & BERT  & CLIP+
            & {\textbf{\score{69}{8}{0}{7}}}
            & {\textbf{\score{79}{5}{0}{3}}}
            & {\textbf{\score{74}{7}{0}{4}}} \\
        & GPT-2 & LLaVA
            & {\score{53}{9}{1}{8}}
            & {\score{59}{3}{0}{9}}
            & {\score{56}{6}{1}{3}} \\
        COCO-MC\textsubscript{shuf100\%} & BERT  & CLIP+
            & {\textbf{\score{70}{5}{0}{3}}}
            & {\textbf{\score{80}{1}{0}{1}}}
            & {\textbf{\score{75}{3}{0}{1}}} \\
        & GPT-2 & LLaVA
            & {\score{53}{6}{2}{6}}
            & {\score{58}{7}{2}{6}}
            & {\score{56}{1}{2}{6}} \\
        \bottomrule
        \end{tabular}%
    }
    \end{subtable}
\end{table}

\begin{callout}{\textbf{Q2:}}
Does cross-modal learning help unimodal learning, especially on children's egocentric data?
\end{callout}

\begin{figure}[t]
    \centering
    \begin{subfigure}[t]{\linewidth}
        \centering
        \includegraphics[width=0.85\linewidth]{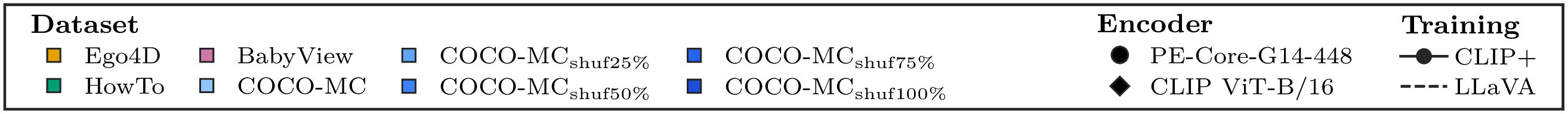}
    \end{subfigure}\\[0.05cm]
    \begin{subfigure}[t]{0.49\linewidth}
        \centering
        \includegraphics[width=\linewidth]{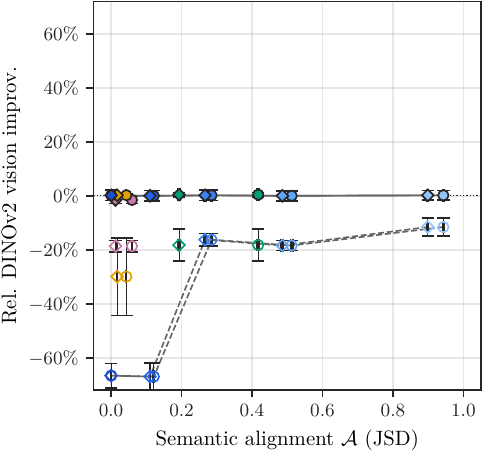}
    \end{subfigure}\hfill
    \begin{subfigure}[t]{0.49\linewidth}
        \centering
        \includegraphics[width=\linewidth]{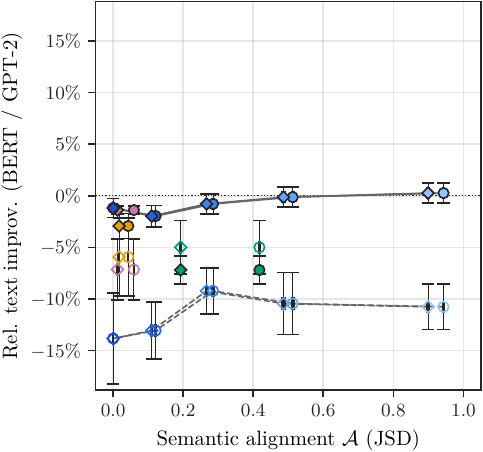}
    \end{subfigure}
    \caption{Effect of cross-modal finetuning on unimodal encoders. Y-axes are the relative improvement on the \textbf{Overall Agg} vision (left) and language (right) scores compared to the unimodal baseline (DINOv2 / BERT or GPT-2 trained on the same data without cross-modal finetuning). Object recognition / Visual properties and Syntax / Semantics subgroup aggregates are reported per-row in Tab.~\ref{tab:unimodal vision} and Tab.~\ref{tab:unimodal text} respectively.}
    \label{fig:q3-figure}
\end{figure}

\paragraph{Motivation and approach.} A key hypothesis in developmental cognitive science is that multimodal alignment guides and improves unimodal learning \citep{vong2024grounded, zhuang2024visual, khorrami2025model}. For instance, seeing objects while hearing their names might help children build better visual representations of object categories. Similar hypotheses have been explored in machine learning~\citep{zhai2022lit, jose2025dinov2, zhuang2024visual}, though it remains unclear how much cross-modal facilitation actually occurs. To assess whether cross-modal training improves unimodal representations under current VLM paradigms, we compared encoder performance before and after CLIP+ or LLaVA finetuning by evaluate the trained vision and text encoders on the unimodal benchmark suites introduced in Sec.~\ref{babyvlm-challenge}.
\paragraph{Results.} For both CLIP+ and LLaVA, the \textbf{Object recognition} subgroup degrades more than \textbf{Visual properties} under cross-modal finetuning---most clearly on COCO-MC and BabyView (Tab.~\ref{tab:unimodal vision}, Fig.~\ref{fig:q3-figure}a; full per-task scores in App.~Tab.~\ref{tab:vision_comparison_full}). On the COCO-MC$_{\text{shuf}}$ series, LLaVA's vision encoder collapses at low alignment and recovers only at high alignment, while CLIP+ degrades more gracefully across the full shuffling range---a robustness we attribute in part to CLIP+'s interleaved unimodal losses, designed to prevent catastrophic forgetting (See App.~\ref{section:ablation} for ablation studies). For language (Tab.~\ref{tab:unimodal text}, Fig.~\ref{fig:q3-figure}b; full per-task scores in App.~Tab.~\ref{tab:text_comparison_full}), cross-modal finetuning never beats the unimodal baseline, and the gap is alignment-independent: LLaVA loses ${\sim}10\text{--}14\%$ across every COCO-MC alignment level, including the well-aligned end, whereas CLIP+ stays within ${\sim}2\%$. The \textbf{Semantics} subgroup (containing the VP-Swap visually grounded text probe) sees essentially no benefit even from the visually grounded objective, suggesting that multimodal fusion provides limited unimodal benefits.

Taken together, these findings present a paradox: infants develop visual and linguistic systems successfully from naturalistic input similar to BabyView, yet our models fail to extract meaningful language grounding from such data or get unimodal boost from cross-modal training. This suggests that current VLMs are limited in the way they exploit learning signals present in noisy, naturalistic data, which is in stark contrast with infants. We hypothesize that solving this challenge will require several advances beyond current methods:
\begin{itemize}
    \item \textbf{Attention mechanisms} that learn which parts of visual input align with which parts of linguistic input, rather than coarse alignment \citep{lee2018stacked}
    \item \textbf{Temporal integration} that tracks references across longer timescales, recognizing that speech often refers to recent past or anticipated future events
    \item \textbf{Social cues} that help identify what speech refers to through gaze direction, pointing, and joint attention  \citep{frank2009using, yu2007unified}
    \item \textbf{Action grounding} that connects language not just to static visual scenes but to interactions and their effects \citep{smith2005development}
    \item \textbf{Robustness to misalignment} through learning objectives that tolerate and learn from weak or absent alignment rather than requiring strong supervision
\end{itemize}

Additionally the development of general architecture that excels on egocentric capabilities would have wide application to generalizable world models and robotics too. Altogether, our findings advocate for more native modality-merging architectural designs as well as algorithmic designs that are more robust to unstructured data for the next generation of adaptive AI.

\section{Conclusion}
We present the \benchmark Challenge, which pairs weakly-aligned naturalistic egocentric data---characterized via multiple complementary alignment metrics in Section~\ref{q1}---with a suite of multimodal and unimodal evaluation tasks, including the novel corpus-grounded Machine-DevBench and Visual Property Swap (VP-Swap) probes.
By benchmarking both contrastive (CLIP+) and generative (LLaVA) VLMs across four distinct datasets, we show that current frameworks overly rely on curated data and fail to extract grounded language from egocentric input: BabyView- and Ego4D-trained models collapse to near chance on Machine-DevBench, while the same architectures trained on the highly aligned COCO-MC approach off-the-shelf CLIP.
We further find that cross-modal training under these recipes provides little to no benefit---and often degrades---the unimodal encoders it is meant to enrich, indicating that the difficulty of learning from naturalistic input is not confined to the multimodal objective alone.
Together, these results may help explain why deployed VLMs fail to generalize to wearable settings such as WearVQA~\citep{chang2025wearvqa} and WAGIBench~\citep{veerabadran2025benchmarking}, where the input distribution is much closer to BabyView than to web-curated image--text pairs.
We hope \benchmark motivates the development of more data-efficient, robust models with human-inspired inductive biases that can learn from the rich but messy multimodal world that humans inhabit.

\section*{Acknowledgements}
We would like to thank Maxime Oquab, Tushar Nagarajan, Robin Algayres, Angelo Ortiz, Jitendra Malik, Yann LeCun, Joelle Pineau for helpful discussions and support on this project. ED in his EHESS role was supported in part by the Agence Nationale pour la Recherche (ANR-17-EURE0017 Frontcog, ANR10-IDEX-0001-02 PSL*) and an ERC grant (InfantSimulator).

\clearpage
\newpage

\bibliography{paper}
\bibliographystyle{icml2026} 

\clearpage
\newpage
\appendix

\section*{Impact Statement}
\label{sec:impact statement}
This paper introduces \benchmark, an evaluation framework for studying data efficiency in vision--language learning under developmentally plausible data regimes, and provides evidence that semantic alignment strongly predicts multimodal grounding performance. We expect the primary positive impact to be improved scientific understanding of what properties of naturalistic multimodal input support language learning, benefiting both machine learning and developmental cognitive science. Progress on this public benchmark may also contribute to more data-efficient and robust multimodal models that rely less on large-scale curated or web-scraped datasets.

Our work also raises ethical considerations because it relies on infant egocentric recordings. Such data can be highly sensitive: it may contain identifying information about children, caregivers, and bystanders, as well as private spaces and conversations. Even when datasets are shared for research, there remains risk of unintended secondary use, re-identification, or use in applications that could enable intrusive monitoring. We therefore encourage future work using these data to follow strict data governance practices (e.g., controlled access where applicable, compliance with dataset terms, minimization of sensitive content, and avoidance of attempts to identify individuals), and to consider whether trained models could be repurposed for surveillance-like or otherwise harmful applications.

Overall, we believe the benefits of a careful, transparent benchmark for studying learning from naturalistic multimodal data outweigh these risks, but we emphasize the importance of responsible data handling and ongoing ethical review as this research area develops.

\section{Limitations}
\label{sec:limitations}

Our study makes several simplifying choices that bound the conclusions one should draw from it. We list them by methodological axis.

\paragraph{Text rather than speech.}
We train and evaluate on WhisperX-transcribed utterances rather than raw audio, sidestepping the central challenges of infant language acquisition: acoustic variability, word segmentation from continuous speech, speaker attribution, and prosodic cues.
Real infants do not receive a clean transcript. Extending EgoBabyVLM to speech-based models---either by replacing the text encoder with an audio encoder, or by training jointly on audio and transcripts---is an important next step.

\paragraph{Frame-- and clip--level vision.} Our vision encoders consume individual frames (CLIP+/LLaVA) or short-clip–level representations (Perception Encoder for alignment scoring).
This discards temporal dynamics that are likely critical for naturalistic egocentric data, where speech often refers to the recent past or anticipated future events, and where caregiver utterances may lag or lead the visually salient moment by seconds. The weak alignment we measure in BabyView/Ego4D may therefore be partly an artifact of frame-level pairing; video-native architectures might recover signal that our setup cannot see.

\paragraph{Backbone scale.} Our reference baselines use small backbones (DINOv2 ViT-B/14 vision tower; BERT-base text encoder for CLIP+, GPT-2 small for LLaVA). This is a deliberate choice to match the data scale of the naturalistic corpora and to remain within an academic compute budget, but it leaves open whether our negative result---current architectures collapse on weakly-aligned egocentric data---generalizes to frontier-scale models.

\textbf{LLM- and VLM-in-the-loop benchmark construction.}
Machine-DevBench's stimuli are generated and filtered by frontier models---captions by Gemma~4, images by FLUX.2, with PerceptionEncoder and Gemma~4 used to filter trials for image--caption alignment and depiction quality.
This couples our notion of ``linguistic competence'' to what these models find generable and discriminable, and could in principle inflate scores for models whose representations are aligned with the generators/filters or deflate scores on phenomena those models handle poorly.
We mitigate this with multi-stage filtering, two visual styles (photorealistic and cartoon), and calibration against the human-curated DevBench, with which Machine-DevBench scores correlate strongly across our 16 trained models (Fig.~\ref{fig:mdb_vs_devbench_scatter}).
A fully model-independent stimulus pipeline---e.g., human-authored or retrieval-only---remains an important complement.

We believe none of these limitations undermines the central finding that semantic alignment between modalities strongly predicts multimodal grounding performance, but each tightens the scope of that claim in a specific way.
We highlight them here so that readers calibrate the conclusions appropriately and so that future work knows where the load-bearing assumptions live.

\section{Per-task results}
\label{app:per-task-tables}

For completeness this appendix reports the full per-task numbers behind the
three subgroup-aggregate tables in the main text.
Tab.~\ref{tab:model_accuracy_full} expands the Machine-DevBench
\textbf{Lexical} / \textbf{Grammatical} / \textbf{Overall} aggregates
from Tab.~\ref{tab:model_accuracy} into all 10 tasks.
Tab.~\ref{tab:vision_comparison_full} expands the vision
\textbf{Object recognition} / \textbf{Visual properties} / \textbf{Overall}
aggregates from Tab.~\ref{tab:unimodal vision} into all 9 metrics.
Tab.~\ref{tab:text_comparison_full} expands the language
\textbf{Syntax} / \textbf{Semantics} / \textbf{Overall} aggregates from
Tab.~\ref{tab:unimodal text} into all 5 tasks.
The aggregation rules are the ones described in
App.~\ref{methods:image-aggregation} (vision),
App.~\ref{methods:text-aggregation} (language) and
App.~\ref{methods:mdb-aggregation} (Machine-DevBench).

\begin{table}[t]
    \caption{
        \textbf{(Appendix companion to Tab.~\ref{tab:model_accuracy})}.
        Per-task language grounding accuracy (\%) on Machine-DevBench, broken
        out across the two \textbf{Lexical} tasks and the eight
        \textbf{Grammatical} tasks. The aggregated subgroup and overall scores
        for each row are reported in the main-text Tab.~\ref{tab:model_accuracy}.
        The top section reports established baselines, while the bottom section
        shows models trained on the small-scale developmental datasets
        introduced in Section~\ref{q1}.
        Best two results for each category are shown in bold and underlined,
        respectively.
        For CLIP+ and LLaVA models we report the mean and std across 3
        training seeds.
    }
    \label{tab:model_accuracy_full}
    \centering
    \resizebox{1.0\textwidth}{!}{%
    \setlength{\tabcolsep}{3pt}
    \setlength{\columnsep}{3pt}
    \begin{NiceTabular}{l c c c c c c c c c c | c}
    \toprule
    \RowStyle{}
    & \multicolumn{2}{c}{\textbf{Lexical}}
    & \multicolumn{8}{c}{\textbf{Grammatical}}
    & \multicolumn{1}{c}{\textbf{}}
    \\
    \cmidrule(lr){2-3}
    \cmidrule(lr){4-11}
    \RowStyle{}
    \textbf{Model}
      & \textbf{Noun}
      & \textbf{Adj.}
      & \textbf{Subj-Verb}
      & \textbf{Subj-Adj}
      & \textbf{Neg.}
      & \textbf{Subj-Obj.}
      & \textbf{Prepos.}
      & \textbf{Compar.}
      & \textbf{Count}
      & \textbf{Relative}
      & \textbf{Aggr.}
      \\
    \midrule
    \rowcolor[gray]{0.95}
    \RowStyle{} Chance
      & 50.0 & 50.0 & 50.0 & 50.0 & 50.0 & 50.0 & 50.0 & 50.0 & 50.0 & 50.0 & 50.0 \\
    \RowStyle{} CLIP-L {\scriptsize\citep{radford2021learning}}
       & 94.6 & 80.0 & 63.2 & 76.9 & 75.7 & 61.7 & 75.4 & 55.9 & \underline{99.5} & 54.5 & 78.8 \\
    \RowStyle{} LLaVA-v1.6-Mistral-7B {\scriptsize{\citep{liu2023visual}}}
      & 92.8 & 87.3 & \textbf{99.3} & \textbf{99.4} & \textbf{98.6} & \textbf{90.6} & \textbf{99.0} & \textbf{91.2} & \textbf{100.0} & \textbf{96.7} & \textbf{93.4} \\
    \RowStyle{} PE-Core-B {\scriptsize\citep{bolya2025perception}}
       & \textbf{97.4} & \textbf{100.0} & 67.1 & \underline{92.9} & \underline{95.9} & \underline{77.8} & 86.2 & 52.0 & \underline{99.5} & 57.1 & \underline{88.6} \\
    \RowStyle{} Gemma 4 {\scriptsize\citep{gemma2024}}
      & \underline{94.9} & \underline{89.3} & \underline{71.1} & 82.2 & 94.6 & 69.4 & \underline{87.2} & \underline{57.8} & 79.2 & \underline{62.9} & 83.8 \\
    \midrule
    \rowcolor[gray]{0.95}
    \RowStyle{} BabyView-CLIP+
      & \score{52}{0}{0}{6}
      & \score{54}{8}{0}{8}
      & \score{54}{3}{3}{4}
      & \score{44}{7}{4}{6}
      & \score{50}{0}{1}{4}
      & \score{51}{2}{0}{4}
      & \score{50}{1}{2}{2}
      & \underline{\score{59}{7}{3}{2}}
      & \score{63}{5}{1}{5}
      & \textbf{\score{56}{8}{2}{1}}
      & \score{53}{6}{0}{7}
      \\
    \rowcolor[gray]{0.95}
    \RowStyle{} BabyView-LLaVA
      & \score{48}{7}{1}{0}
      & \score{53}{7}{4}{1}
      & \score{46}{9}{2}{1}
      & \score{51}{1}{3}{4}
      & \score{44}{6}{6}{8}
      & \score{54}{4}{4}{7}
      & \score{49}{2}{4}{9}
      & \score{48}{7}{6}{4}
      & \score{48}{0}{2}{8}
      & \underline{\score{54}{7}{3}{7}}
      & \score{50}{5}{1}{0}
      \\
    \RowStyle{} Ego4D-CLIP+
      & \score{49}{7}{1}{5}
      & \score{50}{7}{4}{1}
      & \score{48}{7}{4}{2}
      & \score{55}{8}{1}{8}
      & \score{46}{4}{5}{6}
      & \score{51}{6}{2}{5}
      & \score{61}{7}{3}{8}
      & \score{57}{2}{4}{7}
      & \score{43}{0}{3}{3}
      & \score{47}{4}{5}{6}
      & \score{50}{8}{0}{6}
      \\
    \RowStyle{} {Ego4D}-LLaVA
      & \score{49}{7}{1}{2}
      & \score{47}{8}{2}{6}
      & \score{52}{8}{1}{6}
      & \score{54}{9}{6}{6}
      & \score{43}{8}{4}{1}
      & \score{50}{3}{4}{1}
      & \score{49}{8}{1}{5}
      & \score{49}{2}{3}{4}
      & \score{50}{2}{7}{1}
      & \score{48}{2}{3}{8}
      & \score{49}{3}{1}{5}
      \\
    \rowcolor[gray]{0.95}
    \RowStyle{} HowTo-CLIP+
      & \score{59}{8}{2}{3}
      & \underline{\score{61}{4}{1}{9}}
      & \score{53}{8}{3}{9}
      & \score{43}{0}{3}{1}
      & \score{57}{2}{11}{5}
      & \score{54}{2}{2}{1}
      & \score{46}{6}{1}{7}
      & \score{47}{9}{0}{8}
      & \score{58}{9}{1}{6}
      & \score{50}{9}{3}{8}
      & \score{56}{1}{1}{2}
      \\
    \rowcolor[gray]{0.95}
    \RowStyle{} {HowTo}-LLaVA
      & \score{51}{6}{1}{2}
      & \score{53}{1}{1}{2}
      & \score{50}{3}{1}{9}
      & \score{55}{3}{2}{6}
      & \score{53}{9}{8}{6}
      & \score{47}{0}{2}{2}
      & \score{50}{5}{5}{7}
      & \score{49}{0}{3}{9}
      & \score{45}{8}{5}{7}
      & \score{53}{5}{0}{6}
      & \score{51}{5}{0}{5}
      \\
    \RowStyle{} {COCO-MC}-CLIP+
      & \textbf{\score{77}{5}{0}{9}}
      & \score{60}{2}{2}{7}
      & \score{55}{1}{1}{0}
      & \score{60}{8}{5}{3}
      & \underline{\score{59}{0}{7}{8}}
      & \textbf{\score{67}{6}{0}{7}}
      & \score{79}{3}{2}{0}
      & \score{54}{8}{4}{7}
      & \textbf{\score{96}{3}{0}{8}}
      & \score{40}{0}{1}{5}
      & \textbf{\score{66}{5}{1}{9}}
      \\
    \RowStyle{} {COCO-MC}\textsubscript{shuf0\%}-LLaVA
      & \score{56}{0}{2}{3}
      & \textbf{\score{63}{3}{2}{3}}
      & \textbf{\score{65}{6}{1}{3}}
      & \textbf{\score{65}{0}{2}{4}}
      & \textbf{\score{62}{6}{2}{0}}
      & \score{64}{5}{4}{7}
      & \score{70}{1}{2}{2}
      & \score{54}{1}{2}{7}
      & \score{81}{2}{1}{0}
      & \score{50}{3}{2}{3}
      & \score{61}{9}{0}{1}
      \\
    \rowcolor[gray]{0.95}
    \RowStyle{} {COCO-MC}\textsubscript{shuf25\%}-CLIP+
      & \underline{\score{76}{8}{0}{5}}
      & \score{58}{8}{1}{2}
      & \score{54}{0}{5}{3}
      & \underline{\score{62}{6}{0}{9}}
      & \score{54}{5}{4}{1}
      & \score{62}{8}{5}{9}
      & \textbf{\score{83}{3}{3}{2}}
      & \score{54}{6}{6}{0}
      & \underline{\score{95}{4}{1}{0}}
      & \score{44}{4}{2}{7}
      & \underline{\score{65}{9}{0}{7}}
      \\
    \rowcolor[gray]{0.95}
    \RowStyle{} {COCO-MC}\textsubscript{shuf25\%}-LLaVA
      & \score{55}{7}{1}{5}
      & \score{57}{6}{3}{6}
      & \underline{\score{59}{0}{4}{5}}
      & \score{57}{8}{2}{1}
      & \score{58}{2}{2}{9}
      & \score{57}{7}{4}{7}
      & \score{67}{3}{0}{9}
      & \score{48}{5}{2}{0}
      & \score{69}{8}{2}{6}
      & \score{51}{8}{3}{8}
      & \score{57}{7}{0}{6}
      \\
    \RowStyle{} {COCO-MC}\textsubscript{shuf50\%}-CLIP+
      & \score{76}{1}{0}{2}
      & \score{54}{5}{1}{3}
      & \score{54}{5}{2}{4}
      & \score{56}{4}{4}{2}
      & \score{52}{3}{3}{4}
      & \underline{\score{65}{5}{1}{7}}
      & \underline{\score{81}{1}{4}{8}}
      & \textbf{\score{60}{0}{3}{2}}
      & \score{93}{6}{0}{6}
      & \score{48}{0}{2}{0}
      & \score{64}{6}{0}{2}
      \\
    \RowStyle{} {COCO-MC}\textsubscript{shuf50\%}-LLaVA
      & \score{53}{0}{0}{5}
      & \score{56}{1}{1}{4}
      & \score{55}{3}{1}{1}
      & \score{53}{9}{3}{8}
      & \score{56}{2}{2}{3}
      & \score{55}{6}{1}{9}
      & \score{62}{4}{1}{1}
      & \score{50}{8}{3}{8}
      & \score{61}{0}{2}{9}
      & \score{52}{9}{0}{8}
      & \score{55}{3}{0}{6}
      \\
    \rowcolor[gray]{0.95}
    \RowStyle{} {COCO-MC}\textsubscript{shuf75\%}-CLIP+
      & \score{71}{5}{0}{4}
      & \score{57}{2}{3}{6}
      & \score{47}{6}{2}{2}
      & \score{55}{9}{3}{1}
      & \score{45}{0}{2}{1}
      & \score{61}{1}{5}{3}
      & \score{71}{7}{2}{4}
      & \score{50}{6}{8}{3}
      & \score{84}{8}{1}{2}
      & \score{48}{3}{6}{0}
      & \score{61}{2}{0}{6}
      \\
    \rowcolor[gray]{0.95}
    \RowStyle{} {COCO-MC}\textsubscript{shuf75\%}-LLaVA
      & \score{52}{5}{2}{0}
      & \score{48}{8}{1}{0}
      & \score{56}{3}{3}{7}
      & \score{48}{8}{5}{8}
      & \score{48}{5}{4}{6}
      & \score{49}{7}{4}{7}
      & \score{55}{9}{2}{3}
      & \score{43}{8}{5}{0}
      & \score{56}{2}{2}{7}
      & \score{49}{6}{0}{7}
      & \score{50}{9}{1}{0}
      \\
     \RowStyle{} {COCO-MC}\textsubscript{shuf100\%}-CLIP+
      & \score{49}{4}{1}{8}
      & \score{49}{1}{7}{7}
      & \score{50}{2}{1}{7}
      & \score{49}{5}{0}{8}
      & \score{45}{9}{10}{8}
      & \score{46}{6}{1}{6}
      & \score{54}{8}{5}{9}
      & \score{48}{3}{0}{5}
      & \score{45}{9}{7}{6}
      & \score{44}{8}{6}{6}
      & \score{48}{8}{1}{6}
      \\
    \RowStyle{} {COCO-MC}\textsubscript{shuf100\%}-LLaVA
      & \score{50}{2}{1}{8}
      & \score{47}{4}{1}{2}
      & \score{49}{5}{3}{6}
      & \score{49}{7}{3}{2}
      & \score{46}{1}{2}{1}
      & \score{52}{0}{3}{2}
      & \score{53}{9}{6}{0}
      & \score{51}{8}{0}{9}
      & \score{49}{4}{0}{7}
      & \score{47}{5}{2}{8}
      & \score{49}{4}{0}{7}
      \\

    \bottomrule
    \end{NiceTabular}
    }
\end{table}

\begin{table}[ht]
\centering
\caption{\textbf{(Appendix companion to Tab.~\ref{tab:unimodal vision})}. Per-task unimodal vision scores of the DINOv2 encoder with and without finetuning (FT) on multimodal data, for various datasets. The columns labelled ImageNet, MNIST and Seg correspond to our \textbf{Object recognition} subgroup, while Depth and Count correspond to our \textbf{Visual properties} subgroup. \textbf{Subgroup aggregates are not repeated here} -- see the main-text Tab.~\ref{tab:unimodal vision} for the \textbf{Object recognition Agg} and \textbf{Visual properties Agg} columns; the \textbf{Overall Agg} column is reproduced here for convenience. For BabyView/Ego4D/HowTo/COCO-MC, within each Train Data block the highest value of each column is bolded (ties broken by lower std; lower is better for Depth RMSE). For COCO-MC\textsubscript{shufX\%}, the higher between CLIP and LLaVA is bolded. We report the mean and std across 3 training seeds.}
\renewcommand{\arraystretch}{1.1}
\resizebox{\textwidth}{!}{%
\footnotesize
\newcolumntype{s}{>{\scriptsize}r}
\newcolumntype{t}{>{\scriptsize}l}
\begin{NiceTabular}{
    l
    c 
    r@{.}l@{\scriptsize$\,\pm\,$}s@{.}t 
    r@{.}l@{\scriptsize$\,\pm\,$}s@{.}t 
    r@{.}l@{\scriptsize$\,\pm\,$}s@{.}t 
    r@{.}l@{\scriptsize$\,\pm\,$}s@{.}t 
    r@{.}l@{\scriptsize$\,\pm\,$}s@{.}t 
    r@{.}l@{\scriptsize$\,\pm\,$}s@{.}t 
    r@{.}l@{\scriptsize$\,\pm\,$}s@{.}t 
    r@{.}l@{\scriptsize$\,\pm\,$}s@{.}t 
    r@{.}l@{\scriptsize$\,\pm\,$}s@{.}t 
    | r@{.}l@{\scriptsize$\,\pm\,$}s@{.}t 
}

\toprule
    \multirow[c]{2}{*}[-0.5ex]{\textbf{Train Data}}
    & \multirow[c]{2}{*}[-0.5ex]{\textbf{FT}}
    & \multicolumn{12}{c}{\textbf{ImageNet-1k}}
    & \multicolumn{4}{c}{\textbf{Depth}}
    & \multicolumn{4}{c}{\textbf{Seg}}
    & \multicolumn{8}{c}{\textbf{Count}}
    & \multicolumn{8}{c}{\textbf{MNIST}}
    & \multicolumn{4}{c}{\multirow[c]{2}{*}[-0.5ex]{\textbf{Overall Agg$\uparrow$}}}
\\
\cmidrule(lr){3-14} \cmidrule(lr){15-18} \cmidrule(lr){19-22} \cmidrule(lr){23-30} \cmidrule(lr){31-38}
    &
    & \multicolumn{4}{c}{kNN$\uparrow$}
    & \multicolumn{4}{c}{linear$\uparrow$}
    & \multicolumn{4}{c}{ABX$\uparrow$}
    & \multicolumn{4}{c}{RMSE$\downarrow$}
    & \multicolumn{4}{c}{mIoU$\uparrow$}
    & \multicolumn{4}{c}{linear$\uparrow$}
    & \multicolumn{4}{c}{ABX$\uparrow$}
    & \multicolumn{4}{c}{linear$\uparrow$}
    & \multicolumn{4}{c}{ABX$\uparrow$}
    & \multicolumn{4}{c}{}
\\
\midrule
\rowcolor[gray]{0.95}
Chance &
        & 0 & 1 & 0 & 0 & 0 & 1 & 0 & 0 & 50 & 0 & 0 & 0 & 2 & 000 & 0 & 000 & 0 & 3 & 0 & 0 & 11 & 1 & 0 & 0 & 50 & 0 & 0 & 0 & 10 & 0 & 0 & 0 & 50 & 0 & 0 & 0 & 2 & 8 & 0 & 0 \\
\RowStyle{} DINOv2-B/14 {\scriptsize\citep{oquab2023dinov2}} & --
        & 82 & 1 & 0 & 0 & 84 & 5 & 0 & 1 & 97 & 5 & 0 & 0 & 0 & 293 & 0 & 007 & 45 & 0 & 0 & 1 & 36 & 0 & 0 & 9 & 54 & 6 & 1 & 0 & 98 & 4 & 0 & 0 & 76 & 3 & 0 & 6 & 59 & 9 & 0 & 7 \\
\RowStyle{} CLIP-B/16 {\scriptsize\citep{radford2021learning}} & --
        & 74 & 6 & 0 & 0 & 80 & 6 & 0 & 1 & 96 & 6 & 0 & 0 & 0 & 414 & 0 & 004 & 39 & 9 & 0 & 1 & 46 & 2 & 2 & 0 & 55 & 9 & 0 & 6 & 98 & 8 & 0 & 0 & 83 & 0 & 0 & 2 & 60 & 6 & 0 & 6 \\
\midrule
\multirow{3}{*}{BabyView} & \RowStyle{} \xmark
        & \textbf{38} & \textbf{5} & 0 & 0 & \textbf{52} & \textbf{9} & 0 & 1 & \textbf{79} & \textbf{7} & 0 & 0 & 0 & 534 & 0 & 002 & \textbf{27} & \textbf{8} & 0 & 2 & \textbf{34} & \textbf{9} & 0 & 0 & \textbf{54} & \textbf{7} & 0 & 8 & 96 & 4 & 0 & 0 & 65 & 3 & 0 & 6 & \textbf{45} & \textbf{1} & 0 & 4 \\
& CLIP+
        & 30 & 6 & 0 & 2 & 47 & 5 & 0 & 2 & 76 & 2 & 0 & 2 & \textbf{0} & \textbf{526} & 0 & 005 & 24 & 0 & 0 & 1 & 32 & 7 & 0 & 6 & 53 & 8 & 0 & 1 & \textbf{97} & \textbf{8} & 0 & 1 & 76 & 4 & 0 & 8 & 44 & 4 & 0 & 4 \\
\RowStyle{} & LLaVA
        & {18}&{1} & {0}&{4} & {29}&{0} & {0}&{2} & {69}&{7} & {0}&{1} & {0}&{758} & {0}&{013} & {19}&{5} & {0}&{2} & {26}&{1} & {1}&{5} & {51}&{3} & {0}&{8} & {93}&{1} & {0}&{6} & {\textbf{80}}&{\textbf{0}} & {0}&{9} & {36}&{7} & {0}&{9} \\
\midrule
  \multirow{3}{*}{Ego4D} & \RowStyle{} \xmark
            & \textbf{40} & \textbf{8} & 0 & 0 & \textbf{54} & \textbf{1} & 0 & 0 & \textbf{79} & \textbf{8} & 0 & 0 & \textbf{0} & \textbf{514} & 0 & 001 & \textbf{27} & \textbf{4} & 0 & 0 & 29 & 1 & 0 & 0 & 54 & 2 & 0 & 3 & 97 & 2 & 0 & 0 & 69 & 5 & 0 & 7 & 45 & 1 & 0 & 2 \\
  \RowStyle{} & CLIP+
            & 40 & 7 & 0 & 0 & 53 & 9 & 0 & 1 & 79 & 1 & 0 & 1 & 0 & 518 & 0 & 005 & 27 & 3 & 0 & 0 & \textbf{29} & \textbf{9} & 0 & 0 & \textbf{54} & \textbf{4} & 0 & 2 & \textbf{97} & \textbf{3} & 0 & 0 & 70 & 0 & 0 & 4 & \textbf{45} & \textbf{2} & 0 & 2 \\
\RowStyle{} & LLaVA
        & {15}&{7} & {8}&{7} & {20}&{7} & {11}&{1} & {64}&{7} & {4}&{6} & {0}&{881} & {0}&{196} & {20}&{0} & {0}&{3} & {25}&{1} & {2}&{5} & {51}&{3} & {0}&{4} & {83}&{7} & {11}&{3} & {\textbf{71}}&{\textbf{5}} & {5}&{4} & {31}&{6} & {6}&{5} \\
\midrule
\multirow{3}{*}{HowTo} \RowStyle{} & \xmark
        & \textbf{41} & \textbf{5} & 0 & 0 & \textbf{56} & \textbf{0} & 0 & 0 & \textbf{79} & \textbf{9} & 0 & 0 & \textbf{0} & \textbf{477} & 0 & 001 & 29 & 4 & 0 & 2 & 32 & 0 & 0 & 0 & \textbf{54} & \textbf{7} & 0 & 5 & \textbf{97} & \textbf{7} & 0 & 0 & 73 & 0 & 0 & 5 & 47 & 1 & 0 & 3 \\
& CLIP+
\RowStyle{}    & \textbf{41} & \textbf{5} & 0 & 0 & \textbf{56} & \textbf{0} & 0 & 0 & \textbf{79} & \textbf{9} & 0 & 0 & 0 & 477 & 0 & 002 & \textbf{29} & \textbf{4} & 0 & 0 & \textbf{33} & \textbf{0} & 0 & 0 & \textbf{54} & \textbf{7} & 0 & 5 & \textbf{97} & \textbf{7} & 0 & 0 & 73 & 0 & 0 & 3 & \textbf{47} & \textbf{3} & 0 & 2 \\
\RowStyle{} & LLaVA
        & {23}&{8} & {5}&{0} & {31}&{8} & {5}&{9} & {72}&{1} & {2}&{2} & {0}&{706} & {0}&{054} & {19}&{0} & {0}&{4} & {28}&{4} & {2}&{1} & {52}&{7} & {1}&{0} & {95}&{0} & {1}&{2} & {\textbf{75}}&{\textbf{7}} & {1}&{6} & {38}&{5} & {2}&{8} \\
\midrule
\multirow{3}{*}{{COCO-MC}} & \RowStyle{} \xmark
        & \bfseries 62 & \bfseries 4 & 0 & 0 & \bfseries 71 & \bfseries 9 & 0 & 0 & 90 & 0 & 0 & 0 & \textbf{0} & \textbf{383} & 0 & 001 & \bfseries 38 & \bfseries 0 & 0 & 0 & 38 & 7 & 1 & 9 & 56 & 4 & 1 & 4 & 98 & 3 & 0 & 0 & 74 & 7 & 0 & 4 & 55 & 4 & 0 & 9 \\
& CLIP+
\RowStyle{}         & 62 & 0 & 0 & 0 & 71 & 5 & 0 & 0 & \textbf{90} & \textbf{2} & 0 & 0 & 0 & 390 & 0 & 001 & 37 & 4 & 0 & 0 & \bfseries 39 & \bfseries 4 & 1 & 8 & \textbf{56} & \textbf{8} & 0 & 8 & \textbf{98} & \textbf{4} & 0 & 0 & 74 & 5 & 0 & 0 & \textbf{55} & \textbf{5} & 0 & 4 \\
\RowStyle{} & LLaVA
        & {40}&{1} & {0}&{2} & {52}&{4} & {0}&{2} & {85}&{1} & {0}&{3} & {0}&{570} & {0}&{001} & 18 & 9 & 0 & 2 & {38}&{6} & {7}&{7} & {55}&{3} & {0}&{5} & {97}&{9} & {0}&{2} & {\textbf{83}}&{\textbf{7}} & {0}&{5} & 49&0 & 1&7 \\
\midrule
\multirow{2}{*}{{COCO-MC}\textsubscript{shuf25\%}} & \RowStyle{} CLIP+
        & \textbf{62} & \textbf{2} & 0 & 2 & \textbf{71} & \textbf{5} & 0 & 1 & \textbf{89} & \textbf{4} & 0 & 7 & \textbf{0} & \textbf{389} & 0 & 005 & \textbf{37} & \textbf{3} & 0 & 1 & \textbf{39} & \textbf{7} & 2 & 2 & \textbf{56} & \textbf{9} & 0 & 9 & \textbf{98} & \textbf{5} & 0 & 1 & 74 & 7 & 0 & 3 & \textbf{55} & \textbf{4} & 0 & 5 \\
\RowStyle{} & LLaVA
        & {35}&{0} & {0}&{5} & {46}&{2} & {0}&{3} & {81}&{7} & {0}&{6} & {0}&{626} & {0}&{001} & 18 & 5 & 0 & 2 & {30}&{7} & {1}&{0} & {54}&{9} & {0}&{4} & {97}&{4} & {0}&{1} & {\textbf{81}}&{\textbf{6}} & {1}&{0} & 45&2 & 0&7 \\
\multirow{2}{*}{{COCO-MC}\textsubscript{shuf50\%}} & \RowStyle{} CLIP+
        & \textbf{62} & \textbf{3} & 0 & 0 & \textbf{71} & \textbf{5} & 0 & 0 & \textbf{88} & \textbf{9} & 0 & 1 & \textbf{0} & \textbf{388} & 0 & 002 & \textbf{37} & \textbf{3} & 0 & 1 & \textbf{40} & \textbf{5} & 2 & 5 & \textbf{56} & \textbf{7} & 1 & 0 & \textbf{98} & \textbf{5} & 0 & 0 & 74 & 9 & 0 & 2 & \textbf{55} & \textbf{5} & 0 & 5 \\
\RowStyle{} & LLaVA
        & {38}&{5} & {0}&{6} & {48}&{7} & {0}&{3} & {81}&{0} & {0}&{3} & {0}&{608} & {0}&{002} & 18 & 6 & 0 & 1 & {37}&{6} & {2}&{6} & {54}&{1} & {0}&{8} & {96}&{8} & {0}&{3} & {\textbf{80}}&{\textbf{8}} & {0}&{5} & 46&4 & 1&0 \\
\multirow{2}{*}{{COCO-MC}\textsubscript{shuf75\%}} & \RowStyle{} CLIP+
        & \textbf{62} & \textbf{3} & 0 & 0 & \textbf{71} & \textbf{5} & 0 & 0 & \textbf{88} & \textbf{8} & 0 & 0 & \textbf{0} & \textbf{388} & 0 & 001 & \textbf{37} & \textbf{4} & 0 & 1 & \textbf{40} & \textbf{3} & 3 & 3 & \textbf{56} & \textbf{5} & 1 & 0 & \textbf{98} & \textbf{5} & 0 & 0 & \textbf{75} & \textbf{0} & 0 & 2 & \textbf{55} & \textbf{4} & 0 & 6 \\
\RowStyle{} & LLaVA
        & {1}&{3} & {0}&{5} & {4}&{5} & {1}&{7} & {54}&{3} & {1}&{0} & {1}&{066} & {0}&{025} & 12 & 9 & 2 & 6 & {19}&{5} & {4}&{5} & {51}&{3} & {1}&{3} & {43}&{8} & {4}&{5} & {55}&{6} & {2}&{9} & 18&3 & 2&8 \\
\multirow{2}{*}{{COCO-MC}\textsubscript{shuf100\%}} & \RowStyle{} CLIP+
        & \textbf{62} & \textbf{4} & 0 & 0 & \textbf{71} & \textbf{7} & 0 & 0 & \textbf{89} & \textbf{7} & 0 & 0 & \textbf{0} & \textbf{386} & 0 & 001 & \textbf{37} & \textbf{7} & 0 & 0 & \textbf{39} & \textbf{6} & 3 & 1 & \textbf{56} & \textbf{6} & 1 & 0 & \textbf{98} & \textbf{4} & 0 & 0 & \textbf{75} & \textbf{0} & 0 & 3 & \textbf{55} & \textbf{5} & 0 & 6 \\
\RowStyle{} & LLaVA
        & {1}&{2} & {0}&{2} & {3}&{4} & {0}&{7} & {55}&{0} & {0}&{8} & {1}&{062} & {0}&{015} & 13 & 5 & 2 & 2 & {19}&{8} & {4}&{0} & {50}&{1} & {0}&{8} & {47}&{4} & {1}&{7} & {56}&{2} & {5}&{6} & 18&5 & 2&5 \\
\bottomrule
\end{NiceTabular}%
}
\label{tab:vision_comparison_full}
\end{table}

\begin{table}[ht]
    \centering
    \caption{\textbf{(Appendix companion to Tab.~\ref{tab:unimodal text})}. Per-task unimodal language scores of the BERT and GPT encoders with and without finetuning (FT) on multimodal data, for various datasets. The Zorro, InflSwap and AgrSwap columns correspond to our \textbf{Syntax} subgroup; the WordSwap and VP-Swap columns correspond to our \textbf{Semantics} subgroup. \textbf{Subgroup aggregates are not repeated here} -- see the main-text Tab.~\ref{tab:unimodal text} for the \textbf{Syntax Agg} and \textbf{Semantics Agg} columns; the \textbf{Overall Agg} column is reproduced here for convenience. We report the mean and std across 3 training seeds. For BabyView/Ego4D/HowTo/COCO-MC, within each Train Data--Encoder sub-block, the highest value is bolded (ties broken by lower std). For COCO-MC\textsubscript{shufX\%}, the higher between CLIP and LLaVA is bolded.}
    \label{tab:text_comparison_full}
    \newcommand{\val}[2]{{\footnotesize#1}{\scriptsize$\,\pm\,$#2}}
    \resizebox{0.9\textwidth}{!}{%
            \footnotesize
            \begin{NiceTabular}{
                l
                c
                c
                c
                @{\hspace{10pt}}c@{\hspace{4pt}}
                c@{\hspace{4pt}}
                c
                c
                | c
            }[
                code-before = {
                    \rectanglecolor{gray!10}{3-1}{3-9}
                }
            ]
            \toprule
            \multirow[c]{2}{*}[-1ex]{\textbf{Train Data}}
                & \multirow[c]{2}{*}[-1ex]{\textbf{Model}}
                & \multirow[c]{2}{*}[-1ex]{\textbf{FT}}
                & \multirow[c]{2}{*}[-1ex]{\textbf{Zorro$\uparrow$}}
                & \multicolumn{3}{c}{\textbf{LT-Swap}}
                & \multirow[c]{2}{*}[-1ex]{\textbf{VP-Swap$\uparrow$}}
                & \multirow[c]{2}{*}[-1ex]{\textbf{Overall Agg$\uparrow$}}
            \\
            \cmidrule(lr){5-7}
                &
                &
                &
                & {WordSwap$\uparrow$}
                & {InflSwap$\uparrow$}
                & {AgrSwap$\uparrow$}
                &
                &
            \\
            \midrule
            Chance & &
            & \val{50.0}{0.0}
            & \val{50.0}{0.0}
            & \val{50.0}{0.0}
            & \val{50.0}{0.0}
            & \val{50.0}{0.0}
            & \val{50.0}{0.0} \\
            \midrule
            \multirow{4}{*}{BabyView} & BERT & \xmark
            & {\val{\bfseries 76.27}{0.26}}
            & {\val{\bfseries 79.14}{0.30}}
            & {\val{\bfseries 81.59}{0.10}}
            & {\val{\bfseries 60.84}{0.30}}
            & {\val{\bfseries 64.48}{0.61}}
            & {\val{\bfseries 72.4}{0.3}} \\
& BERT & CLIP+
            & {\val{75.16}{0.00}}
            & {\val{78.0}{0.00}}
            & {\val{79.3}{0.00}}
            & {\val{60.8}{0.00}}
            & {\val{64.1}{0.00}}
            & {\val{71.4}{0.0}} \\
\RowStyle{} & GPT-2 & \xmark
            & {\val{\bfseries 59.65}{1.21}}
            & {\val{\bfseries 58.52}{1.49}}
            & {\val{\bfseries 68.81}{1.22}}
            & {\val{\bfseries 56.62}{0.26}}
            & {\val{\bfseries 53.11}{1.20}}
            & {\val{\bfseries 58.8}{1.1}} \\
\RowStyle{} & GPT-2 & LLaVA
            & {\val{52.5}{5.8}}
            & {\val{54.2}{0.8}}
            & {\val{59.7}{0.7}}
            & {\val{55.6}{0.2}}
            & {\val{52.3}{0.5}}
            & {\val{54.6}{1.4}} \\
            \midrule
            \multirow{4}{*}{Ego4D} & BERT & \xmark
            & {\val{\bfseries 75.63}{0.23}}
            & {\val{\bfseries 82.27}{0.31}}
            & {\val{\bfseries 86.60}{0.46}}
            & {\val{\bfseries 68.10}{1.54}}
            & {\val{\bfseries 65.10}{0.79}}
            & {\val{\bfseries 75.2}{0.6}} \\
& BERT & CLIP+
            & {\val{74.13}{0.71}}
            & {\val{80.85}{0.07}}
            & {\val{83.60}{0.14}}
            & {\val{62.85}{2.76}}
            & {\val{64.00}{0.42}}
            & {\val{73.0}{0.7}} \\
\RowStyle{} & GPT-2 & \xmark
            & {\val{\bfseries 57.15}{0.53}}
            & {\val{\bfseries 60.21}{2.04}}
            & {\val{\bfseries 64.23}{0.25}}
            & {\val{50.89}{0.15}}
            & {\val{\bfseries 54.30}{0.90}}
            & {\val{\bfseries 57.3}{0.9}} \\
\RowStyle{} & GPT-2 & LLaVA
            & {\val{52.73}{4.89}}
            & {\val{57.37}{1.51}}
            & {\val{54.85}{2.32}}
            & {\val{\bfseries 51.16}{1.61}}
            & {\val{52.27}{0.63}}
            & {\val{53.9}{2.0}} \\
            \midrule
            \multirow{4}{*}{HowTo} & BERT & \xmark
            & {\val{\bfseries 72.31}{0.59}}
            & {\val{\bfseries 86.37}{1.27}}
            & {\val{\bfseries 89.43}{0.15}}
            & {\val{\bfseries 67.33}{1.46}}
            & {\val{\bfseries 72.53}{1.19}}
            & {\val{\bfseries 77.9}{1.0}} \\
& BERT & CLIP+
            & {\val{70.64}{0.74}}
            & {\val{78.80}{0.35}}
            & {\val{82.10}{0.40}}
            & {\val{62.77}{1.07}}
            & {\val{66.53}{0.25}}
            & {\val{72.3}{0.5}} \\
\RowStyle{} & GPT-2 & \xmark
            & {\val{\bfseries 57.40}{0.06}}
            & {\val{\bfseries 63.29}{1.66}}
            & {\val{\bfseries 62.18}{1.90}}
            & {\val{49.98}{0.20}}
            & {\val{\bfseries 56.11}{1.11}}
            & {\val{\bfseries 58.1}{1.1}} \\
\RowStyle{} & GPT-2 & LLaVA
            & {\val{49.76}{2.72}}
            & {\val{61.24}{0.78}}
            & {\val{57.69}{0.78}}
            & {\val{\bfseries 51.59}{1.18}}
            & {\val{53.58}{0.35}}
            & {\val{55.2}{1.1}} \\
            \midrule
            \multirow{4}{*}{{COCO-MC}} & BERT & \xmark
            & {\val{62.29}{1.44}}
            & {\val{\bfseries 88.37}{0.43}}
            & {\val{\bfseries 82.19}{0.27}}
            & {\val{\bfseries 68.86}{0.58}}
            & {\val{74.37}{0.75}}
            & {\val{76.2}{0.7}} \\
\RowStyle{} & BERT & CLIP+
            & \val{\bfseries 64.96}{0.28}
            & \val{88.17}{0.14}
            & \val{81.08}{0.66}
            & \val{68.49}{0.67}
            & \val{\bfseries 74.59}{0.26}
            & \val{\bfseries 76.4}{0.2} \\
\RowStyle{} & GPT-2 & \xmark
            & {\val{\bfseries 54.90}{3.41}}
            & {\val{\bfseries 80.44}{1.10}}
            & {\val{\bfseries 65.31}{1.23}}
            & {\val{54.92}{1.04}}
            & {\val{\bfseries 63.18}{0.55}}
            & {\val{\bfseries 65.1}{1.4}} \\
\RowStyle{} & GPT-2 & LLaVA
            & {\val{51.60}{0.89}}
            & {\val{65.53}{0.68}}
            & {\val{60.27}{0.22}}
            & {\val{\bfseries 55.62}{1.31}}
            & {\val{55.08}{0.39}}
            & {\val{58.1}{0.7}} \\
            \midrule
            \multirow{2}{*}{{COCO-MC}\textsubscript{shuf25\%}} & \RowStyle{} BERT & CLIP+
            & \val{\bfseries 65.09}{0.16}
            & \val{\bfseries 87.57}{0.45}
            & \val{\bfseries 80.67}{0.26}
            & \val{\bfseries 67.94}{0.34}
            & \val{\bfseries 74.29}{0.30}
            & \val{\bfseries 76.1}{0.2} \\
\RowStyle{} & GPT-2 & LLaVA
            & {\val{50.48}{3.91}}
            & {\val{67.95}{0.80}}
            & {\val{59.14}{0.91}}
            & {\val{54.18}{1.76}}
            & {\val{56.00}{0.62}}
            & {\val{58.3}{1.5}} \\
            \multirow{2}{*}{{COCO-MC}\textsubscript{shuf50\%}} & \RowStyle{} BERT & CLIP+
            & \val{\bfseries 64.79}{0.85}
            & \val{\bfseries 87.53}{0.26}
            & \val{\bfseries 80.37}{0.22}
            & \val{\bfseries 67.07}{0.13}
            & \val{\bfseries 73.47}{0.22}
            & \val{\bfseries 75.6}{0.2} \\
\RowStyle{} & GPT-2 & LLaVA
            & {\val{52.24}{1.64}}
            & {\val{69.97}{0.61}}
            & {\val{59.50}{0.51}}
            & {\val{53.51}{0.99}}
            & {\val{56.10}{0.14}}
            & {\val{59.1}{0.7}} \\
            \multirow{2}{*}{{COCO-MC}\textsubscript{shuf75\%}} & \RowStyle{} BERT & CLIP+
            & \val{\bfseries 63.80}{1.39}
            & \val{\bfseries 86.43}{0.59}
            & \val{\bfseries 79.80}{0.68}
            & \val{\bfseries 65.83}{1.56}
            & \val{\bfseries 72.60}{0.26}
            & \val{\bfseries 74.7}{0.4} \\
\RowStyle{} & GPT-2 & LLaVA
            & {\val{50.62}{3.82}}
            & {\val{64.39}{1.10}}
            & {\val{56.56}{0.51}}
            & {\val{54.55}{1.07}}
            & {\val{54.28}{0.67}}
            & {\val{56.6}{1.3}} \\
            \multirow{2}{*}{{COCO-MC}\textsubscript{shuf100\%}} & \RowStyle{} BERT & CLIP+
            & \val{\bfseries 63.91}{0.65}
            & \val{\bfseries 86.93}{0.18}
            & \val{\bfseries 80.87}{0.20}
            & \val{\bfseries 66.57}{0.36}
            & \val{\bfseries 73.30}{0.23}
            & \val{\bfseries 75.3}{0.1} \\
\RowStyle{} & GPT-2 & LLaVA
            & {\val{52.11}{4.61}}
            & {\val{62.92}{3.96}}
            & {\val{54.90}{1.84}}
            & {\val{53.67}{1.42}}
            & {\val{54.54}{1.26}}
            & {\val{56.1}{2.6}} \\
            \bottomrule
            \end{NiceTabular}
            }
\end{table}

\section{Methods}
In this section, we describe the methods we used to preprocess multimodal data, pretrain encoders, train multimodal models, and evaluate trained models.

\subsection{Data Processing}

\subsubsection{BabyView Voice Type Classification (VTC) Filtering}
To avoid training on the baby's own (often incoherent) speech output, we performed a filtering step based on Voice Type Classification (VTC). We used an off-the-shelf VTC model based on BabyHuBERT \citep{charlot-etal-2025-babyhubert},\footnote{\url{https://github.com/LAAC-LSCP/VTC}} to segment speakers into four classes (``adult female'', ``adult male'', ``key child'', and ``other child''), and discarded all segments labeled as ``key child.''

\subsubsection{WhisperX Transcription}
\label{method:whisperx}
\begin{figure}
    \centering
    \includegraphics[width=0.8\textwidth]{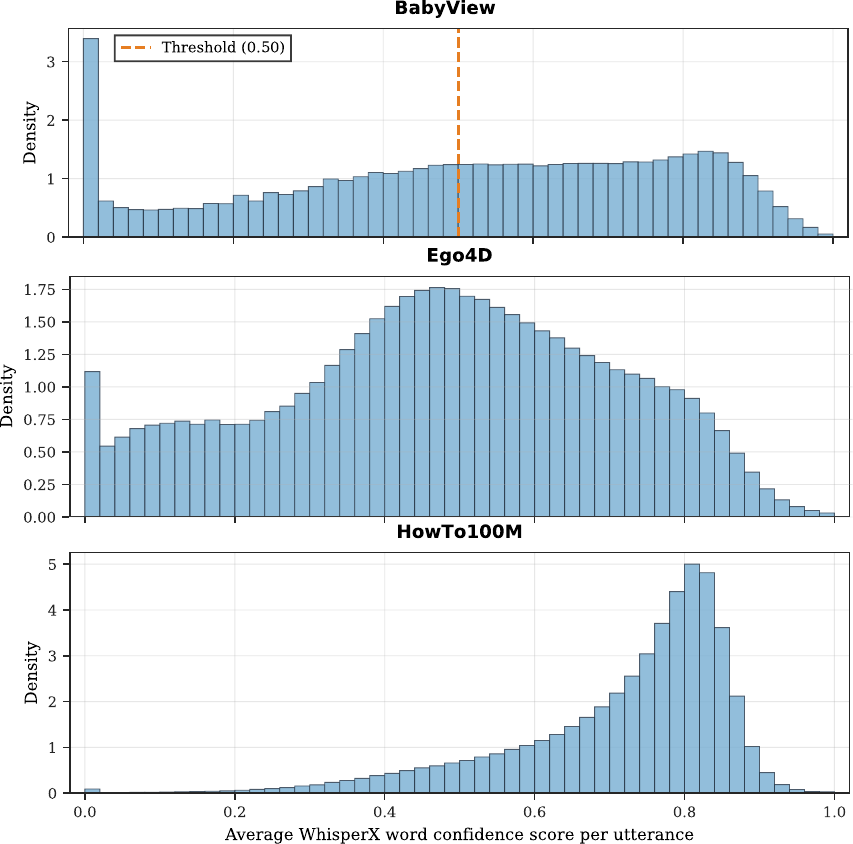}
    \caption{Density plots for the distribution of average WhisperX  word confidence score per utterance for each of the video datasets. Orange line indicates threshold used to filtering; only utterances with a higher average score than the threshold were preserved for the downstream BERT and contrastive training. Total number of utterances in the distribution noted in the upper right corner.}
    \label{fig:word_score_plots}
\end{figure}

For video datasets (BabyView, Ego4D, HowTo), we extracted transcriptions with word-level timestamps using WhisperX~\citep{bain2023whisperx}. To do this, we first extracted single-channel 16kHz audio using FFmpeg, then applied WhisperX with the \texttt{large-v2} model and batch size 1024. WhisperX provides forced alignment for precise word-level timestamps, speaker diarization, and per-word confidence scores enabling quality filtering. We used this confidence score to filter out hallucinated transcript: since different datasets have different data distributions, and hence different average WhisperX word confidence scores per utterance (see Fig.~\ref{fig:word_score_plots}). Egocentric datasets (BabyView, Ego4D) had significantly lower WhisperX score than instructional videos (HowTo), since speakers may speak more clearly in these instructional videos. For BERT / GPT-2 training we sampled utterances that in total make up to 1.12M words from these transcripts, keeping the ones with high WhisperX score and in-distribution sentence length. See Tab.~\ref{tab:dataset_sizes} for a summary of number of words and utterances obtained from WhisperX to train BERT / GPT-2.

\subsubsection{Video Frame and Clip Extraction}
\label{method:frame-extraction}
For DINOv2 training, we extracted \circa4M frames from each video dataset at 1 fps, ensuring diverse videos are included. For cross-modal training, following \citet{vong2024grounded}, we extracted video frames temporally aligned with transcriptions using WhisperX timestamps. For each utterance, we extracted up to 32 frames per utterance at 1 FPS. Frames were extracted using linear interpolation across the utterance timespan. All frames were resized to 256 pixels along the minor edge. See Tab.~\ref{tab:dataset_sizes} for a summary of number of frames obtained for each dataset.

\begin{table}[ht]
\centering
\small
\caption{Dataset sizes used for training: number of utterances/captions, number of images or frames used for DINOv2 vision pretraining, and number of frame-utterance (or image-caption) pairs used for contrastive (VLM) training.}
\renewcommand{\arraystretch}{1.2}
\begin{tabular}{lrrrr}
\hline
 & \textbf{BabyView} & \textbf{Ego4D} & \textbf{HowTo} & \textbf{COCO-MC} \\
\hline
Number of utterances or captions to train BERT/GPT-2 & 237K & 207K & 100K & 105K \\
Number of words to train BERT/GPT-2 & 1.12M & 1.12M & 1.12M & 1.12M \\
Number of images or frames used to train DINOv2 & \circa4M & \circa4M & \circa4M & \circa4M \\
Number of frame-utterance pairs to train CLIP+/LLaVA & 468K & 468K & 407K & 483K \\
\hline
\end{tabular}%
\label{tab:dataset_sizes}
\end{table}

For semantic alignment scoring with video encoders, we also extracted video clips temporally aligned with transcriptions using WhisperX timestamps. For each utterance, we extracted clips with a minimum duration of $0.3$ seconds (shorter clips were extended, if possible) and maximum duration of $60$ seconds (longer clips were truncated) in the video's native framerate and resolution using FFmpeg.




\subsubsection{COCO extended with MetaCLIP images (COCO-MC)}
\label{methods:coco-mc}
We mitigated the discrepancy between the larger number of video frames in BabyView, Ego4D, and HowTo (\circa4M for DINO and \circa500k for VLM training) and the comparably small number of images in the COCO training set (\circa110k) by retrieving additional images from the MetaCLIP~1.2 dataset \citep{xu2024demystifying, xu-etal-2024-altogether}; we refer to the resulting datasets as \textbf{COCO-MC}. Our goal was to scale up the image volume while (i) keeping text volume controlled and (ii) sampling images that are visually correlated with each caption, mirroring the strong visual correlation among video frames sampled for the same utterance.

To this end, for each caption in the Karpathy COCO split we retrieved up to five additional images using a two-stage pipeline. In the first (\emph{recall}) stage, we encoded each caption and a 1B-image subset of MetaCLIP with PE-Core-G-14-448~\citep{bolya2025perception} and kept the 30 images with the highest cosine similarity per caption. In the second (\emph{precision}) stage, we re-ranked the resulting candidate (caption, image) pairs with PerceptionLM-8B~\citep{cho2025perceptionlm} using VQAScore (see App.~\ref{app:semantic_alignment_further}). Finally, we greedily assigned images to captions in descending VQAScore order, accepting an image for a caption when (i) its score exceeded a threshold of $0.9$, (ii) it had not already been assigned to another caption, and (iii) the caption had fewer than five matches. This yielded \circa500{,}000 high-quality image–caption pairs which, combined with the original COCO images, produced \circa600k pairs available for VLM training; we then sampled randomly from these to match the video datasets' number of data pairs. Manual inspection confirmed that the pipeline reliably retrieved images accurately matching the original COCO captions.

To match the BabyView image budget for DINO training, we further sampled \circa3.4M additional uncaptioned MetaCLIP images uniformly at random from the same 1B-image subset (excluding the already-paired images), yielding 4M images in total.

\subsubsection{COCO Captions shuffling procedure}
\label{methods:coco-shuffling}
To study the effects of semantic misalignment on contrastive learning, we generated controlled permutations of COCO-MC image-caption pairs using a derangement approach, ensuring no image retained its original caption within the permuted subset. Permutation metadata was generated with specified percentages (e.g., 25\%, 50\%, 75\%, 100\%) of training pairs from the Karpathy splits  \citep{Karpathy_2015_CVPR}, with fixed random seeds for reproducibility.

\subsection{Semantic Alignment Scoring}
\label{methods:semantic-alignment}
We computed alignment scores as described in Section~\ref{q1} via the Jensen-Shannon divergence between cosine similarity (CLIPScore) distributions with bootstrap confidence intervals ($n=1000$). Below, we detail differences in encoding inputs with our two primary scoring models, ViT-B~CLIP and Perception Encoder.

\paragraph{CLIP} We used the off-the-shelf pretrained ViT-B/16 from \citet{radford2021learning}. We extracted image features for COCO-MC and frame-level visual features for video datasets (Ego4D, HowTo, BabyView). For video datasets, we uniformly sampled $N = 8$ frames from each video clip (extracted as outlined in Section~\ref{method:frame-extraction} above). The sampled frames were paired with the full utterance temporally corresponding to their source video clip. As a result, we obtained several cosine similarity scores for every clip-utterance segment.
When shuffling utterances to compute our alignment metric, we sampled transcriptions from different video clips.

\paragraph{Perception Encoder} We used the off-the-shelf pretrained PE-Core-G-14-448. Following \citet{bolya2025perception}, we embedded a video clip (extracted as outlined in Section~\ref{method:frame-extraction} above), by first uniformly sampling $N = 8$ frames, extracting frame-level embeddings with the Perception Encoder, and then mean-pooling over these frame embeddings to obtain video embeddings. For COCO-MC, we directly encoded the images and used the resulting embeddings to compute cosine similarities. We applied the default image transforms for PE-Core-G-14-448, including resizing to $448$px resolution and normalizing with a mean and std of $0.5$.

\subsection{Models and Training}

\subsubsection{DINOv2 Training on Video and Image Datasets}
\label{methods:dinov2-training}

We trained ViT-B/14 backbones with combined DINO and iBOT losses plus KoLeo regularization,
$\mathcal{L} = \mathcal{L}_{\text{DINO}} + \mathcal{L}_{\text{iBOT}} + 0.1 \cdot \mathcal{L}_{\text{KoLeo}}$,
using separate projection heads for DINO and iBOT and Sinkhorn--Knopp centering of the teacher DINO outputs.
Optimization uses AdamW \citep{kingma-ba-2015-adam, loshchilov2018decoupled} ($\beta_1{=}0.9$, $\beta_2{=}0.999$)
with cosine LR decay, linear warmup, square-root LR scaling w.r.t.\ a reference batch size of 1024,
EMA teacher updates, gradient clipping, and FSDP mixed precision (FP16) across 4 nodes
with FP32 reductions in the DINO/iBOT projection heads for stability.
We coarsely grid-searched the base learning rate per dataset between \num{5e-5} and \num{2e-4}; all other hyperparameters are shared
and summarized in Tab.~\ref{tab:dinov2_hyperparams}.
Remaining settings follow the DINOv2 defaults \citep{oquab2023dinov2}.

\begin{table}[ht]
  \centering
  \caption{DINOv2 ViT-B/14 training hyperparameters. The base learning rate is the only setting
  that varied across datasets; all other values are shared.}
  \small
  \label{tab:dinov2_hyperparams}
  \begin{tabular}{lcccc}
  \toprule
  \textbf{Parameter} & \textbf{BabyView} & \textbf{Ego4D} & \textbf{HowTo} & \textbf{COCO-MC} \\
  \midrule
  Base learning rate & \num{8e-5} & \num{8e-5} & \num{7.5e-5} & \num{5e-5} \\
  \midrule
  Epochs                                  & \multicolumn{4}{c}{500} \\
  Batch size per GPU                      & \multicolumn{4}{c}{64} \\
  LR warmup epochs                        & \multicolumn{4}{c}{80} \\
  LR scaling rule                         & \multicolumn{4}{c}{$\sqrt{\,\text{batch}/1024\,}$} \\
  Min.\ LR (cosine end)                   & \multicolumn{4}{c}{\num{1e-6}} \\
  Weight decay (start$\rightarrow$end)    & \multicolumn{4}{c}{0.04 $\rightarrow$ 0.2} \\
  Layerwise LR decay                      & \multicolumn{4}{c}{1.0} \\
  Gradient clip (max norm)                & \multicolumn{4}{c}{3.0} \\
  Drop path rate                          & \multicolumn{4}{c}{0.3} \\
  Teacher momentum (start$\rightarrow$end)& \multicolumn{4}{c}{0.994 $\rightarrow$ 1.0} \\
  Teacher temp.\ (warmup$\rightarrow$final)& \multicolumn{4}{c}{0.04 $\rightarrow$ 0.07} \\
  Teacher temp.\ warmup epochs            & \multicolumn{4}{c}{30} \\
  DINO / iBOT loss weight                 & \multicolumn{4}{c}{1.0 / 1.0} \\
  KoLeo loss weight                       & \multicolumn{4}{c}{0.1} \\
  DINO / iBOT prototypes                  & \multicolumn{4}{c}{131{,}072 / 131{,}072} \\
  Separate DINO / iBOT heads              & \multicolumn{4}{c}{\cmark} \\
  Teacher centering                       & \multicolumn{4}{c}{Sinkhorn--Knopp} \\
  Global / local crop size (px)           & \multicolumn{4}{c}{224 / 98} \\
  Local crops per image                   & \multicolumn{4}{c}{8} \\
  iBOT mask ratio (min, max)              & \multicolumn{4}{c}{(0.1,\,0.5)} \\
  iBOT mask sample probability            & \multicolumn{4}{c}{0.5} \\
  Register tokens                         & \multicolumn{4}{c}{0} \\
  Mixed precision                         & \multicolumn{4}{c}{FP16 (FP32 reduce in DINO/iBOT heads)} \\
  \bottomrule
  \end{tabular}
  \end{table}

\subsubsection{BERT Training on Multimodal Dataset Text}
\label{methods:bert-training}


For each multimodal dataset (COCO captions, BabyView, Ego4D, and HowTo transcripts) we trained a BERT-base model \citep{devlin2019bert} from scratch with a dedicated WordPiece tokenizer fit on the same corpus. Training uses masked language modeling, \[
  \mathcal{L}_{\text{MLM}} = -\frac{1}{|\mathcal{M}|}\sum_{i \in \mathcal{M}} \log p(x_i \mid x_{\backslash \mathcal{M}}),
  \] where $\mathcal{M}$ is the set of masked positions and $x_{\backslash \mathcal{M}}$ the unmasked context. Models are evaluated on held-out text via cross-entropy loss and word-level masked prediction accuracy. We swept epochs ($\{10,\ldots,300\}$), dropout ($\{0.1, 0.3\}$),
  and weight decay ($\{0.1, 0.3\}$) and selected the best configuration by held-out loss; the resulting shared setup is summarized in Tab.~\ref{tab:bert_hyperparams}.

\begin{table}[ht]
\centering
\caption{BERT-base MLM training hyperparameters, shared across all four datasets.}
\label{tab:bert_hyperparams}
\small
\begin{tabular}{ll}
\toprule
\textbf{Parameter} & \textbf{Value} \\
\midrule
Architecture          & BERT-base (12L, 768d, 12H) \\
Tokenizer             & WordPiece, vocab size 30{,}522 (per-dataset) \\
Mask probability      & 0.15 \\
Optimizer             & AdamW ($\beta_1{=}0.9$, $\beta_2{=}0.999$) \\
Learning rate         & \num{1e-4} \\
LR schedule           & Linear with warmup \\
Weight decay          & 0.1 \\
Dropout               & 0.3 \\
Batch size (per GPU)  & 128 \\
GPUs                  & 4 (single node) \\
Max epochs            & 300 (best by eval loss) \\
\bottomrule
\end{tabular}
\end{table}

\subsubsection{Multimodal Contrastive Training}
\label{sec:multimodal_alignment}
Following \citet{vong2024grounded}, we aligned visual and linguistic representations through contrastive learning on naturally co-occurring image-text pairs.

\paragraph{Data Preparation}
For COCO-MC, we constructed image-caption pairs as described in App.~\ref{methods:coco-mc}, with captions from the Karpathy splits \citep{Karpathy_2015_CVPR}. 
For video datasets, we paired utterances transcribed with WhisperX with co-occurring frames. Since each utterance may span multiple frames, during training we randomly sampled one of these frames as the matching image frame associated with a given utterance. We applied data augmentation consisting of random Gaussian blur and random horizontal flip to images during training.

\paragraph{Encoders}
To compare the efficacy of contrastive finetuning on language grounding, our framework supports multiple encoder architectures. For text encoding, we compared two configurations: (1) randomly initialized embedding layers that learn representations from scratch and (2) BERT models pretrained on target domain data (COCO captions, BabyView transcripts, Ego4D transcripts, or HowTo transcripts). The final text representations are obtained by applying a linear projection layer to map BERT's 768-dimensional \texttt{\texttt{[cls]}} token embeddings to the target embedding dimension of 512. For vision encoding, we similarly compare two configurations: (1) randomly initialized Vision Transformer (ViT) models and (2) DINOv2 models pretrained on target domain images or video frames. All visual encoders output 512-dimensional feature representations through a linear projection from their native dimensionality.

\paragraph{Contrastive Learning Objective}
  We optimize the multimodal alignment using the InfoNCE contrastive loss
  \citep{oord2018representation}, following the CLIP framework \citep{radford2021learning}.
  For a batch of $N$ image--text pairs $\{(x_i, y_i)\}_{i=1}^{N}$, the encoders described
  above produce image embeddings $\mathbf{f}_i = f_{\text{img}}(x_i)$ and text embeddings
  $\mathbf{g}_i = f_{\text{text}}(y_i)$, both in $\mathbb{R}^{512}$. The scaled similarity
  matrix is $S_{ij} = \tau \cdot \mathbf{f}_i^{\top}\mathbf{g}_j$, where $\tau$ is a
  learnable temperature parameter initialized to $0.07$. The symmetric InfoNCE loss
  combines image-to-text and text-to-image classification objectives:
  \begin{equation}
    \mathcal{L} = \tfrac{1}{2}\bigl(\mathcal{L}_{\mathrm{I2T}} + \mathcal{L}_{\mathrm{T2I}}\bigr),
    \quad
    \mathcal{L}_{\mathrm{I2T}} = -\tfrac{1}{N}\sum_{i=1}^{N} \log \tfrac{\exp(S_{ii})}{\sum_{j=1}^{N} \exp(S_{ij})},
    \quad
    \mathcal{L}_{\mathrm{T2I}} = -\tfrac{1}{N}\sum_{i=1}^{N} \log \tfrac{\exp(S_{ii})}{\sum_{j=1}^{N} \exp(S_{ji})}.
  \end{equation}

  \subsubsection{Interleaved Training for Catastrophic Forgetting Prevention}
  \label{methods:interleaved-training}
  To prevent catastrophic forgetting during contrastive finetuning, we implemented
  interleaved training that alternates the contrastive objective with the unimodal
  self-supervised objectives that produced the encoders. We considered three variants
  (Tab.~\ref{tab:interleaved_variants}). BERT-CLIP inserts one MLM step every $N$
  contrastive steps; DINOv2-CLIP inserts $K$ DINOv2 steps every 100 contrastive steps;
  and CLIP+ chains both, performing $Q$ MLM steps and then $R$ DINOv2 steps every 100
  contrastive steps. During MLM steps, only the BERT text encoder and the MLM
  classification head are updated. During DINOv2 steps, only the vision encoder is
  updated, on the current multimodal batch re-augmented with the DINOv2 augmentation
  pipeline, and the CLIP vision encoder is re-synchronized to the DINOv2 teacher
  backbone after each phase. In CLIP+, shared text-encoder weights remain consistent
  across contrastive and MLM phases through gradient routing. For each variant we
  performed a grid search over the step ratios listed in Tab.~\ref{tab:interleaved_variants}
  and reported results for the configuration with the highest held-out retrieval
  accuracy measured after contrastive steps. Section~\ref{section:ablation} reports
  ablations on the CLIP+ regime.

  \paragraph{Optimization and Evaluation}
  All variants, including the plain CLIP baseline, share the optimization setup in
  Tab.~\ref{tab:cl_hyperparams}. Each interleaved objective maintains its own optimizer
  to allow independent learning-rate scheduling and gradient clipping. Models were
  evaluated using retrieval accuracy and entropy, and checkpoints with the lowest
  validation loss are used for downstream evaluations.

  \begin{table}[ht]
  \centering
  \caption{Shared training hyperparameters for contrastive and interleaved runs.}
  \label{tab:cl_hyperparams}
  \small
  \begin{tabular}{ll}
  \toprule
  \textbf{Parameter} & \textbf{Value} \\
  \midrule
  Optimizer              & AdamW \\
  Learning rate          & \num{1e-5} \\
  Weight decay           & 0.01 \\
  LR schedule            & ReduceLROnPlateau (patience 20) \\
  Max epochs             & 30 \\
  Batch size             & 128 \\
  Initial temperature    & 0.07 (learnable) \\
  Checkpoint selection   & lowest validation loss \\
  \bottomrule
  \end{tabular}
  \end{table}

\begin{table}[ht]
  \centering
  \footnotesize
  \setlength{\tabcolsep}{4pt}
  \caption{Interleaving schedules. Each row lists the auxiliary objectives and the
  searched and selected step ratios relative to contrastive (CL) steps.}
  \label{tab:interleaved_variants}
  \begin{tabular}{@{}l l l l l@{}}
  \toprule
  \textbf{Variant} & \textbf{Aux.\ obj.} & \textbf{Schedule} & \textbf{Grid} & \textbf{Selected} \\
  \midrule
  BERT-CLIP      & MLM              & 1 MLM / $N$ CL          & $N \in \{2,5,10,20\}$                  & $N{=}20$ \\
  DINOv2-CLIP    & DINO+iBOT        & $K$ DINO / 100 CL       & $K \in \{10,20\}$                      & $K{=}10$ \\
  CLIP+ (triple) & MLM, DINO+iBOT   & $Q$ MLM, $R$ DINO / 100 CL & $Q \in \{5,10,20\},\ R \in \{10,20\}$ & $Q{=}20,\ R{=}10$ \\
  \bottomrule
  \end{tabular}
  \end{table}

\subsubsection{LLaVA Training and Evaluation}
\label{methods:llava-training}

\paragraph{Architecture}
We adopt a BabyLLaVA-style~\citep{wang2025babyvlm, liu2023visual} architecture comprising three components: a GPT-2 Small language model (12 layers, 768-dimensional hidden size, 12 attention heads), a DINOv2 ViT-B/14 vision encoder, and a 2-layer MLP projector ($768 \rightarrow 256 \rightarrow 256$) mapping visual tokens to the language model hidden dimension. The GPT-2 language model is trained from scratch rather than initialized from pretrained weights, ensuring that all linguistic knowledge is acquired solely from the training data.

\paragraph{Data Preparation}
We use the same datasets and frame extraction pipeline as CLIP+ (BabyView, Ego4D, HowTo, COCO-MC with shuffled variants; see Section~\ref{sec:multimodal_alignment}). For text data, captions are prefixed with ``Describe this image.'' to standardize the prompt format, matching the evaluation prompt exactly. For multimodal data, image-caption pairs are formatted in LLaVA conversation style: the user message contains \texttt{<image>\textbackslash nDescribe this image.} and the assistant response contains the caption.

\paragraph{Three-Phase Training}
Training proceeds in three phases. In \textbf{Phase~0}, the GPT-2 Small language model is trained from scratch on the dataset text using a causal language modeling objective. A custom BPE tokenizer is retrained on the corpus. Text is prefixed with ``Describe this image.'' to match the multimodal training format. We use lr=$\num{1e-4}$, batch size 64, 30 epochs, AdamW optimizer, and dropout 0.1. In \textbf{Phase~1}, the MLP projector is pretrained with both the vision tower and language model frozen, using lr=$\num{3e-3}$, batch size 32, 5 epochs, cosine schedule with warmup ratio 0.03, fp16 precision, and DeepSpeed ZeRO-2. In \textbf{Phase~2}, the entire model is finetuned, using lr=$\num{2e-3}$, batch size 32, 5 epochs, cosine schedule with warmup ratio 0.03, fp16 precision, and DeepSpeed ZeRO-2. 

\paragraph{Text Evaluation}
The GPT-2 checkpoint from Phase~0 and the LLaVA checkpoint from Phase~2 are both evaluated on Zorro and LongTail-Swap. Scoring uses total sentence log-probability; a model receives credit when $P(\text{grammatical}) > P(\text{ungrammatical})$.

\subsection{Evaluation Metrics}

  \subsubsection{Vision Encoder Evaluation}
  \label{methods:vision-evals}
  We evaluated the vision encoder on a frozen-backbone benchmark spanning object recognition,
  fine-grained discrimination, and dense prediction. Unless otherwise noted, classification
  probes are fit on \texttt{[cls]} features from the last transformer block; linear probes
  use SGD with momentum $0.9$, weight decay $0$, and cosine annealing, sweeping the same
  16-point learning-rate grid from \num{1e-5} to $0.5$ and reporting the best validation
  accuracy. ABX evaluations use the triplet protocol of~\citet{poli2025fastabx}: for triplets
  $(X, A, B)$ with $X$ and $A$ sharing a label and $B$ differing, the probe is correct when
  $d(X, A) < d(X, B)$ under cosine distance,
  $d(u, v) = 1 - \tfrac{u^{\top}v}{\|u\|_2\|v\|_2}$, with 100 triplets per class pair.

  \paragraph{ImageNet-1k}
  We measured image recognition on ImageNet-1k~\citep{deng2009imagenet} in three protocols.
  \textbf{k-NN classification} extracts L2-normalized features and assigns labels by
  temperature-scaled softmax voting over cosine-similar neighbors ($T{=}0.07$); we sweep $k\in{10,20,100,200}$ and report top-1 accuracy at $k{=}20$ on the official validation split, consistent with the DINOv2 protocol~\citep{oquab2023dinov2}.
  \textbf{Linear classification} additionally sweeps four pooling strategies that vary
  how many of the last transformer blocks are concatenated (1 or 4) and whether average-pooled
  patch tokens are appended to the \texttt{[cls]} token, trained for 10 epochs.
  \textbf{ABX classification} samples triplets on the validation split, capped at 50{,}000
  images.

  \paragraph{MNIST}
  We evaluated digit recognition with the same three protocols on MNIST. For k-NN we sweep $k\in{3,5,10,20}$ and likewise report $k{=}20$ on the test split. The linear probe trains for 20 epochs at batch size
  $256$. ABX samples triplets across pairs of digits.

  \paragraph{CountBench}
  \label{methods:countbench}
  We evaluated object counting on CountBench~\citep{paiss2023teaching} by treating object
  counts as classification labels. The linear probe trains on the official train split for 50
  epochs at batch size $64$ and is evaluated on the held-out split. ABX samples triplets across
  pairs of counts.

  \paragraph{Depth Estimation}
  We evaluated dense depth on NYU Depth v2~\citep{silberman2012indoor} using a Dense Prediction
  Transformer (DPT) head~\citep{ranftl2021vision} on the frozen backbone. DPT taps features
  from four equally-spaced depths of the encoder (25\%, 50\%, 75\%, 100\%), reassembles each
  to a 256-channel pyramid through 3$\times$3 convolutions, fuses the levels into a 128-channel
  representation, and predicts depth via a 1$\times$1 convolution with Softplus activation.
  The head is trained with SigLoss~\citep{eigen2014depth},
  \begin{equation}
  \mathcal{L}_{\text{Sig}} = \tfrac{1}{n}\sum_i d_i^2 - \tfrac{\lambda}{n^2}\Big(\sum_i d_i\Big)^2,
  \end{equation}
  where $d_i = \log\hat{y}_i - \log y_i$ and $\lambda{=}0.5$, with depths clamped to
  $[0.001, 10]\,\text{m}$. We optimized with AdamW at peak learning rate \num{3e-4} under a
  OneCycleLR schedule for 25 epochs at batch size 16 and report RMSE on the test split with
  horizontal-flip test-time augmentation.

  \paragraph{Semantic Segmentation}
  We evaluated dense recognition on COCO-Stuff~\citep{caesar2018cocostuff} (171 classes:
  80 things plus 91 stuff) with a linear segmentation head on the frozen backbone. The head is
  a single $1{\times}1$ convolution mapping backbone features (concatenating either the last
  block or the last 4 blocks) to per-class logits at the patch grid, followed by bilinear
  upsampling to the input. We trained at $448 \times 448$ with cross-entropy (ignoring void
  pixels) using SGD (lr $0.01$, momentum $0.9$, weight decay \num{1e-4}) for 25 epochs with
  step decay $\gamma{=}0.1$ at epoch 20, and report the best validation mIoU across the two
  pooling strategies.

  \paragraph{Aggregated Vision Score}
  \label{methods:image-aggregation}
  We score each task on a chance-corrected $[0, 100]$ scale and group the nine resulting
  scores into two subgroups (Sec.~\ref{babyvlm-challenge}): an \textbf{Object recognition}
  subgroup (ImageNet-1k k-NN, linear, and ABX; MNIST linear and ABX; COCO-Stuff segmentation;
  six metrics) and a \textbf{Visual properties} subgroup (NYU Depth v2; CountBench linear and
  ABX; three metrics). Each metric is normalized as $\text{Score}=100\cdot\text{acc}$ for the
  classification probes, $\text{Score}=100\cdot(\text{acc}-0.5)/0.5$ for ABX,
  $\text{Score}=100\cdot(1-\text{RMSE}/2.0)$ for depth, and
  $\text{Score}=100\cdot(\text{mIoU}-1/(2K-1))/(1-1/(2K-1))$ for segmentation
  ($K{=}171$, chance mIoU $\approx 0.0029$); all scores are clamped to $[0,100]$. The subgroup
  aggregate $A_g$ is the arithmetic mean of its task scores, and the \textbf{Overall} vision
  score is $A_{\text{overall}} = \tfrac{1}{2}(A_{\text{Object}} + A_{\text{Properties}})$.

\subsubsection{Text Encoder Evaluation}
\label{methods:text-evals}

\paragraph{Zorro Benchmark}
Zorro~\citep{warstadt2020blimp} comprises minimal pairs testing 13 grammatical phenomena (1,000 pairs each): anaphor agreement, argument structure, binding, determiner-noun agreement, ellipsis, filler-gap dependencies, irregular forms, island effects, NPI licensing, quantifiers, subject-verb agreement, case pronouns, local attractors.

For vanilla BERT, the models were loaded via HuggingFace \texttt{hf-mlm} and evaluated zero-shot. For each minimal pair, we compute the model's probability assignment to both the grammatical and ungrammatical sentences. A model receives credit for a minimal pair if it assigns higher probability to the grammatical sentence than to its ungrammatical counterpart.

For BERT that was trained as part of the multimodal model, since the original MLM head was replaced during contrastive training, we first extracted the BERT encoder weights from multimodal checkpoint, then retrained the MLM head for 30 epochs with the same textual data and hyperparameters as the original BERT training (lr \num{5e-5}, batch size 128, MLM prob=0.15), then applied the same Zorro evaluation procedure.

\paragraph{LongTail-Swap Evaluation}
\label{methods:longtail-swap}
We evaluate our models on the LongTail-Swap benchmark~\citep{algayres2025longtail}, designed to assess language models' capabilities on rare words and long-tail linguistic phenomena. The benchmark comprises three tasks: \textit{WordSwap}, which tests lexical semantics by swapping rare words between two distinct sentences; \textit{InflectionSwap}, which targets morphological understanding by swapping inflected forms of the same rare word (e.g., past vs. present tense); and \textit{AgreementSwap}, which evaluates syntactic agreement (e.g., subject-verb, determiner-noun) involving rare words. Target words are sampled from the pretraining corpus and stratified into frequency bins. For each target word, minimal sentence pairs ($S_1, S_2$) are constructed by prompting a large language model (LLM), here we used Llama-3.1-405B-Instruct \citep{dubey2024llama}, to generate natural sentences containing the word. These generations undergo a LLM-based filtering process to ensure grammatical correctness and to verify that swapping the target words (or inflections) between ($S_1, S_2$) where the two words are from the same frequency bin yields an invalid or semantically implausible sentence. Once the swapped sentence pairs are constructed, we perform zero-shot evaluation using a minimal pair classification accuracy metric. For each instance, we compute the model's probability score assignment for the correct sentences $S_1$, $S_2$ and the perturbed sentences $S_1'$, $S_2'$. A model receives credit for a minimal pair if it assigns higher probability to the correct sentence than to its perturbed counterpart (i.e., $P(S_1) > P(S_1')$ and $P(S_2) > P(S_2')$). We report the accuracy averaged across frequency bins and specific linguistic phenomena.

\paragraph{Visual Property Swap (VP-Swap) Evaluation}
\label{methods:vp-swap}
To assess the model's grounding of language in visual perception, we introduce a novel Visual Property Swap (VP-Swap) benchmark. This task extends the LongTail-Swap methodology to evaluate the understanding of visual attributes of concrete nouns, specifically focusing on four properties: Color, Material, Relative Size, and Shape. The pipeline begins by extracting potential concrete nouns from the pretraining corpus using Part-of-Speech tagging. We refine this list by prompting an LLM (Llama 3.1 405B) to verify that each word represents a physical object. Valid nouns are then stratified into frequency bins. For each visual property, we sample pairs of nouns from the same frequency bin and prompt an LLM to generate a pair of correct sentences ($S_1, S_2$) describing the visual property for each noun (e.g., $S_1$: ``The banana is yellow'', $S_2$: ``The television is black''). We then swap the nouns to create invalid or factually incorrect counterparts ($S_1'$: ``The television is yellow'', $S_2'$: ``The banana is black''). A final LLM-based feasibility filtering step ensures that the swapped sentences are distinguishable from the correct ones (e.g., rejecting pairs where swapping the noun does not result in a visual contradiction). We employ the same zero-shot evaluation protocol as in LongTail-Swap. For each quadruplet ($S_1, S_1', S_2, S_2'$), we compare the model's likelihood scores. A model is considered correct on a pair if it assigns a higher probability to the correct sentence than to its swapped counterpart (i.e., $P(S_1) > P(S_1')$ and $P(S_2) > P(S_2')$). We report accuracy averaged across frequency bins and visual properties.

\paragraph{Text Score Aggregation}
\label{methods:text-aggregation}
We group the five text tasks into two subgroups (see Sec.~\ref{babyvlm-challenge}): a \textbf{Syntax} subgroup (Zorro accuracy, LongTail-Swap InflectionSwap and AgreementSwap accuracies---three metrics) and a \textbf{Semantics} subgroup (LongTail-Swap WordSwap accuracy and VP-Swap accuracy---two metrics). The subgroup aggregate is the arithmetic mean of the constituent task accuracies, and the \textbf{Overall} text score is the mean of the two subgroup aggregates: $A_\mathrm{overall} = (A_\mathrm{Syntax} + A_\mathrm{Semantics}) / 2$.

\section{Multimodal Evaluations}
\label{methods:multinodal evals}

\subsection{Machine-DevBench}
\label{methods:machinedevbench}

DevBench~\citep{tan2024devbench} was designed to evaluate language comprehension in VLMs, comparing them to infants using paradigms from developmental psychology.
While pioneering in bridging developmental science and machine learning, several limitations make it unsuitable for reliable model evaluation:

\begin{itemize}
    \item \textbf{Limited statistical power:}  DevBench provides very few samples per task---LWL includes only 75 trials covering only 28 distinct words, TROG has 77 trials, and Winoground contains 170---resulting in high variance evaluations that make it difficult to draw reliable conclusions about model capabilities.
    \item \textbf{Vocabulary–corpus mismatch:} the benchmark's vocabulary is not controlled relative to the model's training data, often containing words that may be absent from the training corpus entirely, especially when benchmarking models trained on small naturalistic datasets.
    Across the four training corpora used in this work, between 4\% and 10\% of DevBench's vocabulary is missing from any given corpus, conflating vocabulary coverage with linguistic competence. The mismatch is most severe on DevBench's VV lexical task, where 21\% of evaluation words are absent from BabyView, 24\% from Ego4D, 16\% from HowTo100M, and 6\% from COCO-MC.
    \item \textbf{Stimuli designed for infants:}  The visual stimuli were created for infant experiments and introduce confounds when applied to VLMs: low-level visual features can act as shortcuts that are unrelated to the linguistic property being tested.
    \item \textbf{Narrow grammatical coverage:} DevBench's grammatical tasks cover only a small set of constructions and fail to comprehensively probe the range of grammatical and compositional phenomena relevant to language understanding.
\end{itemize}

\begin{figure}[t]
    \centering
    \begin{subfigure}[t]{0.33\linewidth}
        \includegraphics[width=\linewidth]{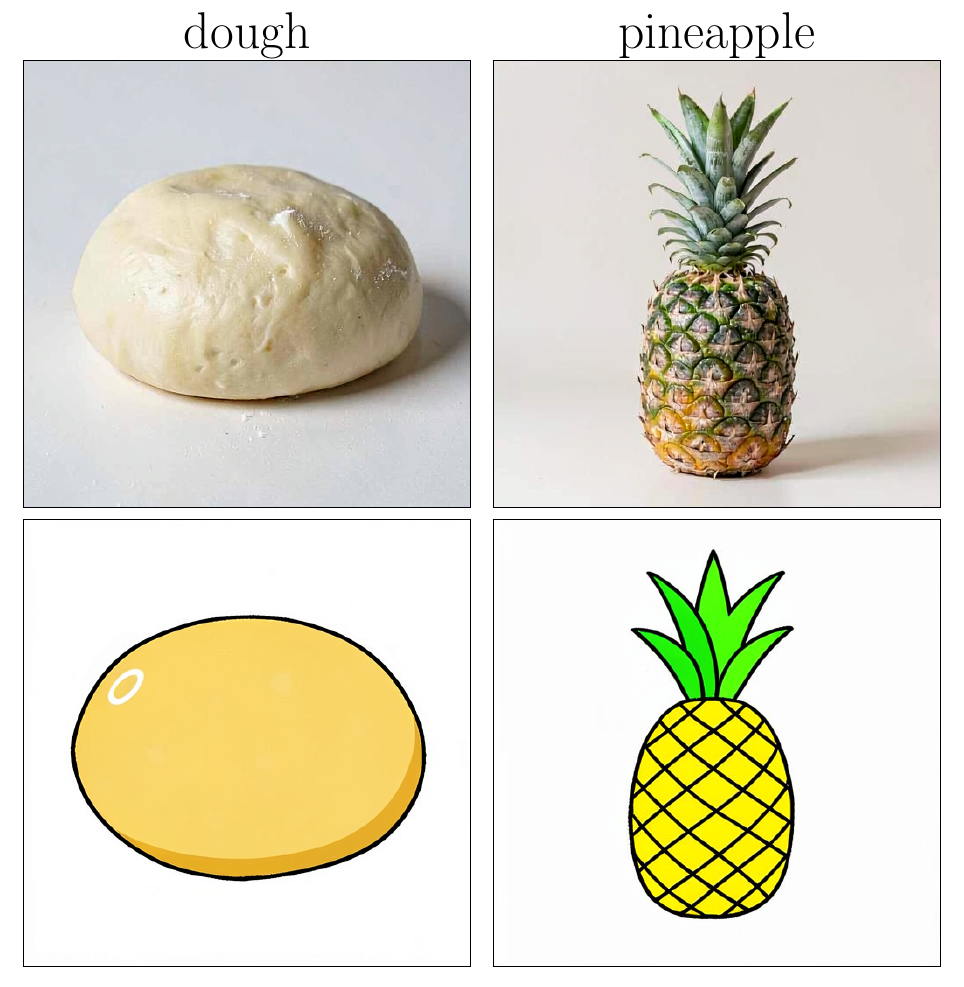}
        \caption{Lexical: Noun recognition}
    \end{subfigure}\hfill
    \begin{subfigure}[t]{0.33\linewidth}
        \includegraphics[width=\linewidth]{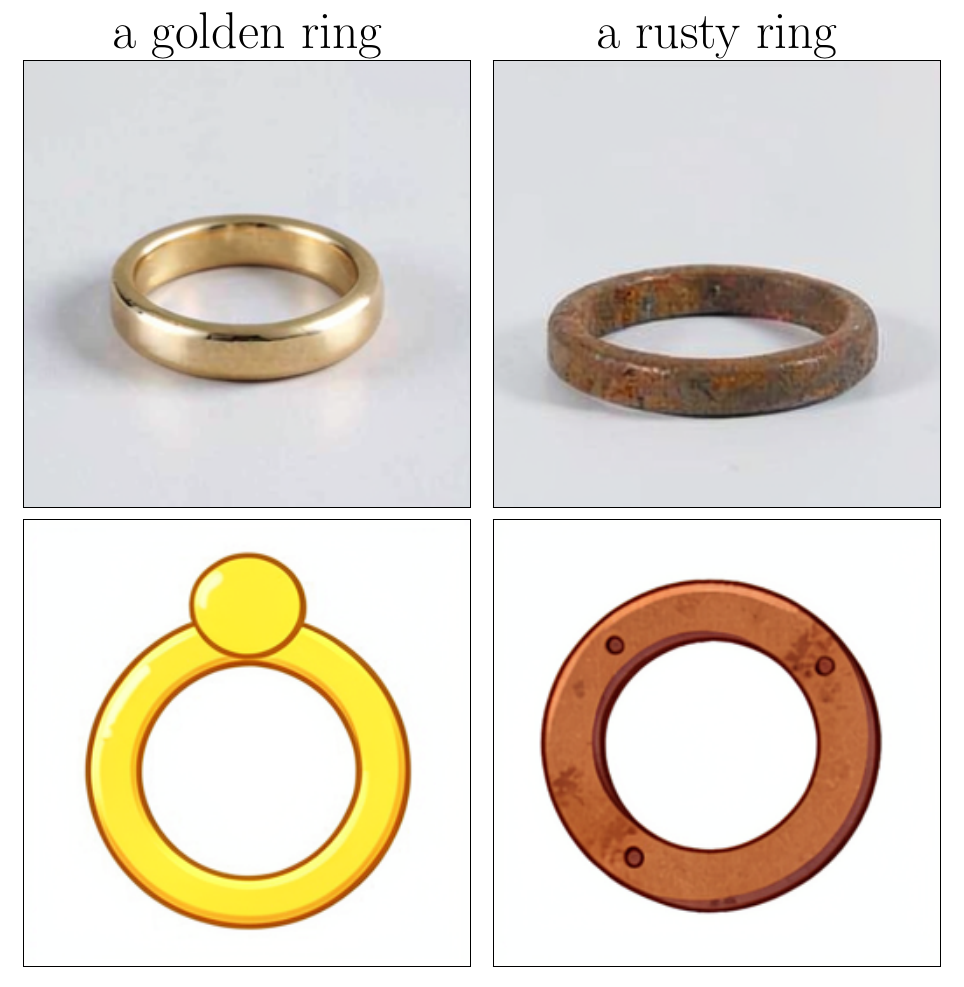}
        \caption{Lexical: Adjective recognition}
     \end{subfigure}\hfill
    \begin{subfigure}[t]{0.33\linewidth}
        \includegraphics[width=\linewidth]{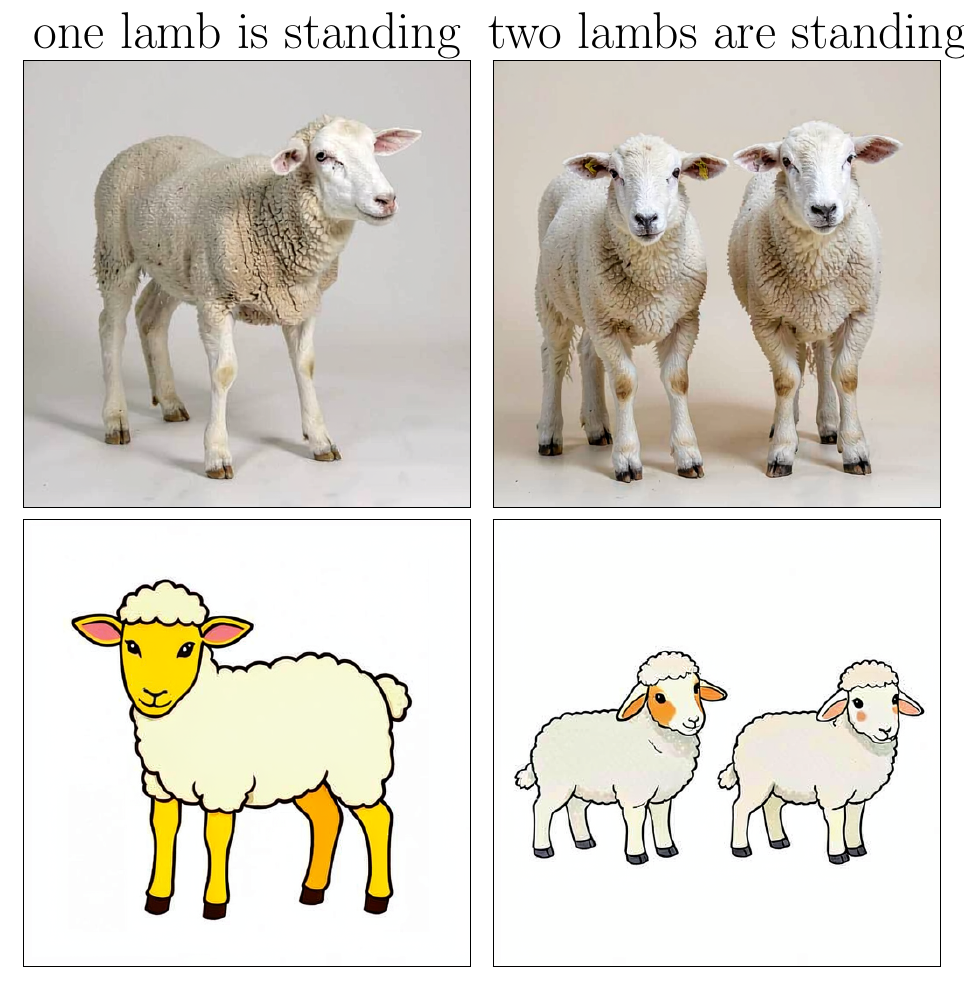}
        \caption{Gram: Counting}
      \end{subfigure}\\[6pt]
    \begin{subfigure}[t]{0.33\linewidth}
        \includegraphics[width=\linewidth]{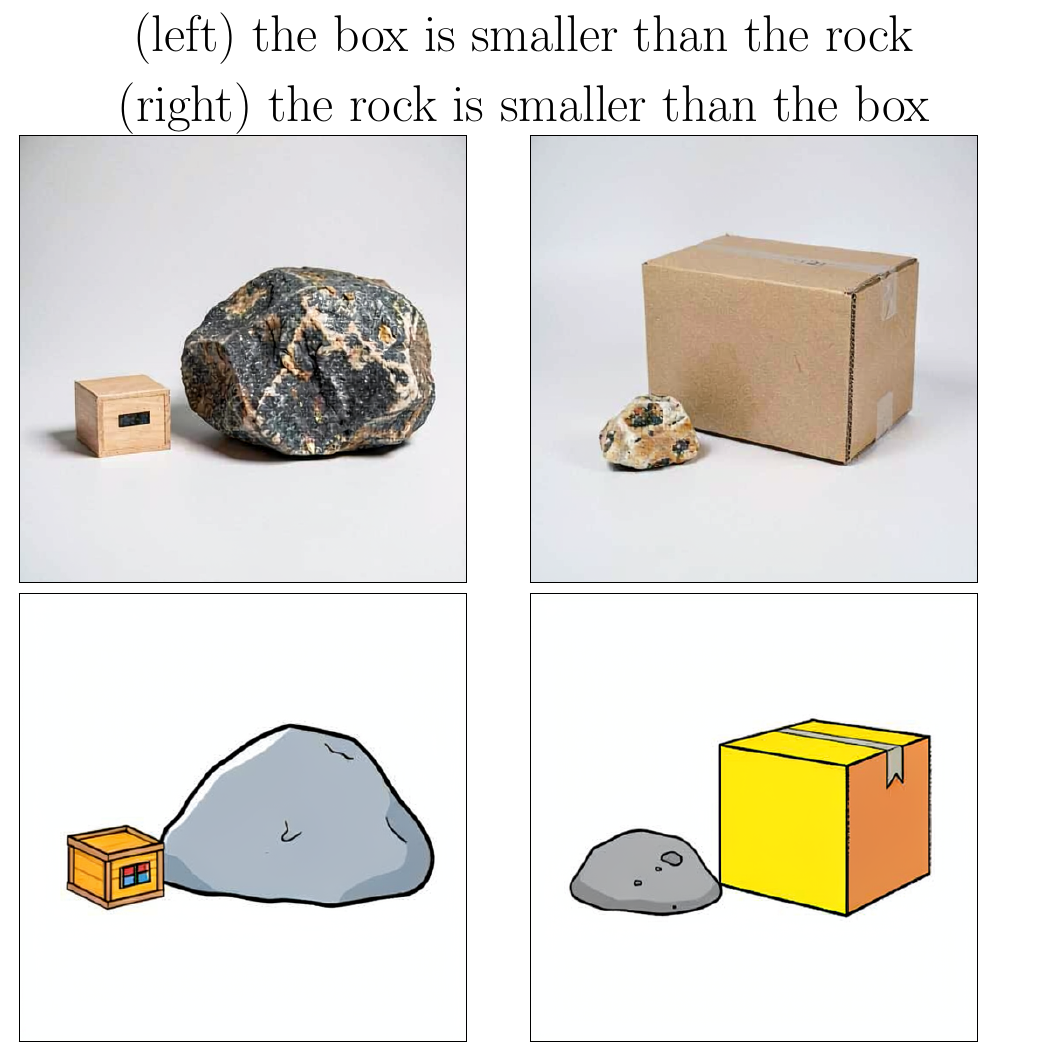}
        \caption{Gram: Comparisons}
     \end{subfigure}\hfill
    \begin{subfigure}[t]{0.33\linewidth}
        \includegraphics[width=\linewidth]{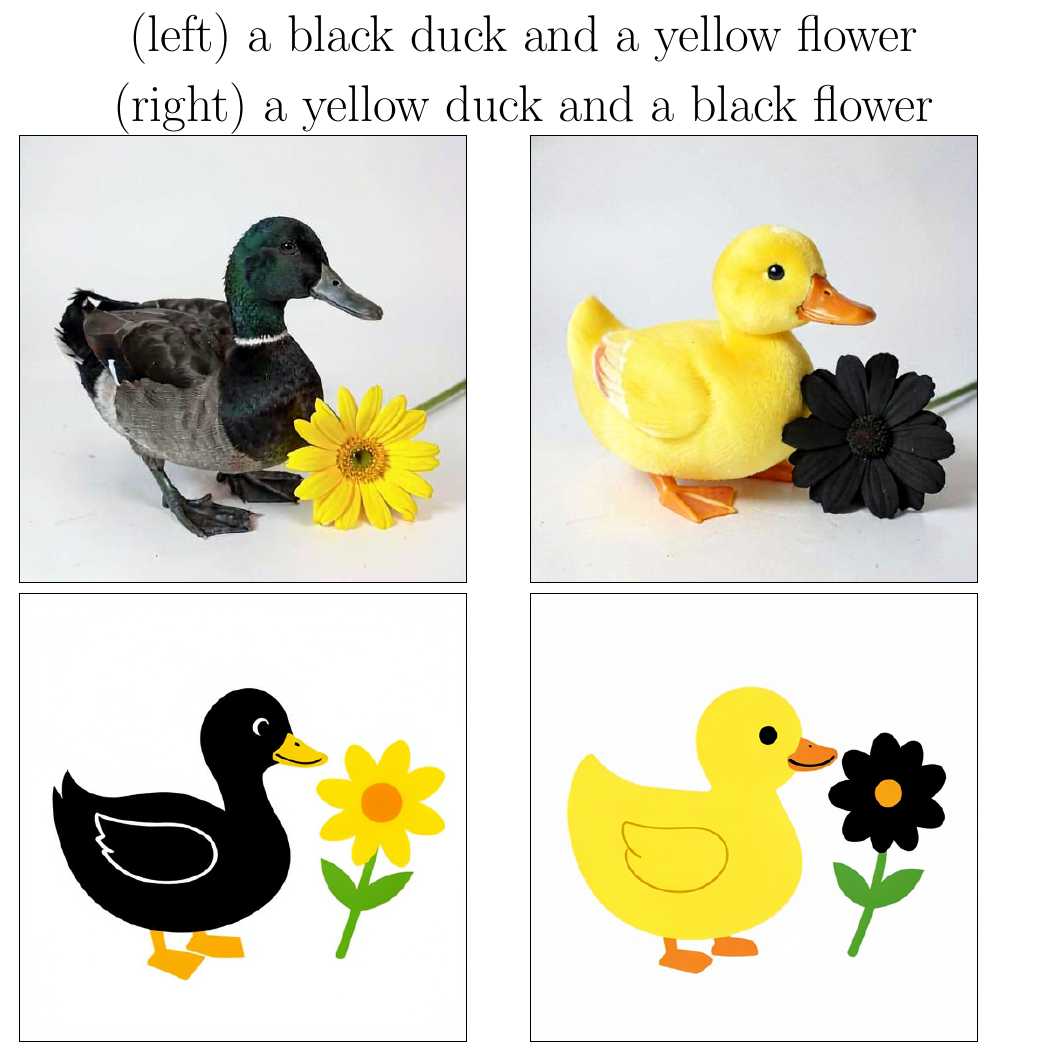}
        \caption{Gram: Attribute assignment}
      \end{subfigure}\hfill
    \begin{subfigure}[t]{0.33\linewidth}
        \includegraphics[width=\linewidth]{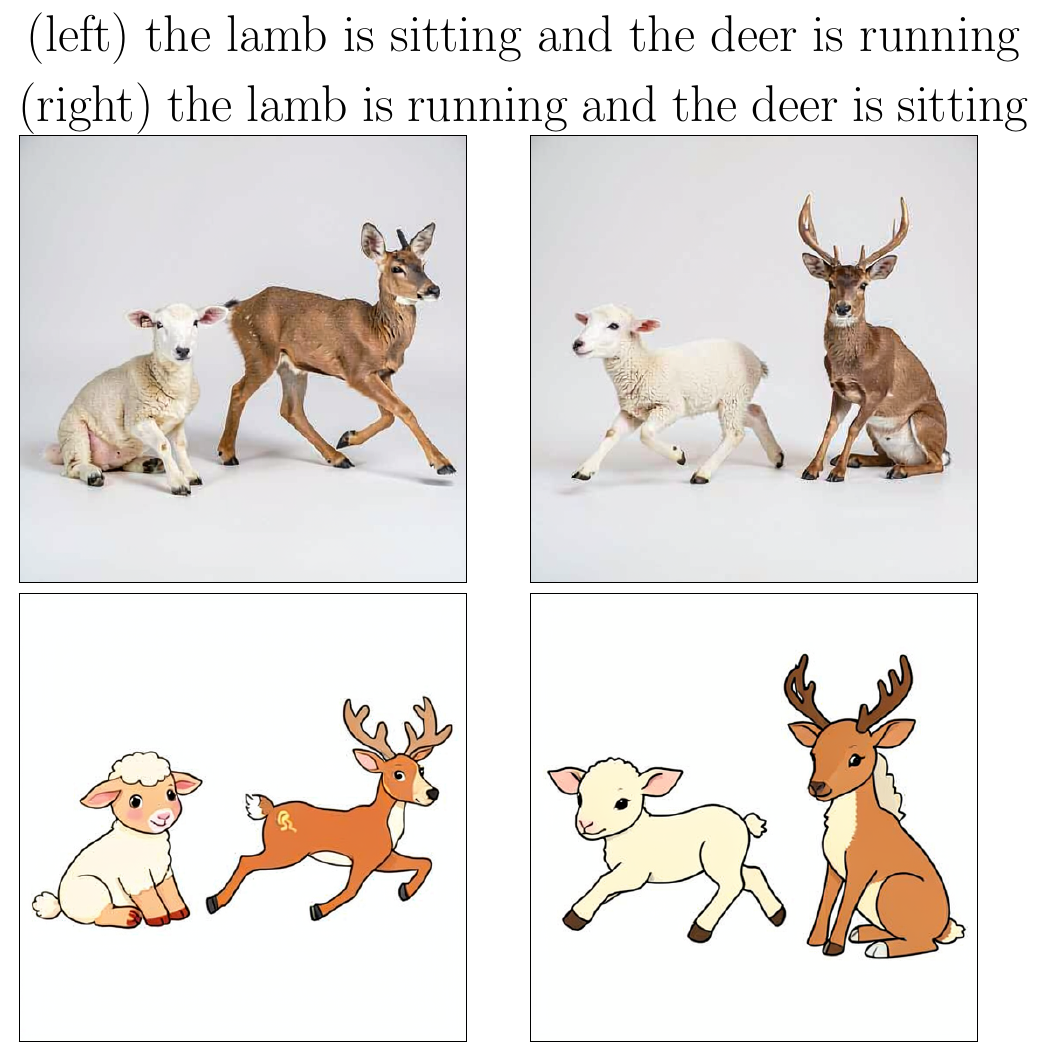}
        \caption{Gram: Action assignment}
     \end{subfigure}\\[6pt]
    \caption{
        Example of Machine-DevBench samples, including lexical (noun and adjective recognition) and several grammatical (gram.) tasks.
        We illustrate trials using both visual styles: realistic and cartoon.
    }
    \label{fig:machinedevbench_tasks}
\end{figure}

These limitations motivate Machine-DevBench, a benchmark that retains the developmental motivation of evaluating emerging linguistic competence while being scalable, corpus-grounded, and more statistically robust.
Machine-DevBench is generated directly from the model's training corpus, ensuring that all evaluation vocabulary consists of concepts the model has been exposed to during training. This eliminates the confound between vocabulary coverage and linguistic competence that affects DevBench.

A central design principle is \emph{long-tail coverage}.
Inspired by LT-Swap~\citep{algayres2025longtail}, words are sampled across the full frequency distribution of the training corpus using logarithmically-spaced frequency bins, ranging from words appearing only once to the most frequent words (appearing over 512 times).
This enables fine-grained analysis of how word frequency in the training data affects model performance---a central question for understanding data-efficient language learning.

The benchmark creation is fully automatic: given any corpus of text utterances, the pipeline produces curated word lists, generates contrastive visual stimuli, and applies multi-stage quality filtering, yielding evaluation-ready data without manual annotation.
The benchmark---built from the vocabulary shared across BabyView, Ego4D, HowTo and COCO-MC---comprises two families of tasks: lexical and grammatical, which are illustrated in Figure~\ref{fig:machinedevbench_tasks} and summarized in Table~\ref{tab:machinedevbench_grammatical}, resulting in over 3{,}700 distinct trials---over 10 times larger that DevBench.
The Machine-DevBench evaluation set used in our experiments, along with code to generate corpus-specific evaluation sets, is available on \href{https://facebookresearch.github.io/egobabyvlm}{our leaderboard page}.

We organize Machine-DevBench around two complementary task families that together cover the core competencies expected of a grounded language model: \textbf{lexical} tasks, which probe whether individual words map to the correct visual referents, and \textbf{grammatical} tasks, which probe whether the model composes those words into structured meaning.
The  two families share a common contrastive format---two images and a linguistic stimulus, with the model selecting the matching image---ensuring that scores are directly comparable across word- and sentence-level competencies. We describe each family below:

\paragraph{Lexical Tasks} Lexical tasks evaluate word-level recognition for nouns and adjectives, testing whether a model can match a word to its correct visual referent.
To evaluate noun recognition, each trial presents a target noun alongside two images: one depicting the target and one depicting a distractor noun drawn from the same semantic category (e.g., dog vs. cat, or fork vs. spoon).
The use of same--category distractors ensures that the model must rely on fine-grained lexical distinctions rather than broad categorical cues.
Nouns are organized into semantic categories (e.g., animals, food, vehicles, clothing) derived from WordNet hypernym hierarchies.
Similarly, to evaluate adjective recognition, each sample presents two images of the same object, one exhibiting the target adjective and one exhibiting a contrasting adjective (e.g., a tall building vs. a short building).
By holding the noun constant, this task isolates the model's ability to recognize adjectival properties.

\paragraph{Grammatical Tasks}
The grammatical tasks evaluate linguistic and compositional understanding through sentence-level contrastive pairs.
Each trial presents two captions and two corresponding images, where the model must correctly match each caption to its image.
We define eight distinct grammatical categories---summarized in Table~\ref{tab:machinedevbench_grammatical}---each targeting a distinct linguistic property.

The construction of Machine-DevBench follows an automated, multi-stage pipeline.
Starting from the raw training corpus, all utterances are canonicalized and word frequencies are computed.
Words are assigned to logarithmically-spaced frequency bins and filtered to retain only concrete, visually representable terms---excluding stopwords, proper nouns, named entities, single-character words, and words without WordNet entries.
For lexical tasks, an LLM (Gemma 4 24B~\citep{gemma2024}) generates descriptive captions for each target word as well as contrastive captions for distractor stimuli, ensuring that each trial presents a meaningful linguistic contrast.
For grammatical tasks, the LLM generates sentence-level minimal pairs via structured prompting.
All generated captions are verified to contain only words present in the source vocabulary.

Given the precomputed captions, visual stimuli are generated using the FLUX.2~\citep{flux-2-2025} (more specifically FLUX.2 [klein] 4B) text-to-image diffusion model.
Each trial's images are produced in two visual styles---photorealistic and cartoon---enabling analysis of whether model performance is robust to visual domain shifts.
Category-specific prompt engineering is applied to maximize the visual distinguishability of contrastive pairs (e.g., explicit spatial cues for prepositions, precise numeral enforcement for counting).

To ensure the integrity of the generated stimuli, we implement a multi-stage filtering pipeline.
First, linguistic filters verify correctness, vocabulary coverage, and structural validity of each trial.
Subsequently, a vision–language model scores every image–caption pair for alignment: lexical trials are filtered using PerceptionEncoder, which checks that the correct image scores higher than the distractor for its corresponding caption, while grammatical trials leverage Gemma 4 to assess depiction quality and caption distinguishability.
Only trials that pass all validation stages are retained in the final benchmark.

By coupling developmental motivation with corpus-grounded construction, long-tail frequency control, and large-scale automated stimulus generation, Machine-DevBench turns the evaluation of emerging linguistic competence in VLMs from an underpowered, vocabulary-confounded exercise into a reproducible and statistically meaningful one.
This makes it possible to track progress on small-corpus VLMs over time and to perform fine-grained linguistic analyses across word frequency bins and grammatical phenomena.

\paragraph{Machine-DevBench Score Aggregation}
\label{methods:mdb-aggregation}
We group the 10 Machine-DevBench tasks into two subgroups (see Sec.~\ref{babyvlm-challenge}): a \textbf{Lexical} subgroup (Lexical-Noun and Lexical-Adjective accuracies---two tasks) and a \textbf{Grammatical} subgroup (the eight sentence-level tasks Subject--Verb, Subject--Adjective, Negation, Word Order, Prepositions, Comparatives, Counting, Embedded Relative). Each per-task accuracy is on a $[0, 100]$ scale with chance at $50$. The subgroup aggregate is the arithmetic mean of the constituent task accuracies, and the \textbf{Overall} Machine-DevBench score is the mean of the two subgroup aggregates: $A_\mathrm{overall} = (A_\mathrm{Lexical} + A_\mathrm{Grammatical}) / 2$.

\begin{table}[t]
\caption{Machine-DevBench lexical and grammatical tasks, linguistic properties evaluated, and example sentence pairs illustrating compositional differences.}
\label{tab:machinedevbench_grammatical}
\centering
\small
\renewcommand{\arraystretch}{1.2}
\setlength{\tabcolsep}{3pt}
\setlength{\columnsep}{3pt}
\begin{tabular}{>{\centering\arraybackslash}m{2.8cm} >{\centering\arraybackslash}m{3.7cm} m{5.55cm} >{\raggedleft\arraybackslash}m{1.0cm}}
\hline
\textbf{Task} & \textbf{Linguistic Property} & \textbf{Example} & \textbf{\# Trials} \\
\hline

Lexical-Noun & Noun recognition &
\begin{tabular}{@{}l@{}}
(a) \textit{dog} \\
(b) \textit{cat}
\end{tabular}
&
1{,}792
\\

Lexical-Adjective & Adjective recognition &
\begin{tabular}{@{}l@{}}
(a) \textit{a red mug} \\
(b) \textit{a blue mug}
\end{tabular} &
359
\\

Subject--Verb & Agent--action binding &
\begin{tabular}{@{}l@{}}
(a) \textit{The dog is eating and the cat is sitting} \\
(b) \textit{The dog is sitting and the cat is eating}
\end{tabular} &
152
\\

Subject--Adjective & Property--object binding &
\begin{tabular}{@{}l@{}}
(a) \textit{A red car and a blue truck} \\
(b) \textit{A blue car and a red truck}
\end{tabular} &
169
\\

Negation & Negation comprehension &
\begin{tabular}{@{}l@{}}
(a) \textit{The bird is small} \\
(b) \textit{The bird is not small}
\end{tabular} &
74
\\

Word Order & Thematic role assignment &
\begin{tabular}{@{}l@{}}
(a) \textit{The dog is chasing the cat} \\
(b) \textit{The cat is chasing the dog}
\end{tabular} &
180
\\

Prepositions & Spatial relation understanding &
\begin{tabular}{@{}l@{}}
(a) \textit{The cup on the table} \\
(b) \textit{The cup under the table}
\end{tabular} &
406
\\

Comparatives & Comparative construction &
\begin{tabular}{@{}l@{}}
(a) \textit{The lamp is taller than the candle} \\
(b) \textit{The candle is taller than the lamp}
\end{tabular} &
102
\\

Counting & Numeral comprehension &
\begin{tabular}{@{}l@{}}
(a) \textit{Two dogs are running} \\
(b) \textit{Four dogs are running}
\end{tabular} &
212
\\

Embedded Relative & Relative clause attachment &
\begin{tabular}{@{}l@{}}
(a) \textit{The man feeds the cat that is small} \\
(b) \textit{The man that is small feeds the cat}
\end{tabular} &
275
\\

\hline
\end{tabular}
\end{table}


\subsection{DevBench}
\label{methods:devbench}

In addition to our proposed Machine-DevBench, we also evaluated our multimodal models on DevBench~\citep{tan2024devbench}, a developmentally-inspired benchmark designed to assess VLM's langauge understanding.
Our evaluation methodology closely follows the original DevBench protocol. In short, DevBench comprises seven tasks spanning three cognitive domains. The \textbf{lexical} domain includes Looking-While-Listening (LWL) and Visual Vocabulary (VV), which assess basic word-referent associations and vocabulary comprehension. The \textbf{grammatical} domain includes the Test for Reception of Grammar (TROG) and Winoground, which evaluate syntactic understanding and compositional reasoning. The \textbf{semantic} domain includes Visual Object Categorization (VOC), THINGS and Word Association Task (WAT), which probe conceptual knowledge and semantic similarity structures for images (VOC, THINGS) and text (WAT). We excluded WAT from our evaluation suite because it did not provide meaningful comparison between models or datasets since all models received the same score, as is the case in \citep{tan2024devbench}. Instead, we used VP-Swap and WordSwap tasks to evaluate model's understanding of the semantics of words.

Following the DevBench methodology, we extracted features for textual and image inputs. For \textbf{lexical} and \textbf{grammatical} tasks, we computed cosine similarity scores between image and text representations, then we computed classification accuracy by determining whether models assign higher similarity scores to correct image-text pairs compared to foils, and calculated the softmax-optimized Kullback-Leibler (KL) divergence between model predictions and human behavioral patterns to assess alignment with human-like processing. For \textbf{semantic} tasks, we extracted image embeddings from the visual encoder to analyze representational structure: for VOC, we examined categorical organization in the embedding space, while for THINGS, we computed representational similarity matrices to compare with human behavioral data. We then computed Spearman correlations between model-derived representational dissimilarity matrices and human behavioral similarity judgments, following standard representational similarity analysis (RSA) protocols.

We aggregated results across tasks by accuracy and model-human similarity. To compute the aggregated accuracy for lexical and grammatical tasks, we calculated the equally-weighted average of the accuracies on the four tasks. To computed the model-human similarity aggregate, we combined the KL divergence measures, converted to higher being better via $\frac{1}{1 + \text{KL}}$, from lexical and grammatical tasks with Spearman correlations from semantic tasks, using an equally-weighted average.

Similarly to the findings reported in Section \ref{sec:findings}, semantic alignment is strongly correlated with downstream language grounding. Across datasets, both aggregated accuracy (Table~\ref{tab:devbench_accuracy}) and human-similarity (Table~\ref{tab:devbench_similarity}) are high for models trained on highly aligned data. As a sanity check, we further verify that performance on our Machine-DevBench tracks performance on the human-curated DevBench: across all 16 of our trained models, the Machine-DevBench Overall Agg is strongly positively correlated with both DevBench Agg Accuracy and DevBench Agg Alignment (Fig.~\ref{fig:mdb_vs_devbench_scatter}), indicating that Machine-DevBench is a reliable proxy for DevBench.

\begin{table}[t]
    \caption{Language grounding accuracy metrics (DevBench). For CLIP+ models we report the mean and std across 3 training seeds, with the best two results in each section / column bold and underlined, respectively (Chance excluded). The top section contains off-the-shelf models; the bottom section contains CVCL and our trained models.}
    \label{tab:devbench_accuracy}
    \centering
    \small
    \newcolumntype{s}{>{\scriptsize}r}
    \newcolumntype{t}{>{\scriptsize}l}

    \begin{NiceTabular}{
        l
        r@{.}l@{\scriptsize$\,\pm\,$}s@{.}t
        r@{.}l@{\scriptsize$\,\pm\,$}s@{.}t
        r@{.}l@{\scriptsize$\,\pm\,$}s@{.}t
        r@{.}l@{\scriptsize$\,\pm\,$}s@{.}t
        | r@{.}l@{\scriptsize$\,\pm\,$}s@{.}t
    }
    \toprule
    \textbf{Model}
      & \multicolumn{4}{c}{\textbf{LWL acc $\uparrow$}}
      & \multicolumn{4}{c}{\textbf{VV acc$\uparrow$}}
      & \multicolumn{4}{l}{\textbf{TROG acc$\uparrow$}}
      & \multicolumn{4}{c}{\textbf{WG acc$\uparrow$}}
      & \multicolumn{4}{c}{\textbf{Agg acc.$\uparrow$}} \\
    \midrule
    \rowcolor[gray]{0.95}
    \RowStyle{} Chance & \multicolumn{4}{c}{50.0} & \multicolumn{4}{c}{25.0} & \multicolumn{4}{c}{25.0} & \multicolumn{4}{c}{50.0} & \multicolumn{4}{c}{37.5} \\
    \RowStyle{} CLIP {\scriptsize\citep{radford2021learning}} & \multicolumn{4}{c}{\underline{98.7}} & \multicolumn{4}{c}{\textbf{95.8}} & \multicolumn{4}{c}{\underline{46.2}} & \multicolumn{4}{c}{\underline{60.5}} & \multicolumn{4}{c}{\underline{75.3}} \\
    \RowStyle{} LLaVA-v1.6-Mistral-7B {\scriptsize\citep{liu2023visual}} & \multicolumn{4}{c}{\textbf{100.0}} & \multicolumn{4}{c}{\textbf{95.8}} & \multicolumn{4}{c}{\textbf{65.4}} & \multicolumn{4}{c}{\textbf{69.9}} & \multicolumn{4}{c}{\textbf{82.8}} \\
    \midrule
    \RowStyle{} CVCL {\scriptsize\citep{vong2024grounded}} & \multicolumn{4}{c}{59.2} & \multicolumn{4}{c}{27.7} & \multicolumn{4}{c}{23.1} & \multicolumn{4}{c}{28.7} & \multicolumn{4}{c}{34.7} \\
    BabyView-CLIP+ & {57}&{0} & {2}&{0} & {30}&{0} & {0}&{5} & {27}&{8} & {1}&{5} & {48}&{1} & {0}&{8} & {40}&{7} & {0}&{3} \\
    \RowStyle{} BabyView-LLaVA & {54}&{4} & {9}&{3} & {28}&{6} & {2}&{2} & {26}&{1} & {4}&{5} & {50}&{0} & {1}&{9} & {39}&{8} & {3}&{1} \\
    Ego4D-CLIP+ & {55}&{3} & {2}&{6} & {28}&{3} & {4}&{2} & {23}&{9} & {1}&{5} & {47}&{9} & {1}&{4} & {38}&{8} & {0}&{8} \\
    \RowStyle{} Ego4D-LLaVA & {57}&{5} & {7}&{7} & {21}&{9} & {4}&{7} & {29}&{1} & {4}&{8} & {49}&{9} & {1}&{2} & {39}&{6} & {0}&{2} \\
    HowTo-CLIP+ & {74}&{6} & {3}&{0} & {37}&{5} & {3}&{8} & {29}&{9} & {6}&{1} & {48}&{9} & {1}&{2} & {47}&{7} & {3}&{4} \\
    \RowStyle{} HowTo-LLaVA & {61}&{4} & {5}&{0} & {25}&{2} & {6}&{6} & {29}&{9} & {4}&{1} & {50}&{9} & {0}&{0} & {41}&{9} & {0}&{2} \\
    {COCO-MC}-CLIP+ & {\underline{96}}&{\underline{1}} & {1}&{4} & {\bfseries 62}&{\bfseries 2} & {3}&{7} & {\underline{40}}&{\underline{1}} & {0}&{8} & {\underline{54}}&{\underline{8}} & {1}&{7} & {\bfseries 63}&{\bfseries 3} & {1}&{1} \\
    \RowStyle{} {COCO-MC}-LLaVA & {80}&{7} & {7}&{7} & {36}&{1} & {4}&{4} & {39}&{7} & {4}&{6} & {54}&{1} & {0}&{8} & {52}&{7} & {3}&{9} \\
    {COCO-MC}\textsubscript{shuf25\%}-CLIP+ & {\bfseries 97}&{\bfseries 8} & {1}&{5} & {\underline{55}}&{\underline{2}} & {1}&{3} & {38}&{5} & {1}&{3} & {\bfseries 56}&{\bfseries 0} & {1}&{2} & {\underline{61}}&{\underline{9}} & {0}&{1} \\
    \RowStyle{} {COCO-MC}\textsubscript{shuf25\%}-LLaVA & {82}&{9} & {4}&{6} & {42}&{3} & {12}&{6} & {32}&{9} & {1}&{5} & {50}&{1} & {0}&{2} & {52}&{0} & {3}&{1} \\
    {COCO-MC}\textsubscript{shuf50\%}-CLIP+ & {\underline{96}}&{\underline{5}} & {1}&{6} & {51}&{3} & {2}&{6} & {\bfseries 41}&{\bfseries 9} & {5}&{4} & {52}&{9} & {1}&{5} & {60}&{7} & {2}&{1} \\
    \RowStyle{} {COCO-MC}\textsubscript{shuf50\%}-LLaVA & {69}&{7} & {1}&{3} & {32}&{5} & {3}&{2} & {31}&{6} & {2}&{7} & {49}&{5} & {3}&{5} & {45}&{8} & {0}&{9} \\
    {COCO-MC}\textsubscript{shuf75\%}-CLIP+ & {89}&{5} & {2}&{7} & {45}&{4} & {5}&{1} & {37}&{6} & {3}&{2} & {52}&{9} & {1}&{3} & {56}&{3} & {2}&{8} \\
    \RowStyle{} {COCO-MC}\textsubscript{shuf75\%}-LLaVA & 60&{5} & {7}&{9} & {26}&{9} & {1}&{5} & {21}&{8} & {5}&{6} & {50}&{3} & {1}&{5} & {39}&{9} & {3}&{2} \\
    {COCO-MC}\textsubscript{shuf100\%}-CLIP+ & {52}&{2} & {2}&{8} & {24}&{9} & {2}&{4} & {24}&{4} & {3}&{9} & {50}&{4} & {0}&{9} & {38}&{0} & {2}&{3} \\
    \RowStyle{} {COCO-MC}\textsubscript{shuf100\%}-LLaVA & {48}&{2} & {5}&{5} & {24}&{6} & {4}&{8} & {23}&{5} & {3}&{0} & {50}&{0} & {0}&{6} & {36}&{6} & {1}&{4} \\
    \bottomrule
    \end{NiceTabular}
\end{table}

\begin{table}[t]
    \caption{Language grounding model-human similarity metrics (DevBench). For CLIP+ models we report the mean and std across 3 training seeds, with the best two results in each section / column bold and underlined, respectively. The top section contains off-the-shelf models; the bottom section contains CVCL and our trained models. LWL/VV/TROG/WG report $D^*_{\mathrm{KL}}$ where lower is better; THINGS/VOC/Agg sim are correlations where higher is better.}
    \label{tab:devbench_similarity}
    \centering
    \resizebox{\textwidth}{!}{%
    \small
    \setlength{\tabcolsep}{3pt}
    \setlength{\columnsep}{3pt}
    \newcolumntype{s}{>{\scriptsize}r}
    \newcolumntype{t}{>{\scriptsize}l}

    \begin{NiceTabular}{
        l
        r@{.}l@{\scriptsize$\,\pm\,$}s@{.}t
        r@{.}l@{\scriptsize$\,\pm\,$}s@{.}t
        r@{.}l@{\scriptsize$\,\pm\,$}s@{.}t
        r@{.}l@{\scriptsize$\,\pm\,$}s@{.}t
        r@{.}l@{\scriptsize$\,\pm\,$}s@{.}t
        r@{.}l@{\scriptsize$\,\pm\,$}s@{.}t
        | r@{.}l@{\scriptsize$\,\pm\,$}s@{.}t
    }
    \toprule
    \textbf{Model}
      & \multicolumn{4}{c}{\textbf{LWL $\bm{D^*_{\textbf{KL}}}$$\downarrow$}}
      & \multicolumn{4}{c}{\textbf{VV $\bm{D^*_{\textbf{KL}}}$$\downarrow$}}
      & \multicolumn{4}{c}{\textbf{TROG $\bm{D^*_{\textbf{KL}}}$$\downarrow$}}
      & \multicolumn{4}{c}{\textbf{WG $\bm{D^*_{\textbf{KL}}}$$\downarrow$}}
      & \multicolumn{4}{c}{\textbf{THINGS r$\uparrow$}}
      & \multicolumn{4}{c}{\textbf{VOC r$\uparrow$}}
      & \multicolumn{4}{c}{\textbf{Agg sim.$\uparrow$}} \\
    \midrule

    \RowStyle{} CLIP        & \multicolumn{4}{c}{\textbf{0.026}} & \multicolumn{4}{c}{\underline{0.201}} & \multicolumn{4}{c}{\underline{0.733}} & \multicolumn{4}{c}{\underline{0.254}} & \multicolumn{4}{c}{\textbf{0.397}} & \multicolumn{4}{c}{\textbf{-0.081}} & \multicolumn{4}{c}{\textbf{0.583}} \\
    \RowStyle{} LLaVA-v1.6-Mistral-7B {\scriptsize\citep{liu2023visual}} & \multicolumn{4}{c}{\underline{0.032}} & \multicolumn{4}{c}{\textbf{0.191}} & \multicolumn{4}{c}{\textbf{0.538}} & \multicolumn{4}{c}{\textbf{0.226}} & \multicolumn{4}{c}{-} & \multicolumn{4}{c}{-} & \multicolumn{4}{c}{-} \\
    \midrule
    \RowStyle{} CVCL        & \multicolumn{4}{c}{0.060} & \multicolumn{4}{c}{0.740} & \multicolumn{4}{c}{0.911} & \multicolumn{4}{c}{0.258} & \multicolumn{4}{c}{0.175} & \multicolumn{4}{c}{\underline{0.138}} & \multicolumn{4}{c}{0.496} \\
    BabyView-CLIP+ & {0}&{107} & {0}&{000} & {0}&{650} & {0}&{000} & {0}&{911} & {0}&{000} & {0}&{258} & {0}&{000} & {0}&{122} & {0}&{001} & {0}&{080} & {0}&{002} & {0}&{505} & {0}&{001} \\
    \RowStyle{} BabyView-LLaVA & {0}&{083} & {0}&{007} & {0}&{651} & {0}&{012} & {0}&{906} & {0}&{007} & {0}&{258} & {0}&{001} & {0}&{101} & {0}&{009} & {\textbf{0}}&{\textbf{165}} & {0}&{015} & {0}&{519} & {0}&{001} \\
    Ego4D-CLIP+ & {0}&{080} & {0}&{006} & {0}&{738} & {0}&{004} & {0}&{909} & {0}&{001} & {0}&{258} & {0}&{000} & {0}&{178} & {0}&{020} & {0}&{105} & {0}&{035} & {0}&{517} & {0}&{008} \\
    \RowStyle{} Ego4D-LLaVA & {0}&{082} & {0}&{008} & {0}&{662} & {0}&{003} & {0}&{898} & {0}&{010} & {0}&{258} & {0}&{000} & {0}&{066} & {0}&{028} & {0}&{108} & {0}&{162} & {0}&{503} & {0}&{029} \\
    HowTo-CLIP+ & {0}&{052} & {0}&{004} & {0}&{655} & {0}&{004} & {0}&{891} & {0}&{007} & {0}&{258} & {0}&{000} & {\textbf{0}}&{\textbf{407}} & {0}&{004} & {0}&{034} & {0}&{018} & {0}&{553} & {0}&{002} \\
    \RowStyle{} HowTo-LLaVA & {0}&{084} & {0}&{003} & {0}&{655} & {0}&{006} & {0}&{904} & {0}&{011} & {0}&{258} & {0}&{000} & {0}&{122} & {0}&{010} & {0}&{102} & {0}&{050} & {0}&{512} & {0}&{007} \\
    {COCO-MC}-CLIP+  & {\bfseries 0}&{\bfseries 019} & {0}&{003} & {\bfseries 0}&{\bfseries 520} & {0}&{007} & {\bfseries 0}&{\bfseries 750} & {0}&{028} & {0}&{258} & {0}&{000} & {\underline{0}}&{\underline{367}} & {0}&{003} & {0}&{015} & {0}&{002} & {0}&{565} & {0}&{002} \\
    \RowStyle{} {COCO-MC}-LLaVA & {0}&{053} & {0}&{009} & {0}&{604} & {0}&{023} & {0}&{806} & {0}&{053} & {\bfseries 0}&{\bfseries 257} & {0}&{000} & {0}&{201} & {0}&{009} & {0}&{115} & {0}&{022} & {0}&{540} & {0}&{003} \\
    {COCO-MC}\textsubscript{shuf25\%}-CLIP+  & {\underline{0}}&{\underline{023}} & {0}&{001} & {\underline{0}}&{\underline{562}} & {0}&{018} & {\underline{0}}&{\underline{786}} & {0}&{002} & {0}&{258} & {0}&{000} & {0}&{348} & {0}&{021} & {0}&{085} & {0}&{056} & {\bfseries 0}&{\bfseries 568} & {0}&{007} \\
    \RowStyle{} {COCO-MC}\textsubscript{shuf25\%}-LLaVA & {0}&{050} & {0}&{011} & {0}&{575} & {0}&{068} & {0}&{864} & {0}&{017} & {0}&{258} & {0}&{000} & {0}&{187} & {0}&{025} & {0}&{081} & {0}&{039} & {0}&{531} & {0}&{002} \\
    {COCO-MC}\textsubscript{shuf50\%}-CLIP+  & {0}&{026} & {0}&{002} & {0}&{595} & {0}&{016} & {0}&{786} & {0}&{028} & {0}&{258} & {0}&{000} & {0}&{352} & {0}&{011} & {0}&{088} & {0}&{033} & {\underline{0}}&{\underline{566}} & {0}&{006} \\
    \RowStyle{} {COCO-MC}\textsubscript{shuf50\%}-LLaVA & {0}&{066} & {0}&{006} & {0}&{639} & {0}&{003} & {0}&{890} & {0}&{016} & {0}&{258} & {0}&{000} & {0}&{160} & {0}&{008} & {0}&{098} & {0}&{023} & {0}&{522} & {0}&{001} \\
    {COCO-MC}\textsubscript{shuf75\%}-CLIP+  & {0}&{036} & {0}&{005} & {0}&{645} & {0}&{014} & {0}&{821} & {0}&{023} & {0}&{258} & {0}&{000} & {0}&{367} & {0}&{012} & {0}&{044} & {0}&{042} & {0}&{555} & {0}&{009} \\
    \RowStyle{} {COCO-MC}\textsubscript{shuf75\%}-LLaVA & {0}&{075} & {0}&{010} & {0}&{649} & {0}&{005} & {0}&{906} & {0}&{009} & {0}&{258} & {0}&{000} & {0}&{016} & {0}&{006} & {0}&{015} & {0}&{114} & {0}&{481} & {0}&{018} \\
    {COCO-MC}\textsubscript{shuf100\%}-CLIP+ & {0}&{085} & {0}&{003} & {0}&{742} & {0}&{001} & {0}&{908} & {0}&{002} & {0}&{258} & {0}&{000} & {0}&{110} & {0}&{015} & {-0}&{001} & {0}&{036} & {0}&{487} & {0}&{008} \\
    \RowStyle{} {COCO-MC}\textsubscript{shuf100\%}-LLaVA & {0}&{089} & {0}&{001} & {0}&{658} & {0}&{000} & {0}&{906} & {0}&{004} & {0}&{258} & {0}&{000} & {0}&{013} & {0}&{004} & {-0}&{008} & {0}&{033} & {0}&{474} & {0}&{005} \\

    \bottomrule
    \end{NiceTabular}
    }
\end{table}

\begin{figure}[t]
    \centering
    \includegraphics[width=\linewidth]{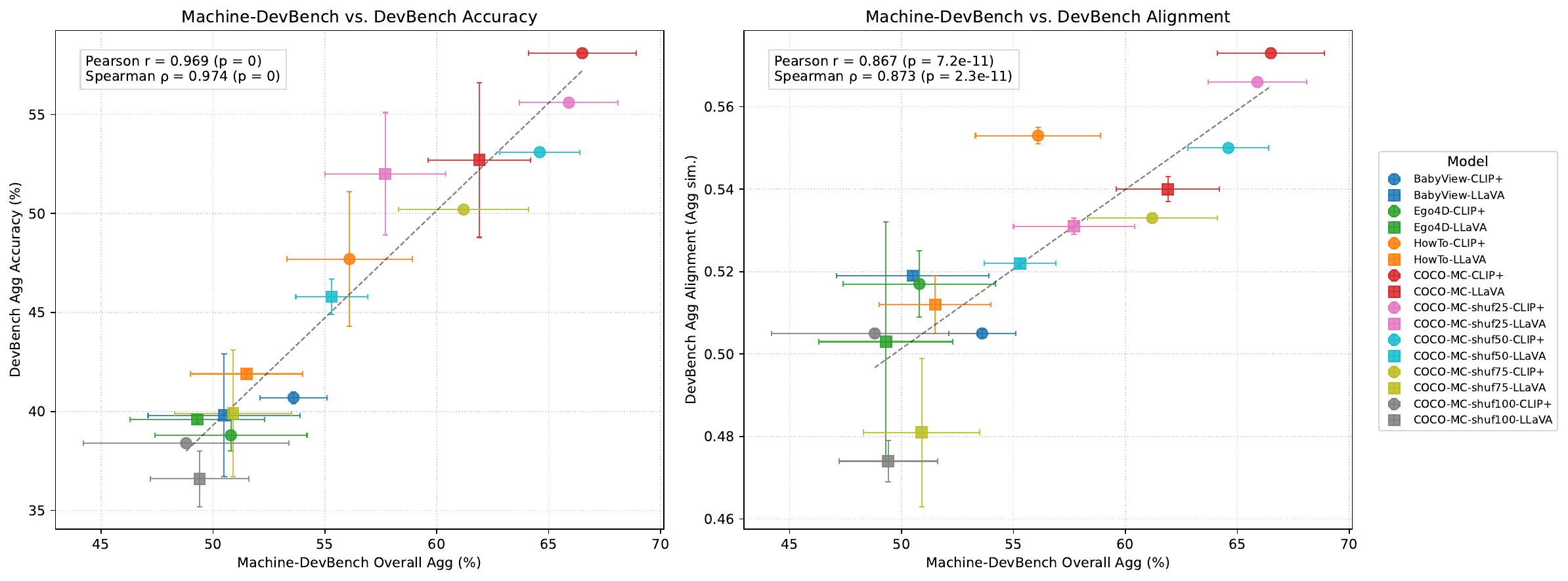}
    \caption{
        \textbf{Machine-DevBench tracks DevBench across our trained models.}
        Each point is one of our 16 trained CLIP+ / LLaVA models (training set
        $\in$ \{BabyView, Ego4D, HowTo, COCO-MC, COCO-MC$_{\text{shuf}\{25,50,75,100\}\%}$\});
        marker shape = encoder family ($\bullet$ CLIP+, $\blacksquare$ LLaVA),
        color = training dataset.
        \textbf{(left)} Machine-DevBench Overall Agg vs.\ DevBench Agg Accuracy.
        \textbf{(right)} Machine-DevBench Overall Agg vs.\ DevBench Agg Alignment
        (model--human similarity, Agg sim.).
        Error bars: $\pm 1$~std across 3 training seeds; dashed line: least-squares fit.
        Pearson $r$ and Spearman $\rho$ (with two-sided $p$-values) are annotated in each panel.
        Strong positive correlations on both panels indicate that performance on
        Machine-DevBench is a reliable proxy for performance on the human-curated
        DevBench, both for grounding accuracy and for human alignment.
    }
    \label{fig:mdb_vs_devbench_scatter}
\end{figure}


\section{CLIP+ Ablation Studies}
\label{section:ablation}
We conducted ablation studies to justify our CLIP+ architecture choices: (1) using pretrained encoders versus random initialization, and (2) interleaving unimodal losses during contrastive training.

\subsection{Pretrained encoders}
How do children learn the mapping between language and the external world so efficiently? We hypothesize that this is because children do not start with random sensory representations; rather, evolution has endowed them with sensory encoders that can be readily mapped to linguistic codes. In machine learning terms, this manifests as the interaction between the objectives that unimodal encoders are optimized on and the objectives used to align multiple modalities. We ask: does pretraining image and text encoders on unimodal objectives improve performance on multimodal language grounding tasks after contrastive learning? And do the benefits of pretraining scale with semantic alignment of the training data?

We compared four configurations: (1) random initialization with contrastive loss only, (2) pretrained BERT + random vision encoder, (3) pretrained DINO + random text encoder, and (4) both pretrained (CLIP+). For configurations 2-4, encoders were pretrained on the target dataset using self-supervised objectives (DINO for vision, masked language modeling for BERT), with interleaved training during contrastive finetuning.

\paragraph{Result.}
Pretraining is consistently beneficial for multimodal grounding, and pretraining \emph{both} encoders yields the strongest overall performance (Tab.~\ref{table:ablation_encoders_devbench}).
Across datasets, BERT-only generally provides larger gains than DINO-only on DevBench, but the best results require both modalities to start from strong unimodal representations. Notably, the benefit of pretraining persists even in weak-alignment regimes (e.g., Ego4D), suggesting that unimodal pretraining provides a crucial foundation even when alignment limits the ultimate ceiling. Model--human similarity follows the same qualitative pattern (Tab.~\ref{table:ablation_encoders_devbench}).

\begin{table}[t]
    \centering
    \caption{Language grounding accuracy and model-human similarity metrics (DevBench) across pretraining strategies. Reporting mean and std across all hyperparameters and seeds for each training scheme. Bolding indicates the highest value in each column.}
    \label{table:ablation_encoders_devbench}
    \resizebox{\textwidth}{!}{%
    \small
    \newcolumntype{s}{>{\scriptsize}r}
    \newcolumntype{t}{>{\scriptsize}l}

    \begin{NiceTabular}{
        l
        r@{.}l@{$\,\pm\,$}s@{.}t r@{.}l@{$\,\pm\,$}s@{.}t
        r@{.}l@{$\,\pm\,$}s@{.}t r@{.}l@{$\,\pm\,$}s@{.}t
        r@{.}l@{$\,\pm\,$}s@{.}t r@{.}l@{$\,\pm\,$}s@{.}t
        r@{.}l@{$\,\pm\,$}s@{.}t r@{.}l@{$\,\pm\,$}s@{.}t
    }

    \toprule
      &
      \multicolumn{8}{c}{\textbf{BabyView}} &
      \multicolumn{8}{c}{\textbf{Ego4D}} &
      \multicolumn{8}{c}{\textbf{HowTo}} &
      \multicolumn{8}{c}{\textbf{COCO-MC}}
      \\
    \cmidrule(lr){2-9} \cmidrule(lr){10-17} \cmidrule(lr){18-25} \cmidrule(lr){26-33}
      &
      \multicolumn{4}{c}{Agg Acc$\uparrow$} & \multicolumn{4}{c}{Agg Sim$\uparrow$} &
      \multicolumn{4}{c}{Agg Acc$\uparrow$} & \multicolumn{4}{c}{Agg Sim$\uparrow$} &
      \multicolumn{4}{c}{Agg Acc$\uparrow$} & \multicolumn{4}{c}{Agg Sim$\uparrow$} &
      \multicolumn{4}{c}{Agg Acc$\uparrow$} & \multicolumn{4}{c}{Agg Sim$\uparrow$}
      \\
    \midrule

    No Pretrain (one loss)
      & {32}&{5} & {2}&{5} & {0}&{506} & {0}&{012}
      & {33}&{5} & {3}&{2} & {0}&{511} & {0}&{011}
      & {30}&{4} & {3}&{3} & {0}&{514} & {0}&{014}
      & 35&8 & 1&4 & 0&520 & 0&009
      \\

    BERT (double interleave)
      & {38}&{2} & {0}&{3} & {0}&{483} & {0}&{003}
      & {35}&{5} & {3}&{2} & {0}&{513} & {0}&{025}
      & {37}&{1} & {4}&{5} & {0}&{512} & {0}&{018}
      & 50&8 & 2&3 & 0&522 & 0&000
      \\

    DINO (double interleave)
      & {33}&{3} & {0}&{8} & {\textbf{0}}&{\textbf{507}} & {0}&{006}
      & {34}&{9} & {4}&{4} & {0}&{506} & {0}&{008}
      & {34}&{4} & {3}&{6} & {0}&{517} & {0}&{025}
      & 47&4 & 2&0 & 0&552 & 0&007
      \\

    \textbf{``CLIP+''}: BERT + DINO (triple interleave)
      & {\textbf{40}}&{\textbf{7}} & {0}&{3} & {0}&{505} & {0}&{001}
      & {\textbf{38}}&{\textbf{8}} & {0}&{8} & {\textbf{0}}&{\textbf{517}} & {0}&{008}
      & {\textbf{47}}&{\textbf{7}} & {3}&{4} & {\textbf{0}}&{\textbf{553}} & {0}&{002}
      & \textbf{63}&\textbf{3} & 1&1 & \textbf{0}&\textbf{565} & 0&001
      \\

    \bottomrule
    \end{NiceTabular}
    }
\end{table}

\subsection{Interleaved training}
To prevent catastrophic forgetting when finetuning pretrained encoders, we interleaved unimodal losses (BERT masked LM, DINO self-supervised) with the contrastive objective. From a developmental neuroscience perspective, children's sensory systems continue to develop and refine throughout early childhood, even as they are acquiring language. Visual processing pathways mature well into adolescence, with ongoing synaptic pruning and refinement of representations \citep{gomez2018development}. Similarly, auditory processing continues to develop during the language learning period, with changes in cortical representations and improved discrimination abilities \citep{sharma2009cortical}. This suggests that infants are simultaneously optimizing multiple learning objectives - refining their sensory encoders while learning cross-modal mappings. We model this by interleaving unimodal self-supervised losses with contrastive learning, hypothesizing that continued sensory learning prevents catastrophic forgetting and may even enhance multimodal learning by maintaining high-quality representations.

We ablated four configurations: (1) no interleaving (contrastive only), (2) interleave BERT only, (3) interleave DINO only, and (4) interleave both (CLIP+).

\subsubsection{DevBench results.}
Interleaving unimodal losses has dataset-dependent effects on multimodal grounding (Tab.~\ref{table:ablation-interleave-devbench}). In low-alignment settings (e.g., Ego4D), interleaving is most important: adding unimodal updates—especially for vision—substantially improves aggregate performance relative to contrastive-only training. In high-alignment settings (COCO-MC), performance is less sensitive to the interleaving choice, although interleaving can still improve human-similarity. Overall, the results support CLIP+ as a robust default across alignment regimes (Tab.~\ref{table:ablation-interleave-devbench}).

\begin{table}[t]
    \centering
    \caption{Language grounding accuracy and model-human similarity metrics (DevBench) across interleaving strategies. Reporting mean and std across all hyperparameters and seeds for each training scheme. Bolding indicates the highest value in each column (ties broken by lower std).}
    \label{table:ablation-interleave-devbench}
    \resizebox{\textwidth}{!}{%
    \small
    \newcolumntype{s}{>{\scriptsize}r}
    \newcolumntype{t}{>{\scriptsize}l}

    \begin{NiceTabular}{
        l
        r@{.}l@{$\,\pm\,$}s@{.}t r@{.}l@{$\,\pm\,$}s@{.}t
        r@{.}l@{$\,\pm\,$}s@{.}t r@{.}l@{$\,\pm\,$}s@{.}t
        r@{.}l@{$\,\pm\,$}s@{.}t r@{.}l@{$\,\pm\,$}s@{.}t
        r@{.}l@{$\,\pm\,$}s@{.}t r@{.}l@{$\,\pm\,$}s@{.}t
    }

    \toprule
      &
      \multicolumn{8}{c}{\textbf{BabyView}} &
      \multicolumn{8}{c}{\textbf{Ego4D}} &
      \multicolumn{8}{c}{\textbf{HowTo}} &
      \multicolumn{8}{c}{\textbf{COCO-MC}}
      \\
    \cmidrule(lr){2-9} \cmidrule(lr){10-17} \cmidrule(lr){18-25} \cmidrule(lr){26-33}
    \cmidrule(lr){2-9} \cmidrule(lr){10-17} \cmidrule(lr){18-25}
      &
      \multicolumn{4}{c}{Agg Acc$\uparrow$} & \multicolumn{4}{c}{Agg Sim$\uparrow$} &
      \multicolumn{4}{c}{Agg Acc$\uparrow$} & \multicolumn{4}{c}{Agg Sim$\uparrow$} &
      \multicolumn{4}{c}{Agg Acc$\uparrow$} & \multicolumn{4}{c}{Agg Sim$\uparrow$} &
      \multicolumn{4}{c}{Agg Acc$\uparrow$} & \multicolumn{4}{c}{Agg Sim$\uparrow$}
      \\
    \midrule

    No interleave
      & {40}&{7} & {1}&{0} & {0}&{511} & {0}&{001}
      & {37}&{3} & {3}&{6} & {0}&{505} & {0}&{016}
      & {46}&{2} & {4}&{2} & {0}&{536} & {0}&{034}
     & 59&1 & 0&5 & 0&553 & 0&006
      \\

    Inter BERT
      & {40}&{4} & {0}&{5} & {0}&{503} & {0}&{002}
      & {35}&{9} & {4}&{8} & {0}&{509} & {0}&{020}
      & {42}&{6} & {6}&{2} & {0}&{524} & {0}&{048}
     & 60 & 9 & 0&6 & 0&548 & 0&005
      \\

    Inter DINO
      & {40}&{4} & {0}&{4} & {\textbf{0}}&{\textbf{522}} & {0}&{002}
      & {37}&{2} & {2}&{5} & {0}&{510} & {0}&{004}
      & {46}&{5} & {5}&{1} & {0}&{539} & {0}&{034}
     & \textbf{64} & \textbf{7} & 0&6 & \textbf{0}&\textbf{576} & 0&003
      \\

    \textbf{``CLIP+''}: Inter BERT + DINO
      & {\textbf{40}}&{\textbf{7}} & {0}&{3} & {0}&{505} & {0}&{001}
      & {\textbf{38}}&{\textbf{8}} & {0}&{8} & {\textbf{0}}&{\textbf{517}} & {0}&{008}
      & {\textbf{47}}&{\textbf{7}} & {3}&{4} & {\textbf{0}}&{\textbf{553}} & {0}&{002}
     & 63 & 3 & 1&1 & 0 & 565 & 0&001      \\

    \bottomrule
    \end{NiceTabular}
    }
\end{table}

\subsubsection{Unimodal encoder preservation.}
Beyond multimodal performance, we examine whether interleaving successfully prevents catastrophic forgetting of unimodal capabilities. Tab.~\ref{table:ablation-interleave-image} and~\ref{table:ablation-interleave-text} compare encoder performance with and without interleaving their respective unimodal losses during contrastive training. For both vision and text, interleaving improves performance on unimodal tasks.

Taken together, these ablations motivate CLIP+ as a practical training recipe that (i) improves multimodal grounding and (ii) better preserves unimodal encoder quality than contrastive-only finetuning, particularly in weak-alignment settings.

\begin{table}[htb]
    \centering
    \caption{Unimodal vision scores with and without interleaving DINO loss during contrastive training. Reporting mean and std across all hyperparameters and seeds for each training scheme. Bolding indicates the higher value between No interleave and Inter DINO for each dataset for each task (lower is better for Depth RMSE).}
    \label{table:ablation-interleave-image}
    \resizebox{\textwidth}{!}{%
    \small
    \setlength{\tabcolsep}{3pt}
    \newcolumntype{s}{>{\scriptsize}r}
    \newcolumntype{t}{>{\scriptsize}l}
    \begin{NiceTabular}{ ll r@{.}l@{$\,\pm\,$}s@{.}t r@{.}l@{$\,\pm\,$}s@{.}t r@{.}l@{$\,\pm\,$}s@{.}t r@{.}l@{$\,\pm\,$}s@{.}t r@{.}l@{$\,\pm\,$}s@{.}t r@{.}l@{$\,\pm\,$}s@{.}t r@{.}l@{$\,\pm\,$}s@{.}t r@{.}l@{$\,\pm\,$}s@{.}t r@{.}l@{$\,\pm\,$}s@{.}t | r@{.}l@{$\,\pm\,$}s@{.}t }
    \toprule
      \multirow[c]{2}{*}[-0.5ex]{\textbf{Train Data}}
      & \multirow[c]{2}{*}[-0.5ex]{\textbf{Variant}}
      & \multicolumn{12}{c}{\textbf{ImageNet-1k}}
      & \multicolumn{4}{c}{\textbf{Depth}}
      & \multicolumn{4}{c}{\textbf{Seg}}
      & \multicolumn{8}{c}{\textbf{Count}}
      & \multicolumn{8}{c}{\textbf{MNIST}}
      & \multicolumn{4}{c}{\multirow[c]{2}{*}[-0.5ex]{\textbf{Overall Agg$\uparrow$}}}
    \\
    \cmidrule(lr){3-14} \cmidrule(lr){15-18} \cmidrule(lr){19-22} \cmidrule(lr){23-30} \cmidrule(lr){31-38}
      &
      & \multicolumn{4}{c}{kNN$\uparrow$}
      & \multicolumn{4}{c}{linear$\uparrow$}
      & \multicolumn{4}{c}{ABX$\uparrow$}
      & \multicolumn{4}{c}{RMSE$\downarrow$}
      & \multicolumn{4}{c}{mIoU$\uparrow$}
      & \multicolumn{4}{c}{linear$\uparrow$}
      & \multicolumn{4}{c}{ABX$\uparrow$}
      & \multicolumn{4}{c}{linear$\uparrow$}
      & \multicolumn{4}{c}{ABX$\uparrow$}
      & \multicolumn{4}{c}{}
    \\
    \midrule
    \multirow{2}{*}{BabyView}
        & No interleave & {26}&{7} & {1}&{3} & {37}&{2} & {1}&{8} & {74}&{2} & {1}&{0} & {0}&{852} & {0}&{011} & {0}&{158} & {0}&{006} & {25}&{8} & {1}&{5} & {54}&{4} & {0}&{5} & {91}&{2} & {1}&{9} & {70}&{6} & {0}&{4} & {37}&{0} & {1}&{2} \\
        & Inter DINO    & {\textbf{34}}&{\textbf{2}} & {2}&{5} & {\textbf{50}}&{\textbf{3}} & {2}&{1} & {\textbf{78}}&{\textbf{1}} & {1}&{2} & {\textbf{0}}&{\textbf{521}} & {0}&{022} & {\textbf{0}}&{\textbf{262}} & {0}&{017} & {\textbf{33}}&{\textbf{5}} & {1}&{0} & {\textbf{54}}&{\textbf{5}} & {0}&{7} & {\textbf{97}}&{\textbf{5}} & {0}&{4} & {\textbf{73}}&{\textbf{2}} & {3}&{3} & {\textbf{45}}&{\textbf{3}} & {1}&{9} \\
    \midrule
    \multirow{2}{*}{Ego4D}
        & No interleave & {22}&{8} & {9}&{9} & {40}&{1} & {11}&{3} & {71}&{2} & {5}&{6} & {0}&{643} & {0}&{120} & {0}&{200} & {0}&{077} & {26}&{6} & {2}&{8} & {54}&{2} & {0}&{7} & {92}&{9} & {9}&{3} & {\textbf{73}}&{\textbf{3}} & {3}&{5} & {39}&{2} & {6}&{4} \\
        & Inter DINO    & {\textbf{38}}&{\textbf{6}} & {0}&{9} & {\textbf{53}}&{\textbf{2}} & {0}&{4} & {\textbf{79}}&{\textbf{6}} & {0}&{1} & {\textbf{0}}&{\textbf{503}} & {0}&{003} & {\textbf{0}}&{\textbf{273}} & {0}&{001} & {\textbf{31}}&{\textbf{1}} & {1}&{4} & {\textbf{54}}&{\textbf{5}} & {0}&{2} & {\textbf{97}}&{\textbf{3}} & {0}&{0} & {70}&{4} & {0}&{6} & {\textbf{45}}&{\textbf{5}} & {0}&{6} \\
    \midrule
    \multirow{2}{*}{HowTo}
        & No interleave & {30}&{0} & {15}&{3} & {47}&{0} & {17}&{2} & {79}&{4} & {8}&{1} & {0}&{597} & {0}&{126} & {0}&{225} & {0}&{093} & {\textbf{31}}&{\textbf{9}} & {4}&{5} & {54}&{1} & {0}&{6} & {91}&{6} & {11}&{8} & {71}&{6} & {4}&{5} & {42}&{8} & {8}&{6} \\
        & Inter DINO    & {\textbf{42}}&{\textbf{1}} & {1}&{0} & {\textbf{56}}&{\textbf{4}} & {0}&{8} & {\textbf{79}}&{\textbf{6}} & {0}&{9} & {\textbf{0}}&{\textbf{478}} & {0}&{005} & {\textbf{0}}&{\textbf{291}} & {0}&{004} & {30}&{6} & {1}&{8} & {\textbf{55}}&{\textbf{2}} & {0}&{2} & {\textbf{97}}&{\textbf{7}} & {0}&{0} & {\textbf{73}}&{\textbf{8}} & {0}&{5} & {\textbf{47}}&{\textbf{2}} & {0}&{8} \\
        \midrule
  \multirow{2}{*}{COCO-MC}
      & No interleave & {49}&{3} & {0}&{3} & {69}&{9} & {0}&{4} & {\textbf{91}}&{\textbf{5}} &
  {0}&{3} & {0}&{432} & {0}&{008} & {\textbf{0}}&{\textbf{374}} & {0}&{011} &
  {\textbf{42}}&{\textbf{2}} & {3}&{2} & {55}&{5} & {0}&{5} & {98}&{0} &
  {0}&{1} & {\textbf{74}}&{\textbf{6}} & {1}&{0} & {57}&{6} & {0}&{5} \\
      & Inter DINO    & {\textbf{61}}&{\textbf{5}} & {0}&{1} & {\textbf{71}}&{\textbf{3}} & {0}&{1} &
  {89}&{5} & {0}&{1} & {\textbf{0}}&{\textbf{386}} & {0}&{002} & {0}&{369} &
  {0}&{017} & {38}&{9} & {1}&{8} & {\textbf{56}}&{\textbf{3}} & {0}&{9} &
  {\textbf{98}}&{\textbf{4}} & {0}&{0} & {73}&{9} & {0}&{7} & {\textbf{58}}&{\textbf{6}} &
  {0}&{5} \\
    \bottomrule
    \end{NiceTabular}%
    }
\end{table}

\begin{table}[htb]
    \centering
    \caption{Unimodal language scores with and without interleaving BERT loss during contrastive learning. Reporting mean and std across all hyperparameters and seeds for each training scheme. Bolding indicates the higher value between No interleave and Inter BERT for each dataset for each task (ties broken by lower std).}
    \label{table:ablation-interleave-text}
    \resizebox{\textwidth}{!}{%
    \small
    \newcolumntype{s}{>{\scriptsize}r}
    \newcolumntype{t}{>{\scriptsize}l}
    \begin{NiceTabular}{
        l l
        r@{.}l@{$\,\pm\,$}s@{.}t
        c
        r@{.}l@{$\,\pm\,$}s@{.}t
        r@{.}l@{$\,\pm\,$}s@{.}t
        r@{.}l@{$\,\pm\,$}s@{.}t
        c
        r@{.}l@{$\,\pm\,$}s@{.}t
        c
        | r@{.}l@{$\,\pm\,$}s@{.}t
    }
    \toprule
    \multirow[c]{2}{*}{\rule{0pt}{3.5ex}\textbf{Dataset}}
      & \multirow[c]{2}{*}{\rule{0pt}{3.5ex}\textbf{Variant}}
      & \multicolumn{4}{c}{\multirow[c]{2}{*}{\rule{0pt}{3.5ex}\textbf{Zorro$\uparrow$}}}
      & &
      \multicolumn{12}{c}{\textbf{LT-Swap}}
      & &
      \multicolumn{4}{c}{\multirow[c]{2}{*}{\rule{0pt}{3.5ex}\textbf{VP-Swap$\uparrow$}}}
      & & \multicolumn{4}{c}{\multirow[c]{2}{*}{\rule{0pt}{3.5ex}\textbf{Overall Agg$\uparrow$}}}
      \\
    \cmidrule(lr){6-17}
      & & \multicolumn{4}{c}{} & &
      \multicolumn{4}{c}{\footnotesize\textbf{WordSwap$\uparrow$}}
      & \multicolumn{4}{c}{\footnotesize\textbf{InflSwap$\uparrow$}}
      & \multicolumn{4}{c}{\footnotesize\textbf{AgrSwap$\uparrow$}}
      & & \multicolumn{4}{c}{}
      & & \multicolumn{4}{c}{}
      \\
    \midrule
    \multirow{2}{*}{BabyView}
      & No interleave & {\textbf{75}}&{\textbf{4}} & {0}&{7} & & {\textbf{78}}&{\textbf{6}} & {0}&{1} & {78}&{9} & {0}&{2} & {59}&{0} & {0}&{2} & & {\textbf{64}}&{\textbf{2}} & {0}&{0} & & {71}&{2} & {0}&{2} \\
      & Inter BERT    & {74}&{8} & {0}&{0} & & {78}&{2} & {0}&{1} & {\textbf{79}}&{\textbf{4}} & {0}&{1} & {\textbf{60}}&{\textbf{1}} & {0}&{1} & & {63}&{9} & {0}&{1} & & {\textbf{71}}&{\textbf{2}} & {0}&{1} \\
    \midrule
    \multirow{2}{*}{Ego4D}
      & No interleave & {74}&{6} & {3}&{6} & & {81}&{8} & {1}&{8} & {84}&{9} & {2}&{4} & {\textbf{65}}&{\textbf{6}} & {1}&{3} & & {65}&{0} & {2}&{1} & & {74}&{2} & {2}&{2} \\
      & Inter BERT    & {\textbf{75}}&{\textbf{1}} & {2}&{0} & & {\textbf{82}}&{\textbf{5}} & {0}&{5} & {\textbf{85}}&{\textbf{3}} & {0}&{6} & {64}&{2} & {2}&{2} & & {\textbf{65}}&{\textbf{7}} & {1}&{1} & & {\textbf{74}}&{\textbf{5}} & {1}&{2} \\
    \midrule
    \multirow{2}{*}{HowTo}
      & No interleave & {68}&{4} & {3}&{2} & & {77}&{8} & {2}&{2} & {81}&{2} & {2}&{5} & {62}&{7} & {2}&{6} & & {65}&{5} & {1}&{9} & & {71}&{2} & {2}&{4} \\
      & Inter BERT    & {\textbf{70}}&{\textbf{9}} & {0}&{6} & & {\textbf{86}}&{\textbf{9}} & {0}&{3} & {\textbf{88}}&{\textbf{0}} & {0}&{4} & {\textbf{63}}&{\textbf{7}} & {0}&{9} & & {\textbf{74}}&{\textbf{0}} & {0}&{9} & & {\textbf{77}}&{\textbf{3}} & {0}&{6} \\
  \midrule
  \multirow{2}{*}{COCO-MC}
    & No interleave & {64}&{7} & {0}&{4} & & {86}&{8} & {0}&{6} & {80}&{3} &
  {0}&{4} & {68}&{0} & {0}&{4} & & {74}&{5} & {0}&{0} & & {74}&{9} &
  {0}&{1} \\
    & Inter BERT    & {\textbf{66}}&{\textbf{1}} & {1}&{2} & & {\textbf{86}}&{\textbf{8}} & {0}&{6} &
  {\textbf{81}}&{\textbf{0}} & {0}&{4} & {\textbf{68}}&{\textbf{4}} & {0}&{2} & &
  {\textbf{74}}&{\textbf{7}} & {0}&{1} & & {\textbf{75}}&{\textbf{4}} & {0}&{3} \\

    \bottomrule
    \end{NiceTabular}
    }
\end{table}


\section{Further Analysis on Semantic Alignment}
\label{app:semantic_alignment_further}

\subsection{Comparison of scoring methods}

To further validate the robustness of our semantic alignment scoring method, we computed Jensen–Shannon divergence alignment using the following additional scoring methods.

\subsubsection{Methods}

\paragraph{VQAScore}

We computed VQAScores \citep{vqascore} using a Perception-LM 8B \citep{cho2025perceptionlm} scoring model. Prior work has shown that VQAScore correlates better with human prompt adherence assessments in text-to-image/video generation than CLIPScore and other automatic metrics \citep{vqascore, wiles2025revisiting}, and might therefore be a good indicator of semantic alignment. The VQAScore for an image/video-text pair is computed by feeding the visual input along with a VQA prompt into the model and extracting the probability of an affirmative output as follows:
\begin{flalign*}
    & P(\text{``Yes''} \hspace{0.25em}| \hspace{0.25em}\textbf{image}, \hspace{0.25em}\text{``Does this figure show `\{\textbf{text}\}'? Please answer `Yes' or `No'.''}), \\
    & P(\text{``Yes''} \hspace{0.25em}| \hspace{0.25em}\textbf{video}, \hspace{0.25em}\text{``Does this video show `\{\textbf{text}\}'? Please answer `Yes' or `No'.''})
\end{flalign*}
We used the default Perception-LM image/video transforms and built on the official implementation.\footnote{\url{https://github.com/facebookresearch/perception_models}} For COCO-MC, we used image-caption pairs, and for video datasets, we extracted video clips with temporally matched utterances as described in Section~\ref{method:frame-extraction}. We generated only a single token with greedy decoding and used the softmax probability of the logits for the \texttt{Yes}-token as $P$(``Yes'').

\paragraph{Semantic Textual Similarity (STS) Score}
While CLIP directly provides cross-modal similarity scores, we also experimented with a cascaded approach, where we first re-captioned the data with Perception-LM (PLM) 8B, and then computed semantic textual similarity scores between original texts and the new captions with an off-the-shelf sentence encoder (SONAR) \citep{duquenne2023sonarsentencelevelmultimodallanguageagnostic}. The motivation for this approach is to reduce the cross-modal alignment problem to an intramodal alignment problem, with re-captioning being an interpretable and somewhat controllable way to translate a visual input into text space.
We followed the same approach as for VQAScore to obtain image/video-text pairs. We used multinomial sampling with temperature $0.6$ to generate captions up to $256$ tokens in length with the pretrained PLM-8B. For SONAR, we used the \texttt{text\_sonar\_basic\_encoder} encoder and official implementation.\footnote{\url{https://github.com/facebookresearch/SONAR}} We computed textual similarity scores as the cosine similarity between SONAR embeddings.

\paragraph{CLIPScore with LoRA-finetuned Perception Encoder}

We hypothesized that off-the-shelf ViT-B~CLIP and Perception Encoder are not well-tuned to our target datasets as they have been trained on large-scale web data, and they may have learned to associate weakly aligned image/video-caption pairs, falsely embedding them with high cosine similarity. To alleviate these issues, we constructed image/video-text pairs using re-captions from Perception-LM 8B, and then used these pairs to LoRA-finetune Perception Encoder via the contrastive InfoNCE loss on the respective target dataset. Through this approach, the model should (1) get accustomed to the target visual inputs and (2) learn to match these visual inputs with highly aligned synthetic captions. At the same time, using LoRA as opposed to full finetuning mitigates effects of catastrophic forgetting.

We used the same re-captions as for STS scores above. We finetuned with LoRA modules in all linear layers of PLM, with $\alpha=32$, $r=32$, and dropout $p=0.05$. We trained for up to $5$ epochs with early stopping using a batch size of $128$. We used the AdamW optimizer with $\beta_1=0.9$,  $\beta_2 = 0.999$, $\epsilon=$\num{1e-8}, weight decay of $0.05$, and a peak learning rate of \num{1e-4} (\num{1e-5} for the logit scale parameter) warmed up linearly over the first epoch, then decayed with a cosine schedule to \num{1e-6}.

After training, we merged the trained LoRA weights into the model and computed cosine similarity scores (CLIPScores) as described in Section~\ref{methods:semantic-alignment}.

\subsubsection{Results}
\begin{figure}[t]
    \centering
    \resizebox{\textwidth}{!}{%
        \includegraphics{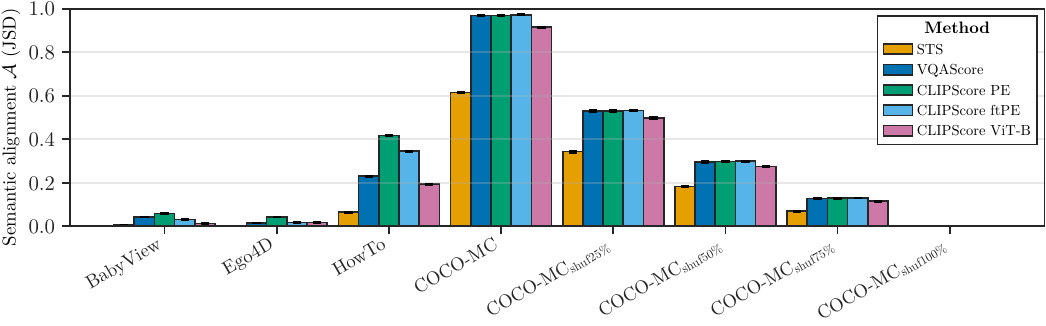}%
    }
    \caption{Comparison of semantic cross-modal alignment scores (JSD) across different scoring methods and datasets.}
    \label{fig:sem_alignment_barplot}
\end{figure}

Fig.~\ref{fig:sem_alignment_barplot} compares JSD alignment scores with bootstrapped 95\% confidence intervals ($n = 1000$) across the different scoring methods and datasets.
The most variable dataset across methods is HowTo, shifting from around 25\% shuffled COCO-MC to around 75\% shuffled COCO-MC, depending on the scoring method. Overall, we find that the five methods are fairly consistent in both their rankings and in terms of absolute JSD scores, validating our hypothesis that naturalistic egocentric video-speech data is largely semantically misaligned, comparable to the level of randomly paired image-caption pairs.

\begin{figure}[htp]
    \centering
    \resizebox{\textwidth}{!}{%
        \includegraphics{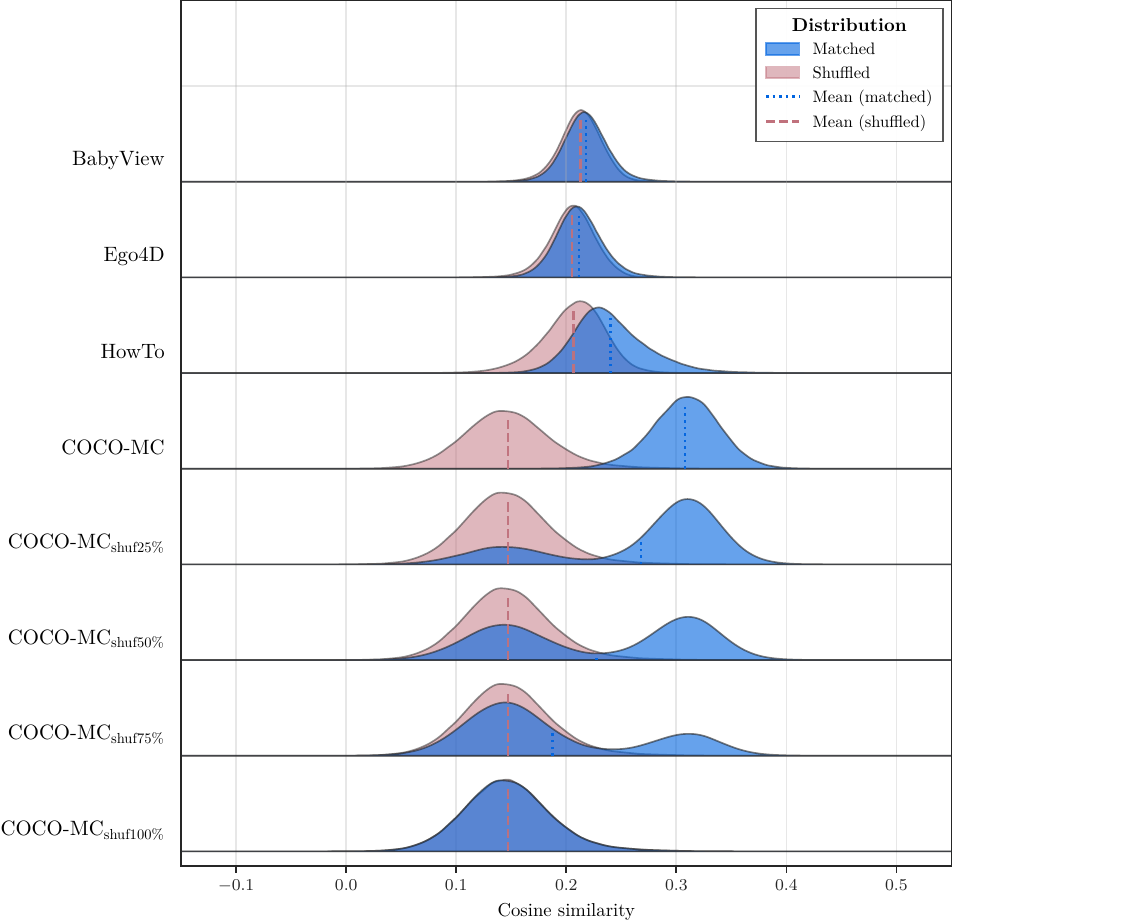}%
    }
    \caption{CLIPScore (ViT-B/16 zero-shot) distributions.}
    \label{fig:cos_sims_clipscore_vitb16}
\end{figure}

\begin{figure}[htp]
    \centering
    \resizebox{\textwidth}{!}{%
        \includegraphics{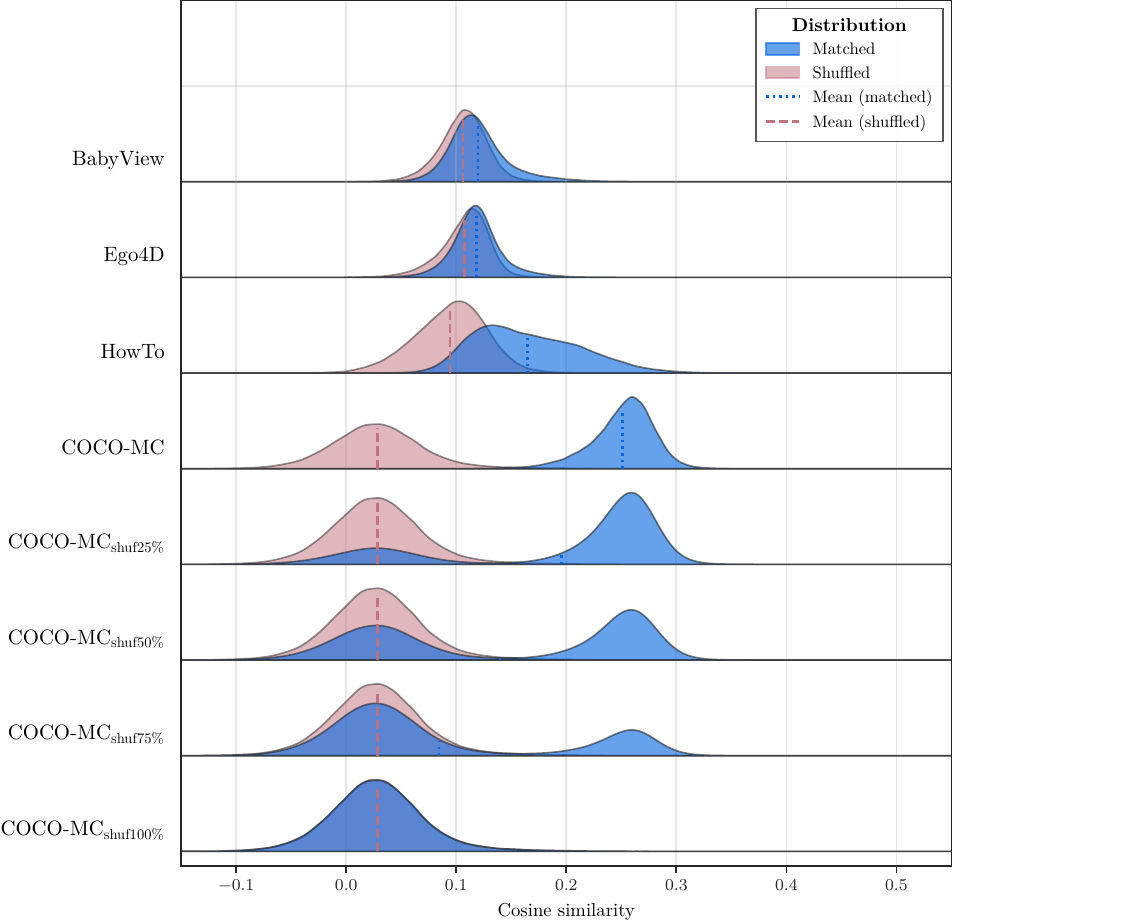}%
    }
    \caption{CLIPScore (PE-Core zero-shot) distributions.}
    \label{fig:cos_sims_clipscore}
\end{figure}

\begin{figure}[htp]
    \centering
    \resizebox{\textwidth}{!}{%
        \includegraphics{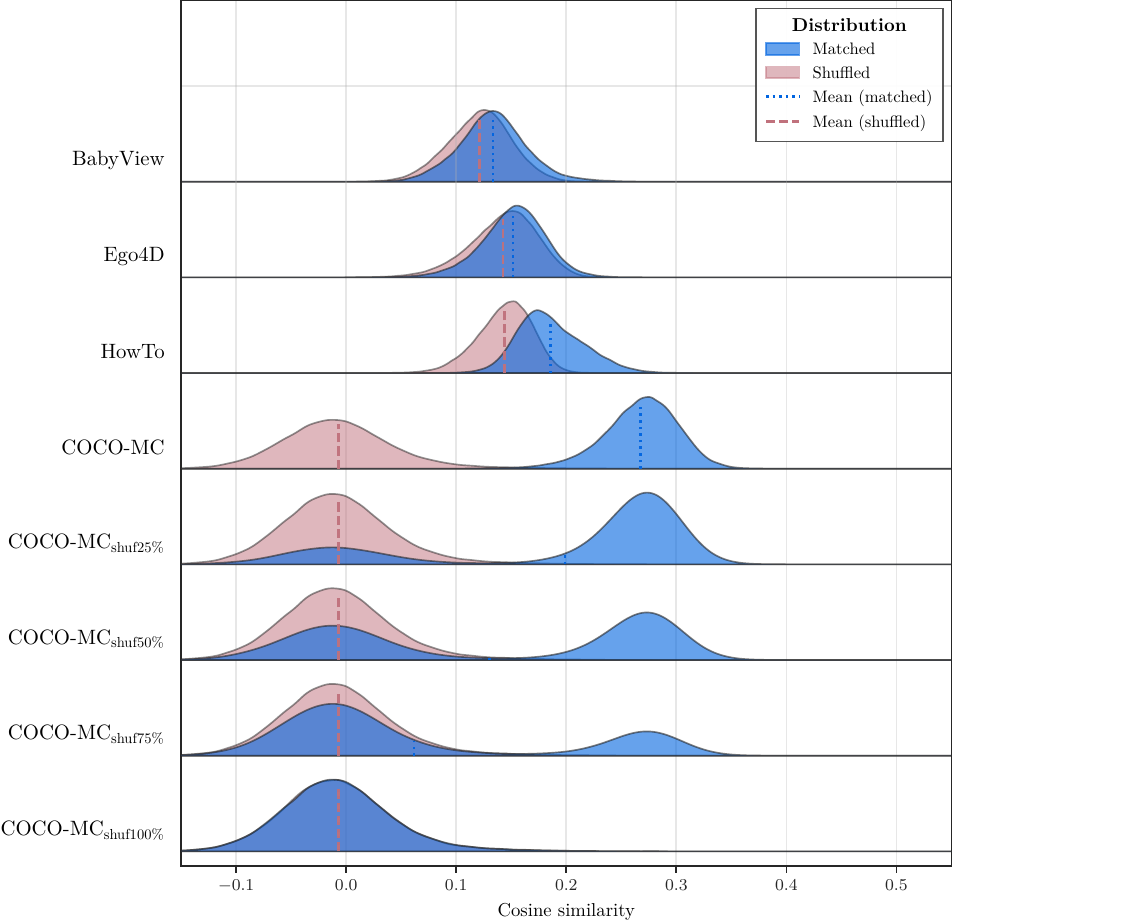}%
    }
    \caption{CLIPScore (PE-Core LoRA-finetuned) distributions.}
    \label{fig:cos_sims_clipscore_v1_ft}
\end{figure}

\begin{figure}[t]
    \centering
    \resizebox{\textwidth}{!}{%
        \includegraphics{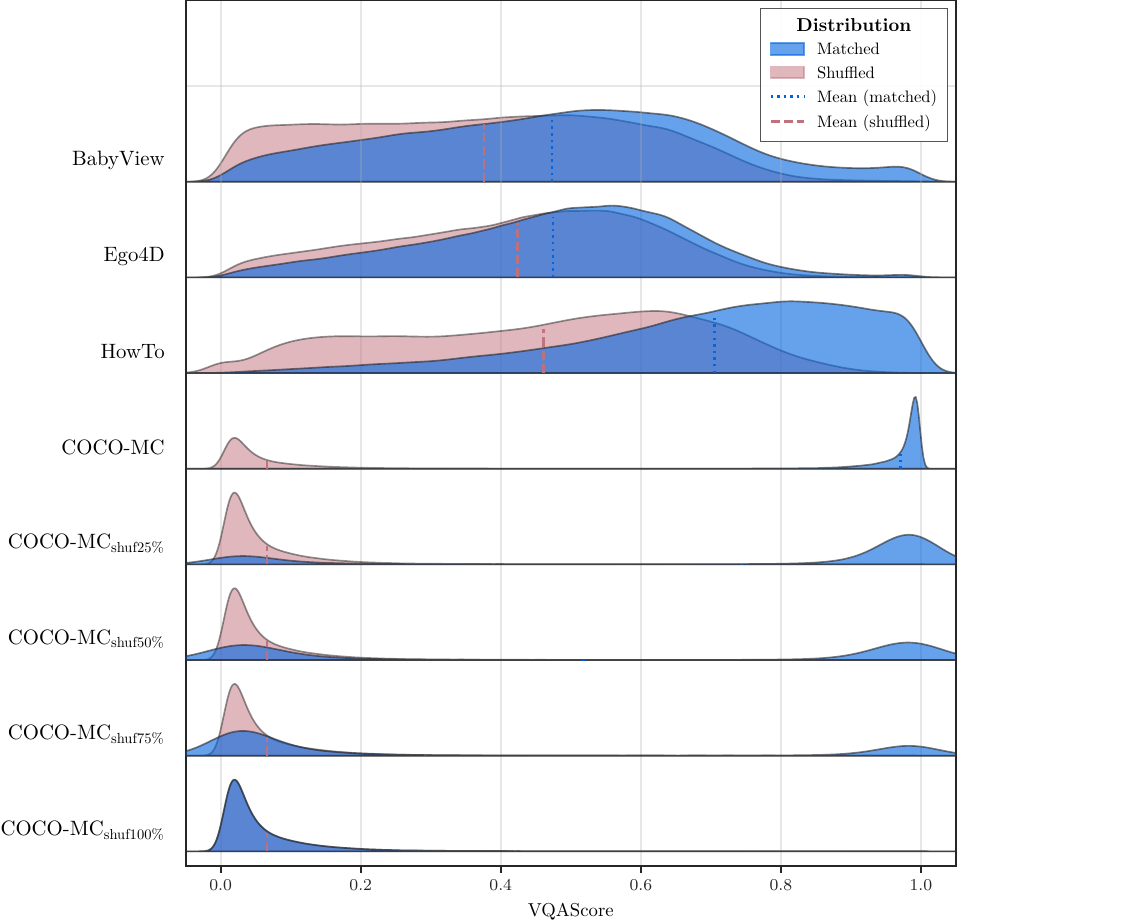}%
    }
    \caption{VQAScore (PLM-8B) distributions.}
    \label{fig:vqascores}
\end{figure}

\begin{figure}[t]
    \centering
    \resizebox{\textwidth}{!}{%
        \includegraphics{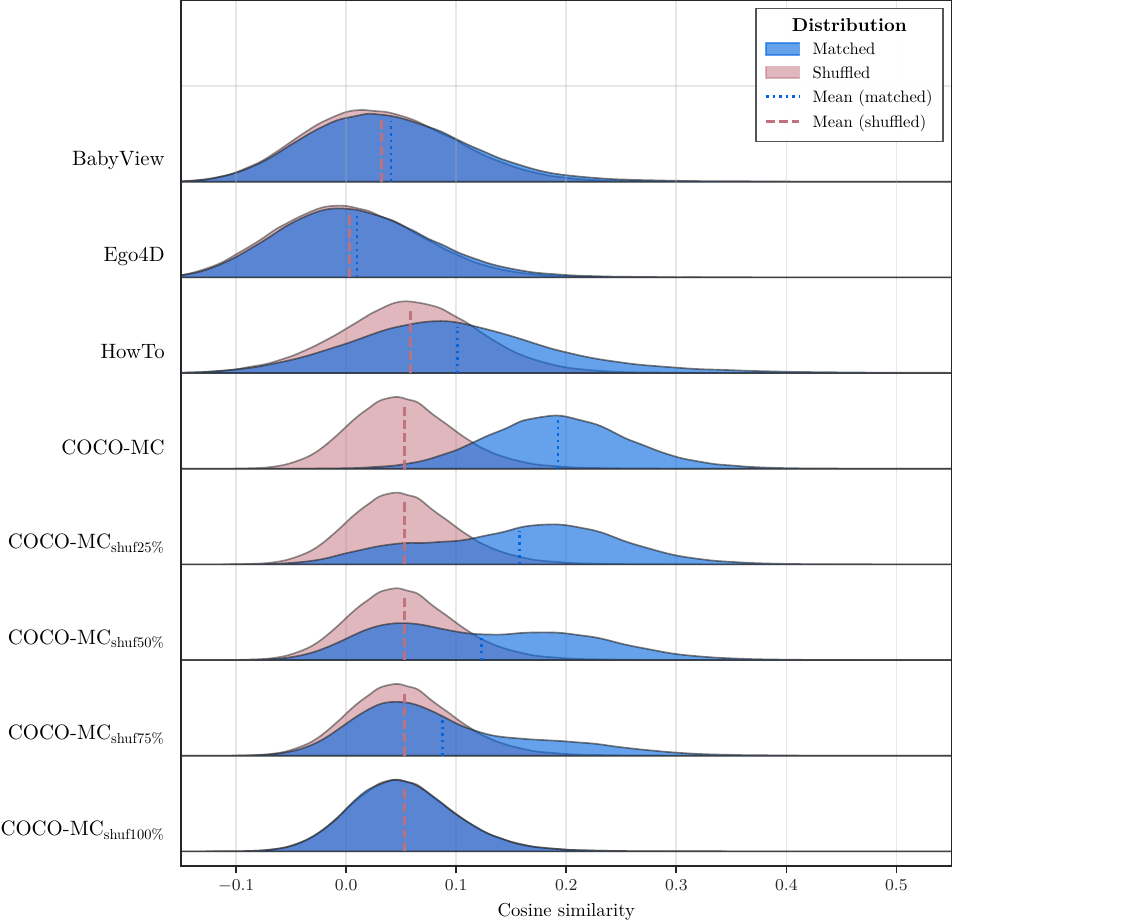}%
    }
    \caption{Semantic textual similarity score distributions (captioning with PLM-8B, text embedding with SONAR).}
    \label{fig:sts_scores}
\end{figure}

Fig.~\ref{fig:cos_sims_clipscore_vitb16}--\ref{fig:sts_scores} show the underlying score distributions (matched and shuffled pairs) for each method and dataset. For the permuted COCO-MC variants, we see how incremental shuffling shifts distribution mass from the mode of the original distribution to the mode of the fully shuffled distribution, with the JSD therefore tending towards $0$. For VQAScore, we see that the metric is well calibrated for COCO-MC, with the original distribution peaking towards $1$ and the fully shuffled distribution peaking towards $0$. The video datasets, on the other hand, are spread out across nearly the entire range between [0, 1], suggesting that sample-level semantic alignment is much more variable in naturalistic videos than in COCO-MC. For the cosine similarity scores, this difference is less pronounced, which may be partly due to relying on unscaled CLIPScores. Overall, however, we see consistent trends across methods. For matched video/image-text pairs, Ego4D and BabyView show the lowest similarity scores across methods, followed by HowTo, while COCO-MC exhibits by far the highest scores. In contrast, for shuffled pairs, the COCO-MC scores tend to be the lowest across datasets (with the exception of Ego4D and BabyView with STS scoring), while the video datasets tend to have slightly higher scores. This result may indicate that the video datasets have more generic transcriptions that are neither particularly descriptive nor completely unrelated to the visual content.

\subsubsection{Robustness check on COCO Karpathy}

A potential concern with the COCO-MetaCLIP results is that COCO-MC was constructed using PLM-VQAScore itself to retrieve high-scoring image--caption pairs from the MetaCLIP pool: the same scoring method that we then used to evaluate the data it helped curate, which is liable to inflate the VQAScore matched distribution. To confirm that our findings are not artifacts of this selection, we additionally scored the original COCO Karpathy training split using the same three pipelines.

Table~\ref{tab:coco_vs_cocomc_robustness} reports matched and shuffled distribution means together with bootstrapped JSD. Despite COCO-MC's selection bias, the absolute JSD scores are remarkably close on all three methods: CLIPScore differs by $0.02$ (0.896 vs.\ 0.916), VQAScore by $0.023$ (0.946 vs.\ 0.969), and STS by $0.042$ (0.657 vs.\ 0.615, with COCO Karpathy slightly higher). Crucially, the qualitative ordering across datasets is preserved -- COCO Karpathy still vastly outranks any of the video datasets on every method -- confirming that our cross-dataset alignment claims are robust to the choice of curation pipeline used to construct COCO-MC.

 \begin{table}[t]
  \centering
  \caption{Alignment-scoring JSD on the original COCO Karpathy split vs.\ COCO-MC, with bootstrapped 95\% CIs ($n{=}1000$). Matched and shuffled distribution means are also reported. Despite the VQAScore-based curation of COCO-MC, results on the independent Karpathy split are very similar across all three methods.}
  \label{tab:coco_vs_cocomc_robustness}
  \small
  \begin{tabular}{@{}llcccc@{}}
  \toprule
  \bfseries Method & \bfseries Dataset & \bfseries Matched mean & \bfseries Shuffled mean & \bfseries JSD & \bfseries 95\% CI \\
  \midrule
  \multirow{2}{*}{CLIPScore (ViT-B/16)}
    & COCO     & 0.307 & 0.150 & 0.896 & [0.892, 0.905] \\
    & COCO-MC  & 0.308 & 0.147 & 0.916 & [0.913, 0.918] \\
  \midrule
  \multirow{2}{*}{VQAScore}
    & COCO     & 0.916 & 0.126 & 0.946 & [0.944, 0.948] \\
    & COCO-MC  & 0.971 & 0.066 & 0.969 & [0.968, 0.971] \\
  \midrule
  \multirow{2}{*}{STS}
    & COCO     & 0.216 & 0.062 & 0.657 & [0.654, 0.660] \\
    & COCO-MC  & 0.192 & 0.053 & 0.615 & [0.611, 0.618] \\
  \bottomrule
  \end{tabular}
  \end{table}



\clearpage

\end{document}